\pdfoutput=1

\documentclass[11pt]{article}

\usepackage[]{acl}
\usepackage{hyperref}

\usepackage{times}
\usepackage{latexsym}
\usepackage{caption}
\usepackage{graphicx}
\usepackage{latexsym}
\usepackage{booktabs}
\usepackage{xcolor,colortbl}
\usepackage{amsmath}
\usepackage{multirow}
\usepackage{paralist}
\usepackage{cleveref}
\usepackage{xspace}
\usepackage[normalem]{ulem}
\usepackage{lipsum}
\useunder{\uline}{\ul}{}

\usepackage{amssymb}
\usepackage{pifont}
\newcommand{\cmark}{\ding{51}}%
\newcommand{\xmark}{\ding{55}}%

\usepackage[T1]{fontenc}

\usepackage[utf8]{inputenc}

\usepackage{microtype}
\usepackage{xurl}
\usepackage{pdflscape}
\usepackage{geometry}
\usepackage{longtable}
\usepackage{enumitem}

\setlist{topsep=2pt,itemsep=2pt,partopsep=2pt, parsep=2pt}

\newcommand{\ex}[1]{\textit{#1}\xspace}

\title{NusaCrowd: Open Source Initiative for Indonesian NLP Resources}

\author{
\textbf{Samuel Cahyawijaya$^{\spadesuit,1,2}$, Holy Lovenia$^{\spadesuit,1,2}$, Alham Fikri Aji$^{\spadesuit,3}$, Genta Indra Winata$^{\spadesuit,4}$,} \\
\textbf{Bryan Wilie$^{\spadesuit,1,2}$, Fajri Koto$^{\spadesuit,3,2}$, Rahmad Mahendra$^{5,2}$, Christian Wibisono$^{6}$}, \\ \textbf{Ade Romadhony$^{7,2}$,
Karissa Vincentio$^{8,2}$, Jennifer Santoso$^{9}$, David Moeljadi$^{10}$,} \\ 
\textbf{Cahya Wirawan$^{12}$, Frederikus Hudi$^{11,20}$, Muhammad Satrio Wicaksono$^{13}$,}, \\ 
\textbf{Ivan Halim Parmonangan$^{14}$, Ika Alfina$^{5}$, Ilham Firdausi Putra$^{13}$, Samsul Rahmadani$^{15}$,} \\ 
\textbf{ Yulianti Oenang$^{13}$, Ali Akbar Septiandri$^{16}$, James Jaya$^{13}$, Kaustubh D. Dhole$^{17}$,} \\ 
\textbf{ Arie Ardiyanti Suryani$^{7}$, Rifki Afina Putri$^{18}$, Dan Su$^{1}$, Keith Stevens$^{19}$,} \\  
\textbf{Made Nindyatama Nityasya$^{13}$, Muhammad Farid Adilazuarda$^{6}$, Ryan Ignatius$^{13}$,} \\ 
\textbf{ Ryandito Diandaru$^{6}$,Vito Ghifari$^{6}$, Tiezheng Yu$^{1}$, Wenliang Dai$^{1}$, Yan Xu$^{1}$,} \\
\textbf{Dyah Damapuspita$^{5}$, Haryo Akbarianto Wibowo$^{13}$, Cuk Tho$^{14}$,} \\ 
\textbf{Ichwanul Muslim Karo Karo$^{21}$, Tirana Noor Fatyanosa$^{22}$, Ziwei Ji$^{1}$, Graham Neubig$^{23}$,} \\
\textbf{Timothy Baldwin$^{3}$, Sebastian Ruder$^{24}$, Pascale Fung$^{1}$, Herry Sujaini$^{25,2}$, } \\
\textbf{Sakriani Sakti$^{26,11}$, Ayu Purwarianti$^{6,27,2}$} \\[7px]
$^{\spadesuit}$Main Authors \\[5px]
$^{1}$HKUST \quad $^{2}$INACL \quad $^{3}$MBZUAI \quad $^{4}$Bloomberg \quad $^{5}$Universitas Indonesia \\ $^{6}$Institut Teknologi Bandung \quad $^{7}$Telkom University \quad
$^{8}$JULO \quad $^{9}$University of Tsukuba \\
$^{10}$Kanda University of International Studies $^{11}$NAIST \quad $^{12}$AI-Research.id \\
$^{13}$Independent Researcher \quad $^{14}$BINUS \quad $^{15}$Bahasa.ai \quad $^{16}$Universitas Al Azhar Indonesia \\
$^{17}$Emory University \quad $^{18}$KAIST \quad $^{19}$Surface Data \quad $^{20}$ Works Applications \\
$^{21}$State University of Medan \quad $^{22}$Kumamoto University \quad $^{23}$CMU \quad $^{24}$Google \\ 
$^{25}$Tanjungpura University \quad $^{26}$JAIST \quad $^{27}$Prosa.ai \\
}

\begin{document}
\maketitle
\begin{abstract}
We present NusaCrowd, a collaborative initiative to collect and unify existing resources for Indonesian languages, including opening access to previously non-public resources. Through this initiative, we have brought together 137 datasets and 118 standardized data loaders. The quality of the datasets has been assessed manually and automatically, and their value is demonstrated through multiple experiments.
NusaCrowd's data collection enables the creation of the first zero-shot benchmarks for natural language understanding and generation in Indonesian and the local languages of Indonesia. Furthermore, NusaCrowd brings the creation of the first multilingual automatic speech recognition benchmark in Indonesian and the local languages of Indonesia.
Our work strives to advance natural language processing (NLP) research for languages that are under-represented despite being widely spoken.

\end{abstract}

\section{Introduction}


Indonesia is one of the most linguistically diverse and populous countries in the world, with over 270 million people living across 18,000+ islands. It covers more than 700 spoken languages, making up $\sim$10\% of all languages in the world~\cite{grimes2000ethnologue,lewis2009ethnologue,cohn2014local}.
However, the progress of NLP research in Indonesian languages has been held back by factors including language diversity~\cite{anderbeck2008malay,haryono2012perubahan,siregar2014code,fauzi2018dialect}, orthographic variation~\cite{Diksi5265}, resource limitation~\cite{wilie2020indonlu,koto2020indolem}, and other societal challenges~\cite{nurjanah2018pengembangan,antara-NTT,aji-etal-2022-one}.

\begin{figure}[t]
    \centering
    \includegraphics[width=\linewidth, trim={24cm 8cm 24cm 8cm}, clip]{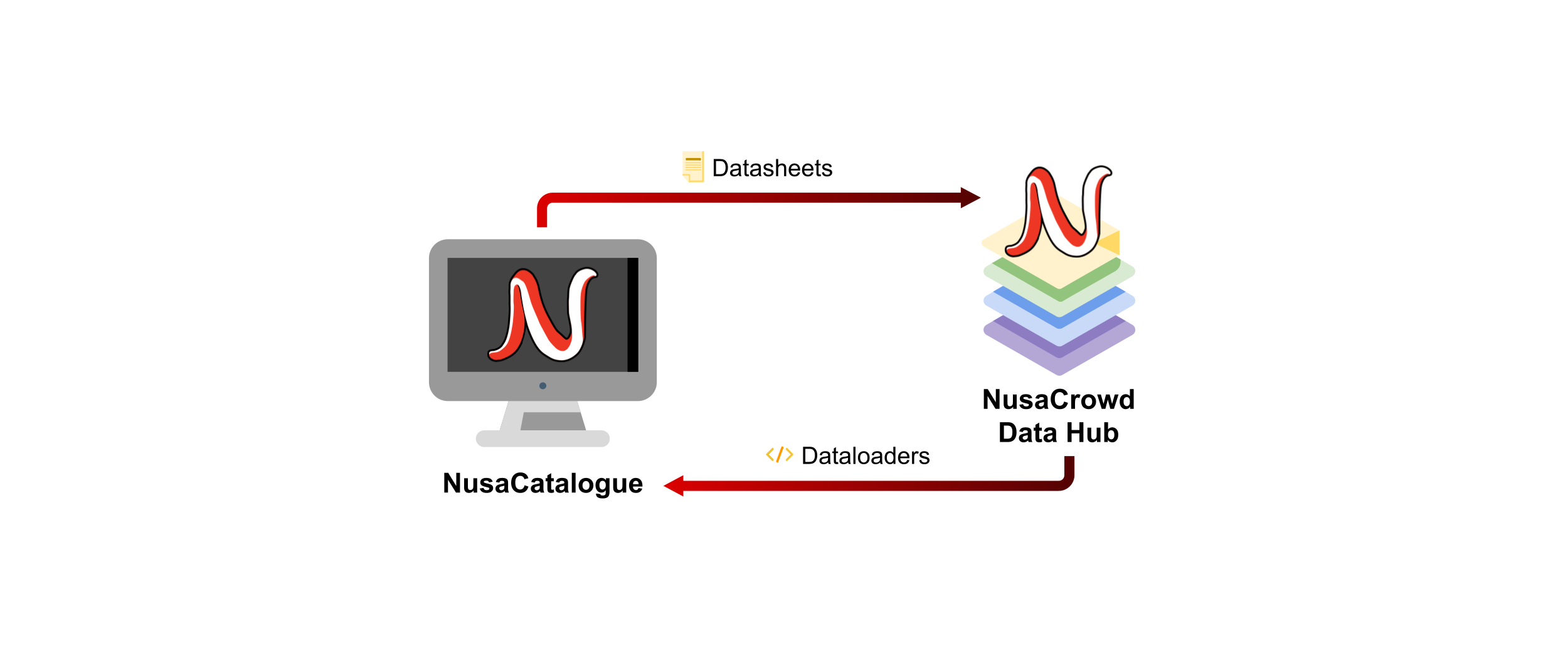}
    \caption{
    System architecture of NusaCrowd. Open access to the datasheets is provided through \textbf{NusaCatalogue}, while dataloader scripts to access the resources are implemented in \textbf{NusaCrowd Data Hub}.}
    \label{fig:nusacatalogue-nusacrowd}
\end{figure}

Existing NLP research mainly focuses on high-resource languages~\cite{wang2018-glue,xu2020clue, ruder2022statemultilingualai}, while the vast majority of languages with limited data---including most languages spoken in Indonesia---are neglected \cite{joshi-etal-2020-state}. Specifically, many Indonesian NLP resources are scattered, undocumented, and not publicly available. These issues cause a severe data scarcity problem, which hinders NLP research in Indonesian and other local languages spoken in Indonesia from progressing.



In this work, we introduce NusaCrowd,\footnote{NusaCrowd is a portmanteau of the words \textbf{Nusantara} and \textbf{Crowd}. The word \textbf{Nusantara} is derived from an old Javanese term referring to the territories of the Majapahit empire that corresponds to present-day Indonesia.} an open collaborative effort to gather and unify existing resources in Indonesian languages for public use, and liberate non-public resources. This initiative has successfully collected a total of 137 datasheets with 118 standardized data loaders in NusaCrowd Data Hub\footnote{We publicly release NusaCrowd's data hub at \url{https://github.com/IndoNLP/nusa-crowd} and the NusaCatalogue at \url{https://indonlp.github.io/nusa-catalogue/}}. 
The datasets were manually assessed for data quality by multiple native speakers and experts in NLP.
Utilizing the datasets collected in NusaCrowd, we introduce the first zero-shot NLU benchmark (\textbf{NusaNLU}), zero-shot NLG benchmark (\textbf{NusaNLG}), and multilingual ASR benchmark (\textbf{NusaASR}) for Indonesian languages. We evaluate various Indonesian and multilingual models on the benchmarks.

Our contributions can be summarized as follows:

\begin{compactitem}
    \item We introduce the first large-scale resource hub of standardized  Indonesian corpora, covering 100+ datasets and 200+ tasks, spanning 19 Indonesian languages in text, speech, and image modalities. As part of this, we provide first-time access to 14 previously private datasets.
    \item 
    We develop the first Indonesian multilingual zero-shot benchmarks for natural language understanding (NusaNLU) and natural language generation (NusaNLG), which cover 40 NLU and NLG tasks in 12 languages. 
    \item We conduct a comprehensive analysis of the collected datasets across various factors. Our analysis reflects the quality and diversity of existing NLP datasets in Indonesian and other languages spoken in the region.
    \item For speech, our initiative opens up access to a wide variety of ASR corpora ($\sim$800 hours) covering 10 Indonesian languages. Using these resources, we build NusaASR and develop various Indonesian monolingual and multilingual ASR models.
\end{compactitem}

\section{Related Work}

\paragraph{Indonesian NLP Resources}
The lack of labeled datasets for training and evaluation has impeded the advancement of NLP research in Indonesian languages~\cite{aji-etal-2022-one}. As a result, research has focused on using unlabeled data by building large language models (LLMs) to enable zero-shot and few-shot transfer learning. In recent years, multiple efforts have worked on language models (LMs) in Indonesian languages by exploring and developing different LM structures. Several efforts have focused on encoder-only LMs, such as IndoBERT~\cite{wilie2020indonlu,koto2020indolem}, SundaBERT~\cite{wongso2022pre}, and IndoBERT-Tweet~\cite{koto2021indobertweet}. Elsewhere, a number of generative models have been proposed, i.e., IndoBART and IndoGPT, along with the generation task benchmark, IndoNLG~\cite{cahyawijaya-etal-2021-indonlg}. 

\paragraph{Open and Community-based Initiatives}
Open source/open science initiatives are a core part of the motivation behind this paper.
Large-scale collaborations have made their mark in various research areas through developing a variety of resources, e.g., LMs~\cite{scao2022bloom,muennighoff2022crosslingual}, datasets~\cite{ardila2020common,adelani2021masakhaner,mager2021findings}, catalogues~\cite{alyafeai2022masader, altaher2022masader, mcmillan2022documenting}, and benchmarks~\cite{srivastava2022beyond,dhole2021nl,fries2022bigbio}.

\section{NusaCrowd}

In this section, we provide an overview of NusaCrowd, a detailed description of the NusaCrowd framework, the dataset curation process, as well as a detailed summary and statistics of the datasets contained in NusaCrowd.

\subsection{Overview of NusaCrowd}
NusaCrowd is a crowdsourcing initiative to collect, open-source, and standardize access to datasets in Indonesian and 700+ local languages in Indonesia. NusaCrowd aims to address the resource limitation problem in Indonesian NLP across three dimensions: (1) complete datasheets for each curated, ready-to-use dataset; (2) an open-access and centralized data hub for accessing datasets through standardized data loading scripts; and (3) promoting public data access for published non-public datasets. Through promoting public data access, NusaCrowd provides access to 14 previously non-public datasets, some of which are multilingual, covering a total of $\sim$40 tasks over 12 languages. It also serves as a portal for retrieving and loading a wide variety of Indonesian NLP datasets, in text and other modalities (e.g., speech and images). NusaCrowd does not store or copy any of the hosted datasets, and control and ownership of the hosted datasets belong to the original owners.


\begin{figure}[!t]
    \centering
    \includegraphics[width=\linewidth]{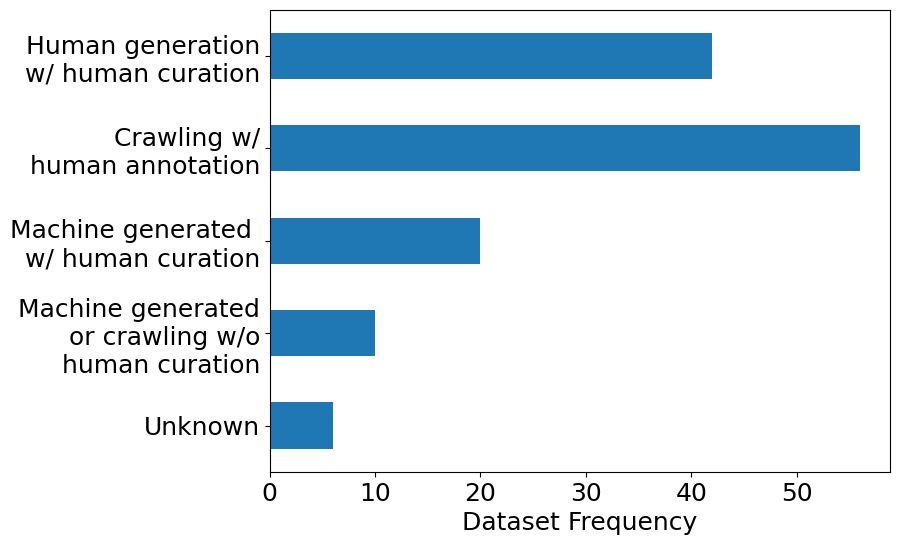}
    \caption{Distribution of dataset curation approaches used in datasets contained in NusaCrowd.}
    \label{fig:annotation}
\end{figure}

\subsection{NusaCrowd Framework}

As shown in \Cref{fig:nusacatalogue-nusacrowd}, NusaCrowd consists of two platforms: NusaCatalogue
and NusaCrowd Data Hub.
The two platforms interact to support dataset registration and provide a standardized pipeline for NusaCrowd. In general, NusaCatalogue stores the datasheets (metadata) of all datasets, and NusaCrowd Data Hub stores the standardized data loaders for all of the datasets. The two systems share information about the datasheets and the data loaders, enabling users to seamlessly explore and use the datasets.\footnote{All code in NusaCrowd will be made publicly available under Apache License 2.0.}

\paragraph{NusaCrowd Workflow} The dataset registration and standardization pipeline in NusaCrowd consists of four stages: (1) submission of datasheet information through an online form;
(2) manual curation of the datasheet information by an expert in NLP, which, once approved (\Cref{sec:data-curation}), is made available via the \textbf{NusaCatalogue} portal and a data loader implementation request is submitted to \textbf{NusaCrowd Data Hub}; (3) implementation of a data loader; and (4) review and approval of the implemented data loader by two maintainers, which is then published on \textbf{NusaCrowd Data Hub}. In addition to the datasheets, we also provide instructions on how to use the data in \textbf{NusaCatalogue}.

\subsection{Dataset Standardization and Curation}
\label{sec:data-curation}

We standardize the tasks from the datasets in NusaCrowd into several categories according to a specific schema, defined as a common set of attributes required to perform the task. We use the schema to cover similar tasks across the datasets. We define 13 schemas to cover all the tasks and modalities in the datasets, e.g., text classification, text generation, image captioning, and speech recognition. For instance, in the single-label text classification schema (\texttt{TEXT}), each example consists of three attributes \texttt{(id, text, label)}, where \texttt{id} denotes a unique row identifier, \texttt{text} denotes the input text, and \texttt{label} denotes a discriminative target variable. We elaborate on the attributes of each schema in \Cref{app:task-schema}.

\begin{table}[!t]
\small
\centering
  \resizebox{0.49\textwidth}{!}{
      \begin{tabular}{@{ ~ }c@{ ~ }c@{ ~ }c@{ ~ }c@{ ~ }c@{ ~ }c@{}}
      \toprule
        \multirow{2.5}{0.1\textwidth}{\centering \textbf{Language}} & \multicolumn{2}{c}{\textbf{langid.py}} & \multicolumn{2}{c}{\textbf{FastText}} & \textbf{CLD3} \\ \cmidrule(l{2pt}r{2pt}){2-3} \cmidrule(l{2pt}r{2pt}){4-5} \cmidrule(l{2pt}r{2pt}){6-6} 
         & \textbf{Top-1} & \textbf{Top-3} & \textbf{Top-1} & \textbf{Top-3} & \textbf{Top-1} \\
         \midrule
        Eng & 98.33 & 99.33 & 94.05 & 99.03 & 99.69 \\
        Ind & 72.11 & 90.39 & 82.42 & 89.92 & 60.27 \\
        Sun & --- & --- & 34.28 & 75.21 & 50.53 \\
        Jav & 48.97 & 79.07 & 28.08 & 69.43 & 46.88 \\
        \bottomrule
      \end{tabular} 
  }
\caption{Language identification accuracy based on different languages. For Sundanese and Javanese, several datasets consist of informal Indonesian utterances including Ind--Sun and Ind--Jav code-mixed sentences.}
  \label{tab:language-check}
\end{table}

\begin{figure*}[!t]
    \centering
    \includegraphics[width=\linewidth, trim={0 0 0 0}, clip]{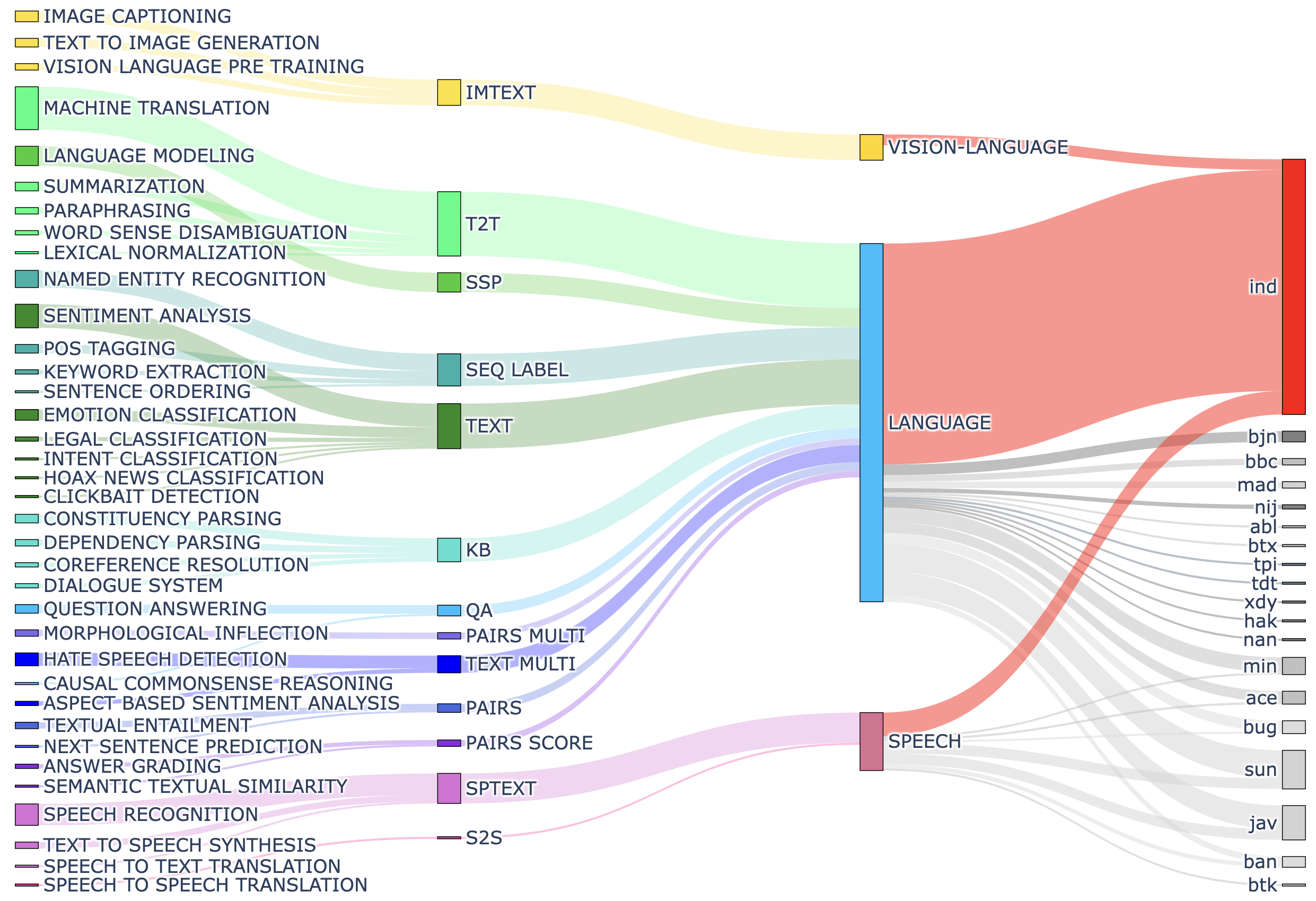}
    \caption{Summary of tasks, schemas, modalities, and languages\footnotemark\hspace{0.25mm} in NusaCrowd. $\sim$75\% of the datasets are textual language data in Indonesian, with the other two modalities being vision-language and speech. Textual language data covers 19 Indonesian languages (Indonesian and 18 other languages spoken in the region), the speech data covers 8 languages (Indonesian and 7 local languages), while vision-language data only covers Indonesian.}
    \label{fig:nusa-sankey}
\end{figure*}

To assess the quality of the datasets in NusaCrowd, we perform manual curation for each datasheet submission based on two criteria: language correctness, and the annotation process. We provide the results as metadata for each dataset. We check the correctness of the reported language using off-the-shelf language identification (LID) tools. We perform LID in 4 languages: English, Indonesian, Sundanese, and Javanese,  We measure the LID accuracy compared to the reported languages in the metadata on all tasks containing text modality in NusaCrowd. Since many datasets consist of a large number of samples, language correctness checking is done both automatically and manually. 

We conduct automatic language identification for 4 languages, i.e., English, Indonesian, Sundanese, and Javanese~\footnote{We only perform language identification as these are the only languages supported by most of the existing off-the-shelf language identification tools.} using 3 off-the-shelf language identification tools, i.e., langid.py~\cite{lui2012langid}, FastText LID~\cite{ooms2022cld3}, and Google CLD3~\cite{ooms2022cld3}. For other languages, since there is no language identification library available, the curation is done manually through sampling. Based on the automatic language identification result in \Cref{tab:language-check}, the correctness of languages is quite high, indicated by the top-3 accuracy of each language identification tools ~\footnote{we don’t consider Top-1 for Sundanese and Javanese since the languages are low-resource and often mispredicted}. Additionally, the accuracy of Indonesian is not as high as English, we conjecture that this is caused by there are many English terms from tasks that are collected from online platforms.

For assessing the annotation process for each dataset, we manually check the dataset annotation process from relevant publications and/or other descriptions and classify them into five categories, i.e., \textit{human-generated}, \textit{crawling with human annotation},
\textit{machine-generated with human curation},
\textit{machine-generated or crawling without human curation}, and
\textit{unknown}. The statistics of the dataset annotation assessment are shown in \Cref{app:data-description}. In general, $\sim$90\% of all the datasets listed in NusaCrowd are human-curated, showing that most of the datasets in NusaCrowd are high-quality and well-suited for building and evaluating Indonesian NLP models. Moreover, almost half of the datasets are collected through crawling and are annotated manually by humans, usually for NLP tasks such as sentiment analysis, emotion recognition, hate speech detection, named entity recognition, and machine translation. The crawling often comes from sources such as social media, news platforms, online reviews, etc. 



\subsection{Datasets in NusaCrowd}
\label{sec:dataset-nusa}

NusaCrowd includes 137 datasheets and 118 dataloaders, including access to 14 previously non-public datasets, and a variety of tasks and languages. We list all of the previously private datasets in \Cref{app:private-dataset}. NusaCrowd covers 36 task types, including: machine translation, summarization, sentiment analysis, part-of-speech (POS) tagging, and question answering, which are standardized into 13 different schemas. The datasets in NusaCrowd stem from three modalities---image, text, and speech---with the majority of the data coming from the text modality. In terms of languages, NusaCrowd covers 19 Indonesian languages, i.e., Indonesian and 18 regional languages, in addition to some non-Indonesian languages such as Japanese, English, Spanish, and Russian, which come into the mix as machine translation language pairs. A summary of the datasets is shown in \Cref{fig:nusa-sankey}. A list of language codes with the complete language name and family is provided in \Cref{app:language-nusacrowd}. We present comprehensive details of the datasets in \Cref{app:data-description}, and a comparison of NusaCrowd with other initiatives in \Cref{app:initiative-comparison}.

\paragraph{Modalities}
NusaCrowd comprises datasets from three different modalities, i.e., image, text, and speech, all of which are related to language tasks. Most datasets contain text data for natural language understanding (e.g., sentiment analysis, named entity recognition, and parsing) and natural language generation tasks (e.g., machine translation, paraphrasing, and abstractive summarization). These account for 29 out of 36 task types in NusaCrowd. In addition, NusaCrowd covers three vision tasks: vision-language pre-training, image captioning, and text-to-image generation. For speech, NusaCrowd covers four tasks: automatic speech recognition (ASR), text-to-speech synthesis (TTS), speech-to-text translation (S2T), and speech-to-speech translation (S2S).

\footnotetext{Based on ISO639-3 language codes: \url{https://iso639-3.sil.org/code_tables/639/data}.}


\begin{figure*}[!t]
\centering
\includegraphics[width=0.515\linewidth, trim={0 0 0 0}, clip]{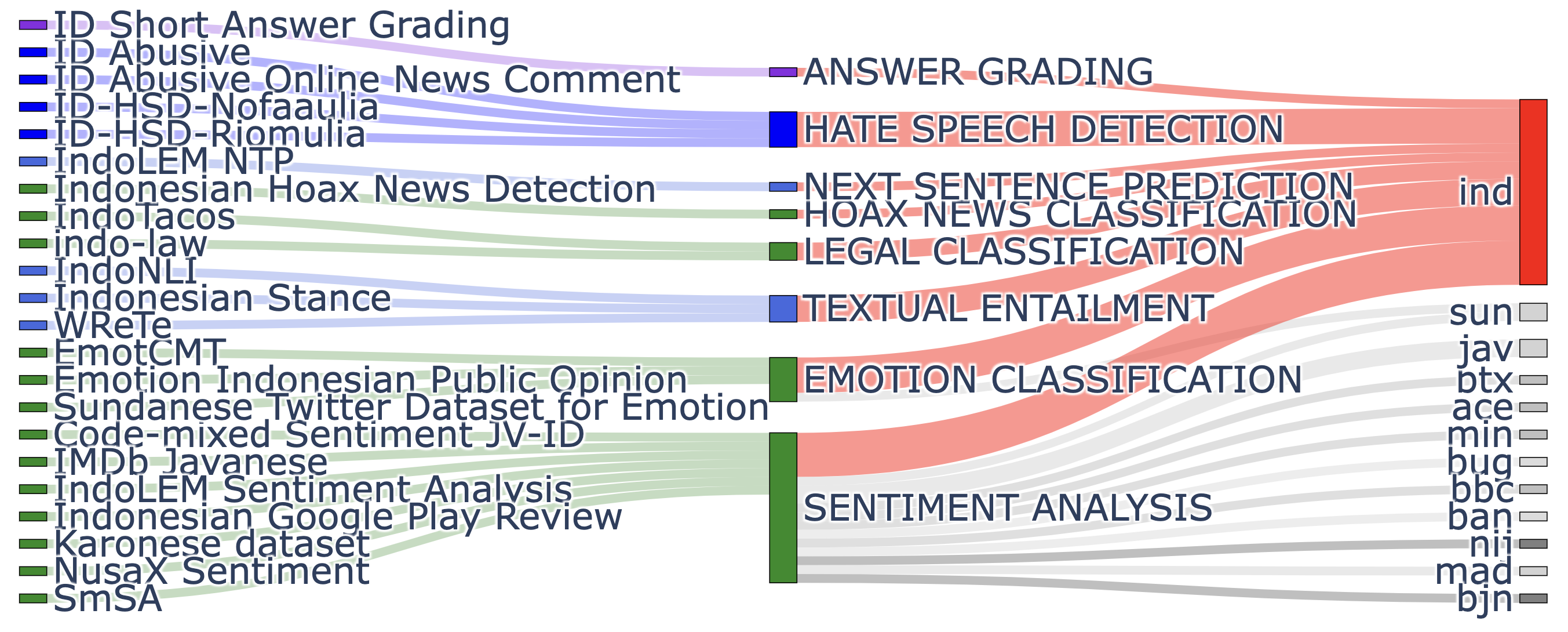}
\includegraphics[width=0.475\linewidth, trim={0 0 22.6cm 0}, clip]{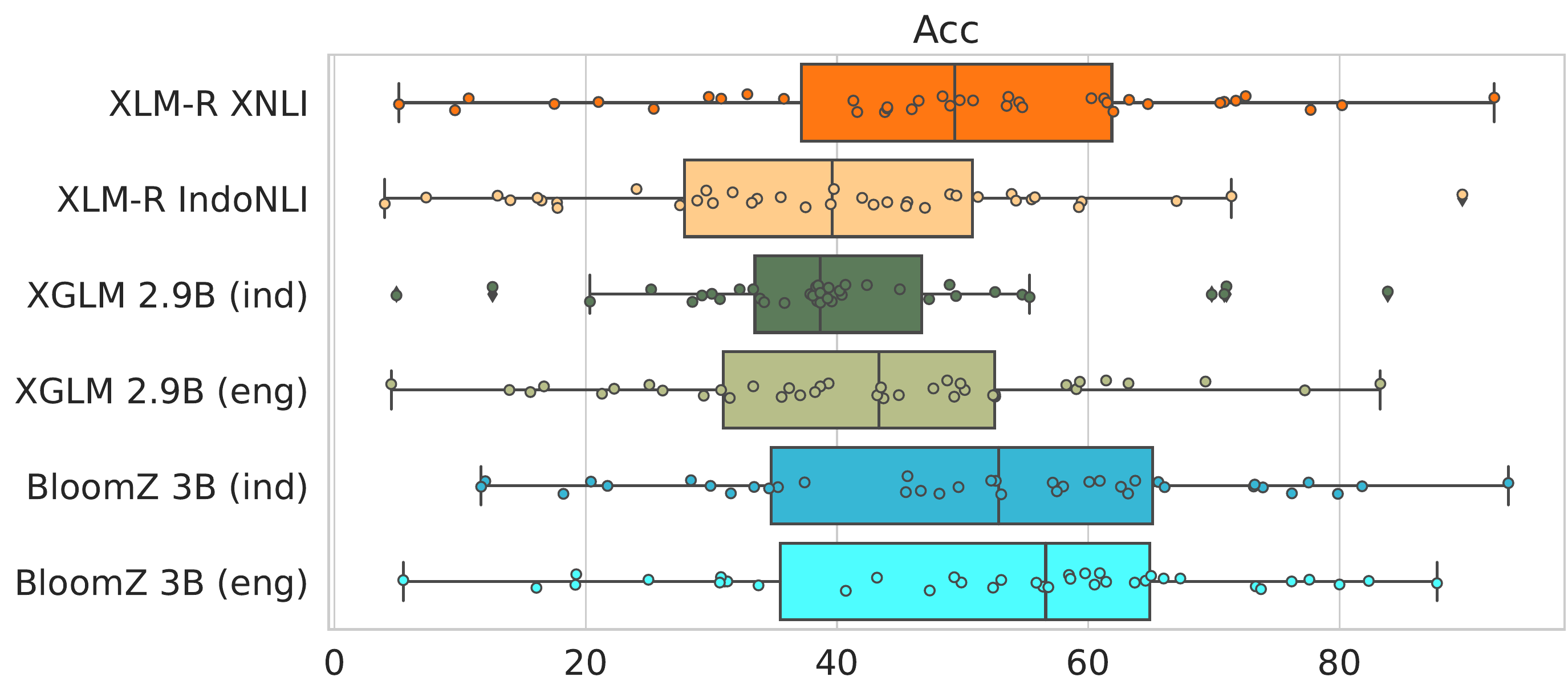}
\caption{\textbf{(left)} The datasets used in NusaNLU and \textbf{(right)} zero-shot generalization in NusaNLU. Box plots show summary statistics of accuracy scores. For XGLM and BLOOMZ, each point denotes average per-dataset performance using three different prompts. \textbf{(ind)} and \textbf{(eng)} denote the prompt language used for prompting, i.e., Indonesian and English, respectively.}
\label{fig:zeroshotnlu}
\end{figure*}

\paragraph{Languages} NusaCrowd covers Indonesian and 18 regional languages. Most languages covered in NusaCrowd belong to the Austronesian language family, 14 of which are part of Malayo-Polynesian family (including Indonesian), 2 of which are creole languages, i.e., Tok Pisin (tpi) and Tetun Dili (tdt).\footnote{The two languages are not spoken in Indonesia, but instead used in neighboring countries: Papua New Guinea and Timor Leste.} The other two languages --- Hakka/Khek (hak) and Min Nan (nan) with Teochew dialect --- are Sinitic and belong to the Sino-Tibetan language family. Detailed descriptions of each language are provided in \Cref{app:language-nusacrowd}.



\section{NusaCrowd Benchmarks}

To showcase the benefit of NusaCrowd, we develop three different benchmarks from subsets of the datasets. Specifically, we develop benchmarks for Indonesian and other local languages including a zero-shot NLU benchmark (NusaNLU), a zero-shot NLG benchmark (NusaNLG), and a multilingual ASR benchmark (NusaASR).

\subsection{NusaNLU}
\label{sec:nusanlu}


Existing benchmarks~\cite{wilie2020indonlu,koto2020indolem} in Indonesian NLU only cover one language, i.e., the national language, Indonesian. Moreover, these benchmarks only focus on comparing traditional machine learning approaches with the fine-tuning approaches of pre-trained LMs. Following recent work in other high-resource languages that explore zero-shot generalization of large LMs~\cite{scao2022bloom,lin2021few,muennighoff2022crosslingual,fries2022bigbio}, we develop NusaNLU, the first zero-shot NLU benchmark in Indonesian and regional languages to benchmark zero-shot techniques over 26 datasets using both Indonesian monolingual and multilingual LMs. NusaNLU covers 12 languages across various tasks, including 3 emotion classification tasks~\cite{saputri2018emotion,Yulianti2021-kx,ricossan2022emotpubopini}, 18 sentiment analysis tasks~\cite{winata-etal-2023-nusax,nurlaila2017natashacare,hidayatullah2020attention,wongso2021imdbjv,koto2020indolem,purwarianti2019improving}, one review score rating task\footnote{\url{https://huggingface.co/datasets/jakartaresearch/google-play-review}}, one hate speech detection task~\cite{ibrohim2019multihs}, one abusive language detection task~\cite{putri2021abusive}, one next tweet prediction task~\cite{koto2020indolem}, and one natural language inference (NLI) task~\cite{mahendra2021indonli}. A visual overview of the datasets in NusaNLU is provided in \Cref{fig:zeroshotnlu}.

\paragraph{Models} We evaluate three state-of-the-art multilingual language models: XLM-R~\cite{conneau2020unsupervised}, XGLM~\cite{lin2021few}, and BLOOMZ~\cite{muennighoff2022crosslingual}. We generally evaluate in a zero-shot cross-lingual transfer setting \cite{hu2020xtreme}. For XLM-R, we employ intermediate-task training on NLI by predicting the entailment relation between the input text and the label \cite{phang-etal-2020-english}. We explore both XLM-R fine-tuned on XNLI~\cite{conneau2018xnli} and Indonesian IndoNLI~\cite{mahendra2021indonli}. For XGLM and BLOOMZ, we employ zero-shot prompt-based learning with prompts in English and Indonesian. For each language and task, we employ three different prompts and take the average score for the evaluation of each task. More details about fine-tuning hyperparameters and the prompt used in the NLU experiments are shown in \Cref{app:exp-zeroshot-nlu}.

\paragraph{Results} \Cref{fig:zeroshotnlu} shows the zero-shot NLU results of all the models. Overall, the prompting performance of BLOOMZ outperforms other models. Prompting with BLOOMZ outperforms XGLM by a huge margin, providing evidence of the benefit of instruction tuning for prompting. Interestingly, zero-shot cross-task transfer using XLM-R trained on XNLI (XLM-R XNLI) outperforms prompting using XGLM and performs on par with prompting using BLOOMZ, despite the huge difference in their model sizes. This suggests that large LMs are not always needed to perform zero-shot NLU tasks and better efficiency can be achieved through cross-task transfer using much smaller models while achieving similar performance.

Comparing the performance of cross-task fine-tuning across monolingual and multilingual NLI, XLM-R XNLI (122k training instances) outperforms XLM-R IndoNLI (11k training instances) by a large margin, suggesting that using large-scale multilingual data is more effective than using smaller-scale data from closely-related or even the same language to fine-tune a multilingual model in a zero-shot cross-task setting. Comparing the language of the prompts, both BLOOMZ and XGLM with English prompts perform better than the corresponding models with Indonesian prompts. Our findings align with prior work~\cite{muennighoff2022crosslingual,lin2021few,shi2022language}, which shows that, in most cases, the corresponding models perform better in English than on human-translated prompts, despite the language distance between the prompt template and the corresponding text data. 

\begin{figure}[!t]
    \centering
    \resizebox{\linewidth}{!}{
    \includegraphics[trim={0 5 0 45}, clip]{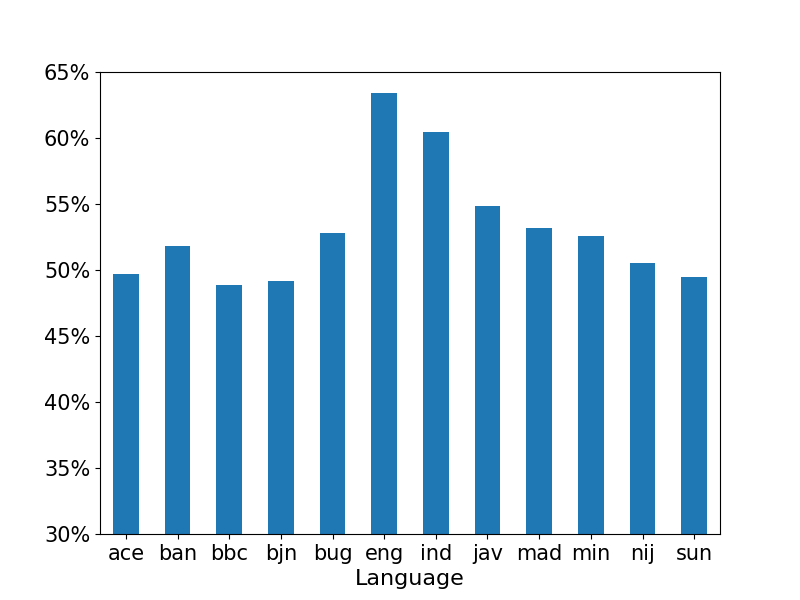}
    }
    \caption{Average zero-shot performance per language across all models on the NusaX subset. All models achieve higher scores for Indonesian (ind) and English (eng).}
    \label{fig:nlu-per-lang}
\end{figure}

Comparing the performance across different languages, as shown in \Cref{fig:nlu-per-lang}, we can conclude that the performance of all models is generally better for Indonesian and English compared to regional Indonesian languages, suggesting that existing multilingual models are unable to generalize well on these languages, and better language representations are vital to close the gap. A full breakdown of per-task performance is provided in \Cref{app:zeroshot-nlu}.

\begin{figure*}[!t]
\centering
    \includegraphics[width=0.515\linewidth, trim={0 0 0 0}, clip]{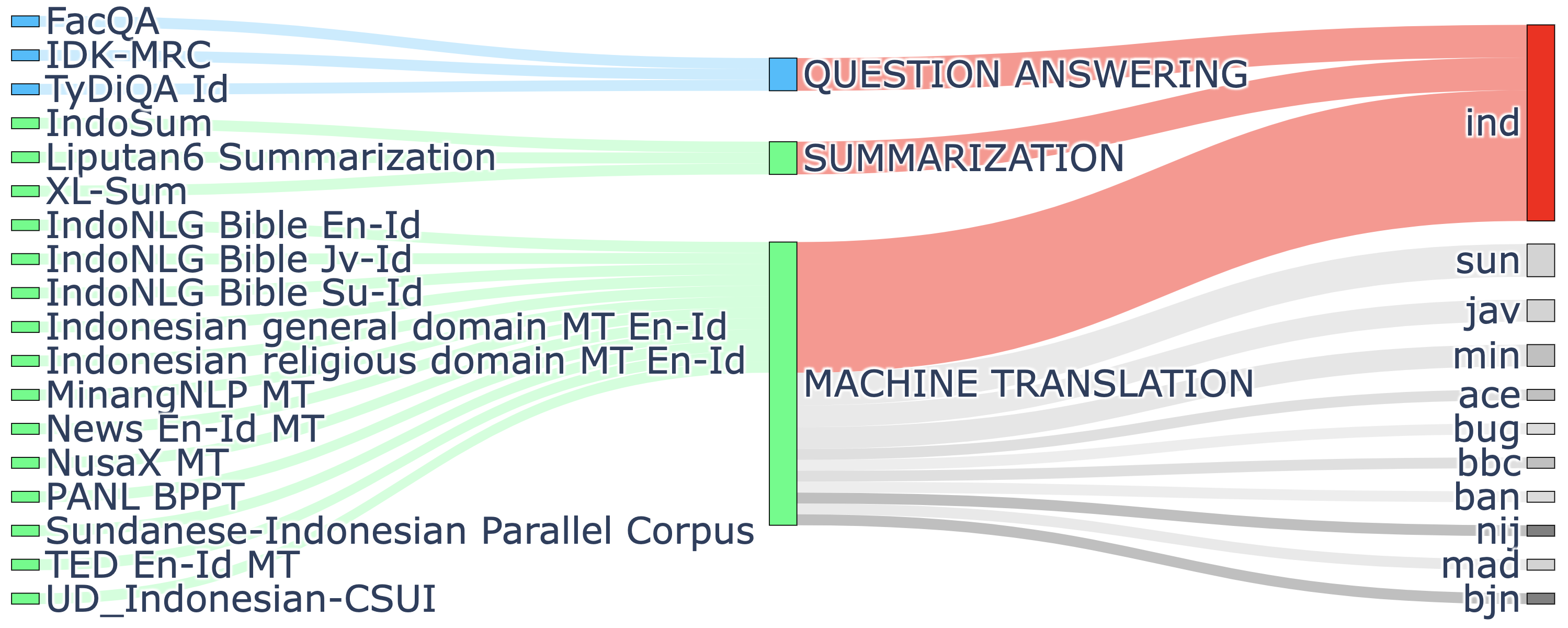}
    \includegraphics[width=0.475\linewidth, trim={0 0 23.5cm 0}, clip]{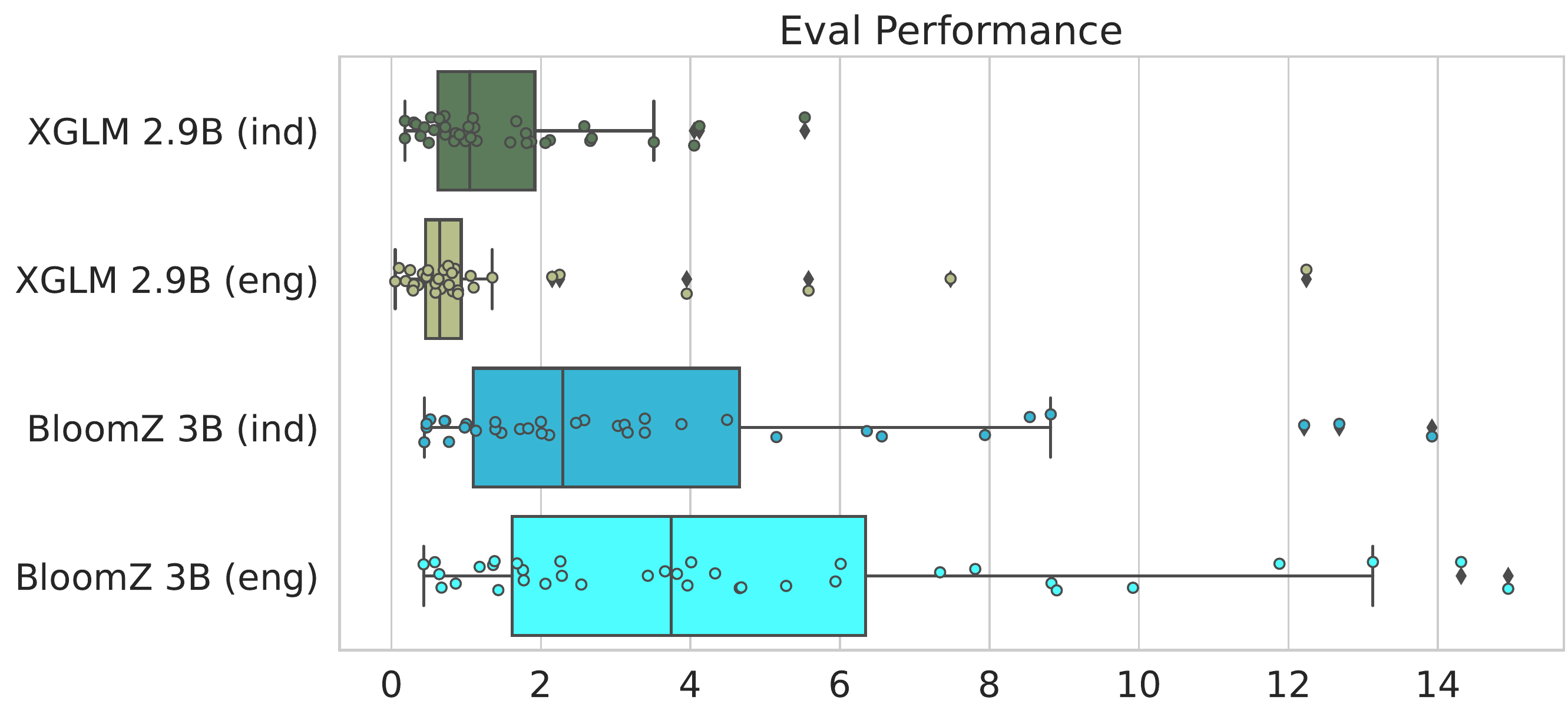}
    \caption{\textbf{(left)} The datasets used in NusaNLG and \textbf{(right)} Zero-shot generalization to machine translation and summarization tasks in NusaNLG. Box plots show summary statistics of the evaluation performance. Points are per-dataset scores from the average of performances over 3 different prompts. \textbf{(ind)} and \textbf{(eng)} denote the prompt language used for prompting, i.e., Indonesian and English, respectively.}
    \label{fig:zeroshotnlg}
\end{figure*}

\subsection{NusaNLG}
\label{sec:nusanlg}

Recent work on Indonesian NLG benchmarks~\cite{cahyawijaya-etal-2021-indonlg,guntara2020benchmarking} has employed transformer-based models, both decoder-only (e.g., IndoGPT) and encoder--decoder (e.g., IndoBART) architectures. To further broaden NLG research in Indonesian and other regional languages, we develop an NLG benchmark, NusaNLG, which covers NLG tasks in 12 languages including English, Indonesian, and 10 local languages. NusaNLG incorporates a total of 36 datasets across various tasks covering 33 machine translation tasks~\cite{guntara2020benchmarking,cahyawijaya-etal-2021-indonlg} and 3 summarization tasks~\cite{kurniawan2018indosum,koto2020liputan6} (\Cref{fig:zeroshotnlg}). We use SacreBLEU for machine translation evaluation, and ROUGE-L for summarization evaluation.

\paragraph{Models} Following recent work in prompting, we explore the possibility of zero-shot generalization of various large LMs on generation tasks through prompting on two NLG tasks, i.e., machine translation and summarization. To explore the effect of different prompt languages on the zero-shot generalization performance, we evaluate prompts in English and Indonesian. We employ two large LMs: XGLM~\cite{lin2021few}, and BLOOMZ~\cite{muennighoff2022crosslingual}. For each task and prompt language, we provide three different prompts and average the result. More details about the hyperparameters and the prompt used in the NLG experiments are shown in \Cref{app:exp-zeroshot-nlg}.

\paragraph{Results} The zero-shot NLG results of all models are shown in \Cref{fig:zeroshotnlg}. Outputs obtained by prompting BLOOMZ outperform those obtained from XGLM for both English and Indonesian prompts. The performance is better on average when prompting BLOOMZ with English prompts than when using Indonesian prompts, which aligns with the results of BLOOMZ on XNLI~\cite{conneau2018xnli}, where BLOOMZ with English prompts performs better than the human-translated prompts~\cite{muennighoff2022crosslingual}.

\begin{table}[t]
    \centering
    \resizebox{0.9\linewidth}{!}{
        \begin{tabular}{lcc}
        \toprule
        \textbf{Language} & \textbf{Ind prompt} & \textbf{Eng prompt} \\
        \midrule
        \textbf{eng $\rightarrow$ ind} & 5.11 & 6.04 \\
        \textbf{ind $\rightarrow$ eng} & 4.65 & 7.90 \\
        \textbf{local $\rightarrow$ ind} & 2.11 & 2.72 \\
        \textbf{ind $\rightarrow$ local} & 1.66 & 2.96 \\
        \bottomrule
        \end{tabular}
    }
    \caption{Average SacreBLEU performance of BLOOMZ for different language pairs. \textbf{Local} denotes all Indonesian local languages in NusaCrowd.}
    \label{tab:bloom-nlg-translation}
\end{table}

Prompting using XGLM yields better quality outputs using Indonesian language prompts than using English prompts. A similar result is reported in XGLM evaluation for Spanish XNLI and Chinese XCOPA~\cite{ponti2020xcopa}, which shows that prompting with the human-translated prompt to the target language produces a better score than the English one. For the BLOOMZ models, the result for English is better since we use the BLOOMZ checkpoint fine-tuned only on English prompts. Additionally, we found that the zero-shot translation quality across all models and prompt languages is poor, especially for local languages, as shown in \Cref{tab:bloom-nlg-translation}. This is even more severe when local languages are involved, yielding $\sim$2\% SacreBLEU. This finding suggests that existing large multilingual LMs still fail to learn representations for these local languages. A full breakdown of per-task results over NusaNLG is provided in \Cref{app:zeroshot-nlg}.

\begin{table*}[]
\centering
\resizebox{0.92\linewidth}{!}{%
\begin{tabular}{lcccccccc}
\toprule
\multicolumn{1}{c}{\textbf{Model}} & \textbf{ace} & \textbf{ban} & \textbf{btk} & \textbf{bug} & \textbf{ind} & \textbf{jav} & \textbf{min} & \textbf{sun} \\ \toprule
\multicolumn{9}{c}{\textit{\textbf{Single-task Training}}} \\ \midrule
wav2vec 2.0-pt & 100.00 & 71.99 & 64.77 & 100.00 & 12.51 & 85.78 & 100.00 & 83.01 \\
wav2vec 2.0-ft & {\ul 49.31} & {\ul 28.74} & {\ul 40.92} & 90.09 & {\ul 2.13} & {\ul 32.11} & {\ul 24.29} & {\ul 26.62} \\
 \midrule
\multicolumn{9}{c}{\textit{\textbf{Monolingual Multi-task Training}}} \\ \midrule
wav2vec 2.0-pt (ind) & 95.14 & $>$100 & $>$100 & {\ul 96.70} & \cellcolor{yellow!25}4.20 & $>$100 & {\ul 46.19} & $>$100 \\
wav2vec 2.0-pt (jav) & $>$100 & 67.02 & 81.24 & $>$100 & 88.87 & \cellcolor{yellow!25}46.97 & 68.10 & 69.89 \\
wav2vec 2.0-pt (sun) & 92.36 & 82.37 & 74.67 & $>$100 & 91.22 & 93.43 & 98.57 & \cellcolor{yellow!25}40.42 \\ \midrule
wav2vec 2.0-ft (ind) & 91.67 & $>$100 & $>$100 & $>$100 & \cellcolor{yellow!25}\textbf{1.87} & $\geq$100 & 70.48 & $>$100 \\
wav2vec 2.0-ft (jav) & 90.28 & {\ul 52.63} & {\ul 59.79} & $>$100 & 78.87 & \cellcolor{yellow!25}{\ul 27.23} & 52.86 & 54.31 \\
wav2vec 2.0-ft (sun) & {\ul 89.58} & 76.52 & 61.34 & $>$100 & 89.59 & 88.50 & 79.05 & \cellcolor{yellow!25}{\ul 25.11} \\ \midrule
\multicolumn{9}{c}{\textit{\textbf{Multilingual Multi-task Training}}} \\ \midrule
wav2vec 2.0-pt & 40.85 & \textbf{16.73} & \textbf{18.98} & \textbf{41.59} & 8.05 & \textbf{18.57} & \textbf{16.94} & \textbf{13.93} \\
wav2vec 2.0-ft & \textbf{31.94} & 21.05 & 35.99 & 53.30 & {\ul 1.90} & 27.55 & 18.10 & 20.79 \\

\bottomrule
\end{tabular}%
}
\caption{Speech recognition results in terms of average word error rate (WER) per language over NusaASR (lower is better). For monolingual multi-task training, the language in brackets denotes the language used for training. \textbf{Bold} denotes the best performance across all groups. {\ul{Underline}} denotes the best performance within the group. In monolingual multi-task training, \colorbox{yellow!25}{Highlight} denotes that the model is trained in the corresponding language.
} 
\label{tab:asrmain}
\end{table*}

\subsection{NusaASR}
\label{sec:nusaasr}

In addition to zero-shot benchmarks for textual language data, we showcase the benefit of NusaCrowd by extending the NLP benchmark in Indonesian languages to speech. 
We develop the first multilingual ASR benchmark for Indonesian and other local languages covering 17 ASR datasets in eight languages: 5$\times$Indonesian (ind), 3$\times$Sundanese (sun), 3$\times$Javanese (jav), 2$\times$Balinese (ban), 1$\times$Acehnese (ace), 1$\times$Batak (btk), 1$\times$Buginese (bug), and 1$\times$Minangkabau (min).

\paragraph{Models} We employ pre-trained wav2vec 2.0~\cite{baevski2020wav2vec} models in our experiment. We explore three training settings: single-task monolingual training, where we fine-tune and evaluate the model on the corresponding ASR dataset; multi-task monolingual training, where we fine-tune the model using multiple ASR datasets on a single language (we evaluate three languages with the largest resources, i.e., Indonesian, Javanese, and Sundanese); and joint multi-task multilingual training, where we fine-tune the model using all 17 ASR datasets listed on NusaASR. We experiment with two wav2vec 2.0$_{LARGE}$ ($\sim$300M parameters) checkpoints, i.e., an unsupervised pre-trained XLS-R wav2vec 2.0 (\textbf{wav2vec 2.0-pt})\footnote{\url{https://huggingface.co/facebook/wav2vec2-large-xlsr-53}} and an Indonesian, Javanese, and Sundanese ASR fine-tuned XLS-R wav2vec 2.0 (\textbf{wav2vec 2.0-ft}).\footnote{\url{https://huggingface.co/indonesian-nlp/wav2vec2-indonesian-javanese-sundanese}} In addition to wav2vec 2.0, we also employ Whisper$_{SMALL}$~\cite{radford2022whisper}\footnote{\url{https://huggingface.co/openai/whisper-small}} ($\sim$250M parameters). Full details of the experiment setup are provided in \Cref{app:exp-multilingual-asr}.

\paragraph{Results} 
\Cref{tab:asrmain} shows the per-language task-averaged performances of wav2vec 2.0 models over NusaASR. The complete per-task results of NusaASR along with the performance of Whisper$_{SMALL}$ are provided in \Cref{app:nusaasr-results}. Based on the results, single-task training on \textbf{wav2vec 2.0-pt} performs poorly due to the limited training data to adapt from unsupervised contrastive pre-training to the ASR task, while the ASR fine-tuned \textbf{wav2vec 2.0-ft} model yields decent results in most languages, except for Buginese (bug) with 90.09\% WER. This suggests limited transferability from  Indonesian, Sundanese, and Javanese to Buginese, consistent with the analysis from NusaX~\cite{winata-etal-2023-nusax} regarding the low overlap between Buginese and other local languages included in NusaCrowd. For monolingual multi-task training, all models perform well only in the languages that they were trained on. This shows that there is a large difference between vocabulary and speech features from one language to another.

For all models evaluated over NusaASR (\textbf{wav2vec 2.0-pt}, \textbf{wav2vec 2.0-ft}, and \textbf{Whisper}), the best performance is achieved through multilingual multi-task training, yielding as low as $\sim$20\% average WER across all languages, 
suggesting transferability of speech features from one language to the others~\cite{fung1998salsa,plu2000salsa,sakti2012multilingual-st,nakayama2019multilingaulspeechchain}. Unlike prior work~\cite{winata-etal-2023-nusax}, where Acehnese (ace) yields similar performance to other languages in sentiment analysis, the same behavior is not reflected in ASR. This suggests that there is a distinction between the speech of Acehnese (ace) and other regional languages, despite vocabulary overlap and shared language structure.



\section{Discussion}


\paragraph{Multilinguality for Low-Resource Languages}
Despite the higher pre-training cost relative to monolingual LMs~\cite{cahyawijaya-etal-2021-indonlg}, multilingual LMs are more versatile and transferable. Recent low-resource monolingual language LMs are on the scale of a hundred million parameters, while the size of multilingual LMs, within a period of three years, has increased by around 1,000$\times$ from $\sim$100M to $\geq$100B parameters~\cite{devlin2019bert,xue2021mt5,tang2021multilingual,muennighoff2022crosslingual,scao2022bloom}. This benefit comes from the data scale of multilingual LMs, which is orders of magnitude larger than monolingual LMs. Additionally, multilingual LMs benefit from positive transfer between related languages, which is especially beneficial for low-resource languages. Moving forward, we expect that multilingual LMs will play a significant role in the exploration of low-resource languages. 

\paragraph{Viability of Large Models for Indonesian}
Computational resources are limited among Indonesian research institutions and in industry, even among the top Indonesian universities~\cite{indonesia2020national,nityasya2020costs}. Focusing solely on large LMs will limit accessibility, and adoption will likely be low.
Therefore, although larger LMs empirically offer better quality, we instead suggest investing more effort in efficiency. This includes smaller sizes LMs and modularized LMs~\cite{pfeiffer2020madx,ansell2021madg, pfeiffer2022lifting}. Furthermore, more work on efficiency through factorization~\cite{winata2020lrt,cahyawijaya2021greenformer}, pruning~\cite{frankle2019lottery,dai2021multimodal}, quantization~\cite{shen2020qbert,aji2020compressing}, or distillation~\cite{zhang2020ternarybert,bai2021binarybert,dai-etal-2022-enabling} are also likely to be beneficial.




\section{Conclusion}
We have introduced NusaCrowd, a combined resource for Indonesian and regional languages, covering 137 datasets, 118 of which have a standardized loader. NusaCrowd covers Indonesian and 18  regional languages, encompassing 3 different data modalities. Manual and automatic curation processes were conducted to verify the quality of the collected datasets. The effectiveness of NusaCrowd is shown in three use cases: zero-shot NLU (NusaNLU), zero-shot NLG (NusaNLG), and multilingual ASR (NusaASR) benchmarks. Our experiments provide evidence regarding the efficiency of cross-tasks method over prompting for zero-shot NLU, the limited capabilities of existing large LMs for handling NLG tasks in local languages, and the potential of joint multilingual multi-task learning for Indonesian ASR. We hope NusaCrowd will benefit the research community as a data hub for Indonesian and regional languages by facilitating easy access to datasets, as well as accelerating research and development.

\section{Limitations}
\label{sec:limitation}

\paragraph{Dataset Utilization} We have collected 137 datasets, yet we have only conducted experiments over a minority of these ($\sim$40 datasets), leaving the remaining datasets unexplored. Since the datasets are already curated, future work should further explore these datasets in additional experiments. In this work, we do not experiment on image-text datasets for two reasons: (1) all of the image-text datasets are translated from English versions; and (2) there is no large LM available for zero-shot image-to-text generation.

\paragraph{Experiments} We did not attempt few-shot or fully-supervised learning experiments in NusaCrowd since prior work has explored these approaches on some of the datasets~\cite{wilie2020indonlu,koto2020indolem,cahyawijaya-etal-2021-indonlg,winata-etal-2023-nusax}. We specifically conduct our experiments on zero-shot methods to explore the generalization of zero-shot cross-lingual and zero-shot prompting approaches to extremely low-resource languages. 


\paragraph{Task Diversity} The tasks represented in NusaCrowd are skewed towards MT, sentiment, abusive text classification, and ASR. Many other tasks remain unexplored for Indonesian and regional languages. Furthermore, most ASR work come from the same authors or research groups. While these topics are prevalent among Indonesian researchers, it is also important to expand to other tasks.

\paragraph{Domain Diversity} The datasets in NusaCrowd are primarily from the domains of social media, news, and other general domain sources. Despite having a huge potential, narrow-domain datasets, such as clinical, biomedical, legal, financial, and educational datasets remain underrepresented for Indonesian and regional languages. Exploration of domain-specific data and use cases for Indonesian and regional languages is critical.

\paragraph{Language Diversity} There are 700+ languages in Indonesia. However, we have only focused on a small fraction of these languages. In addition, there are also other regional languages similar to the two Sinitic languages in NusaCrowd, i.e., Hakka (Khek) and Min Nan (Teochew). More focus on under-represented languages is an interesting future direction.

\paragraph{Multimodality} The datasets in NusaCrowd are mainly in the text modality. Exploration of speech, image, and other modalities for Indonesian and regional languages is still limited, and there are potentially exciting opportunities to capture locally-relevant Indonesian culture in such modalities.

\paragraph{Utilization of Datasets} There are 137 datasets contained in NusaCrowd. While we showcased three different use cases for the datasets (i.e., zero-shot NLU, zero-shot NLG, and multilingual ASR benchmarks), there is much greater potential to use the datasets in NusaCrowd. Potential areas of focus include experimenting with various approaches and analyses over multiple datasets, such as multi-task learning, continual learning, or few-shot learning.

\section{Ethical Statement}

Our work highlights the importance of democratizing access to Natural Language Processing (NLP) technology for underrepresented and extremely low-resource languages with a focus on the Austronesian language family specifically in Indonesian languages. Within our study, we are well aware of the ethical responsibility associated with language research and the potential impact that comes with it. Our study prioritizes diversity, inclusivity, and fairness. Within this work, the contribution of each 
collaborator is calculated following a fair and transparent scoring guideline that empowers the core principles of NusaCrowd. We have obtained informed consent from all dataset authors to provide publicly open-access corpora and benchmarks. Throughout our research process, we have made conscious efforts to engage with the language communities, involve local experts, and respect their linguistic and cultural nuances. We encourage further collaboration and engagement with underrepresented language communities to ensure that their voices are heard and their needs are addressed in future NLP development. We remain committed to the principles of ethical research, diversity, inclusivity, and fairness, striving to promote social good through our work in the field of language technology.




\bibliography{acl_latex}
\bibliographystyle{acl_natbib}

\clearpage

\newcommand{\appendixhead}{\centering\textbf{\huge Appendix}\vspace{0.5in}}
\twocolumn[\appendixhead]

\appendix

\setcounter{table}{0}
\renewcommand{\thetable}{A}
\setcounter{figure}{0}
\renewcommand{\thefigure}{A}

\section{Languages in NusaCrowd}
\label{app:language-nusacrowd}

\begin{figure}[!t]
    \centering
    \resizebox{\linewidth}{!}{
        \includegraphics{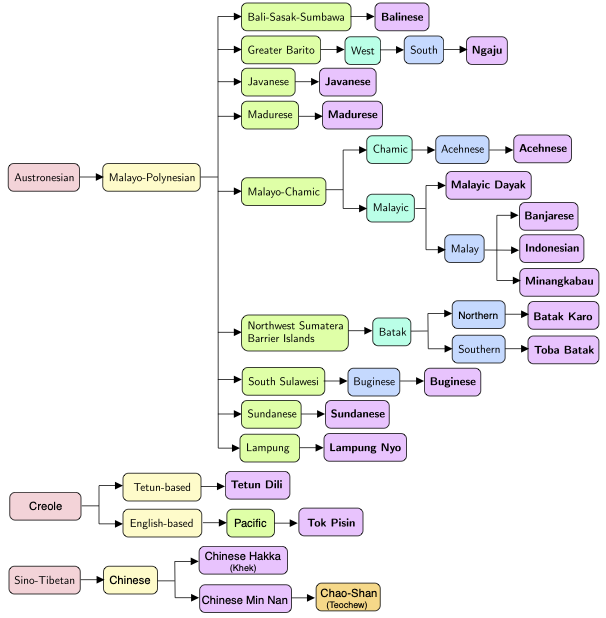}
    }
    \caption{Language family tree for all the languages covered in NusaCrowd. Most languages are Austronesian with two Creole languages and two other languages are under Sino-Tibetan language family.}
    \label{fig:language-nusacrowd}
\end{figure}

\Cref{tab:language-nusacrowd} provides the language code, name, and family
for all 19 languages listed in NusaCrowd. The language family information is collected from Ethnologue~\cite{ethnologue}. We follow the ISO 639-3 standard\footnote{\url{https://iso639-3.sil.org/}} for language coding in NusaCrowd. The language tree of all languages in NusaCrowd is shown in \Cref{fig:language-nusacrowd}.

\begin{table}[!t]
    \centering
    \resizebox{0.765\linewidth}{!}{
        \begin{tabular}{c|c|c}
            \toprule
            \textbf{Lang Code} & \textbf{Lang Name} & \textbf{Family} \\
            \midrule
            ace & Acehnese & MP\\
            abl & Lampung Nyo & MP\\
            ban & Balinese & MP\\
            bbc & Batak Toba & MP\\
            bjn & Banjar & MP \\
            btk & Batak & MP \\
            btx & Batak Karo & MP\\
            bug & Buginese & MP\\
            hak & Hakka/Khek & ST\\
            ind & Indonesian & MP \\
            jav & Javanese & MP\\
            mad & Madura & MP\\
            min & Minangkabau & MP\\
            nan & Min Nan (Teochew) & ST\\
            nij & Ngaju & MP\\
            sun & Sundanese & MP\\
            tpi & Tok Pisin & CR\\
            tdt & Tetun Dili & CR\\
            xdy & Malayic Dayak & MP\\
            \bottomrule
        \end{tabular}
    }
    \caption{Language codes and its complete names for all 19 languages listed in NusaCrowd. \textbf{MP} denotes Malayo-Polynesian, \textbf{CR} denotes Creole, and \textbf{ST} denotes Sino-Tibetan language family.}
    \label{tab:language-nusacrowd}
\end{table}

\textbf{Acehnese} (ace) is a language spoken mainly in the Aceh province.  Although it is the de facto language of provincial identity of Aceh, language use is shifting to Indonesian in urban areas. Acehnese has features typical of the Mon-Khmer languages of mainland Southeast Asia, a result of its former status as part of the early Chamic dialect continuum on the coast of Vietnam. It has at least ten contrasting vowels and as many distinct diphthongs, as well as voiceless aspirated stops and murmured voiced stops~\cite{blust2013austronesian}.
In addition to the large number of diphthongs, it has a high percentage of monosyllabic root morphemes. Prefixes and infixes play an active role while suffixes are absent~\cite{Durie:1985}. It is of the `active’ or so-called `Split-S’ type: some intransitive verbs take arguments, which have the properties of `transitive subjects’ while others take arguments with the properties of ‘transitive objects’~\cite{durie1988preferred}.

\textbf{Lampung Nyo} (abl) is a language spoken in three enclaves east between Kanan and Seputih rivers in Lampung province.
It is one of the three languages under the subgroup Lampung. The other two languages are Komering and Lampung Api. 
It has four dialects: Abung, Tulangbawang, Sukadana, and Melinting, with 77\% of lexical similarity among dialects.  
It was written in Kaganga script but it is written mainly in Latin script~\cite{ethnologue}.

\textbf{Balinese} (ban) is a language spoken mainly in the Bali province and in the West Nusa Tenggara province. It has three main dialects: Highland Balinese, Lowland Balinese, and Nusa Penida. It is mainly written in the Latin script since the early 20th century although it has its own Balinese script.
The word order in Balinese is SVO. It is non-tonal and has 17 consonant and 6 vowel phonemes. Stress is on the penultimate syllable. It has three sociolinguistic registers. 
Regarding patterns of verb affixation, Balinese is an `active' or `split-S' language: verbs with Undergoer-like subject arguments are marked in one way (with a `zero prefix'), while verbs with Actor-like subject arguments---intransitive or transitive---are marked in another (either with the nasal prefix `N-', or with `ma-')~\cite{arka2003balinese}.

\textbf{Toba Batak} (bbc) is a language spoken in the North Sumatra province. 
Similarly to Acehnese, it is slowly being replaced by Indonesian in urban and migrant areas. It used to be written in the Batak script but is mainly written in Latin script now.
The Batak languages are predicate-initial, and have verb systems reminiscent of Philippine languages, although they differ from them in many details~\cite{blust2013austronesian}.

\textbf{Banjarese} (bjn) is a language spoken in Kalimantan (Central, East, South, and West Kalimantan provinces). It became a language of wider communication through trade in the market, in business, and in media. It is dominant in the South Kalimantan Province and also growing rapidly in the Central and Eastern Kalimantan provinces. It has two main dialects: Kuala and Hulu dialects. Although it is a Malayic language, it has many Javanese loanwords, probably acquired during the Majapahit period from the late thirteenth century until the fifteenth century~\cite{blust2013austronesian}. It has 73\% of lexical similarity with Indonesian\footnote{i.e., 73\% of its words also occur in Indonesian.} and it is written in Arabic and Latin scripts.

\textbf{Batak languages} (btk) are a subgroup of the languages of Northwest Sumatra-Barrier Islands spoken by the Batak people in the North Sumatra province and surrounding areas. 
Batak languages can be divided into three groups: Northern, Simalungan, and Southern. The Northern group consists of three languages: Batak Alas-Kluet (btz), Batak Dairi (btd), and Batak Karo (btx). The Simalungan group has one language only, i.e., Batak Simalungun (bts). The Southern group consists of three languages: Batak Angkola (akb), Batak Mandailing (btm), and  Batak Toba (bbc)~\cite{ethnologue}.
The Batak languages were written using the Batak script, but the Latin script is now used for most writing.

\textbf{Batak Karo} (btx) is a language spoken in Aceh province and North Sumatra province. The language status is threatened. 
The lexical similarity is 81\% with Batak Dairi (btd), 80\% with Batak Simalungun (bts), and 76\% with Batak Alas-Kluet (btz)~\cite{woollams2005karo}.
It has 17 consonants and 7 vowels. The stress is on the penultimate syllable.
Similar to Indonesian, it has inclusive/exclusive pronouns.
The basic word order is SVO with prepositions. It is a head initial language, except for the order of quantifiers.
It has two voices: actor-voice and undergoer-voice.
It is written in Batak script and also Latin script.

\textbf{Buginese} (bug) is a language spoken mainly in the South Sulawesi, Southeast Sulawesi, Central Sulawesi, and West Sulawesi provinces. 
The word order is SVO. Verb affixes are used to mark persons. It is non-tonal and has 19 consonant and 6 vowel phonemes. Stress is on the penultimate syllable. It was written in the Buginese script in the past (derived from Brahmi script) but is mainly written in Latin script now~\cite{ethnologue}.
In Buginese, the pronoun `I' has three forms: the independent form `iyya', the ergative form `-ka', and the absolutive form/clitic `u-'. Buginese employs sentence patterns, pronouns, and certain terms to express politeness~\cite{weda2016syntactic}.

\textbf{Hakka} (hak) is a language spoken in Southeastern China, mainly in Guangdong province, also in Fujian, Guangxi, Hainan, Hunan, south Jiangxi, and Sichuan provinces. It is also spoken by Chinese descendants in some parts in Indonesia, such as in Singkawang in West Kalimantan province~\cite{stenberg2015multilingualism}, in Medan in North Sumatra province~\cite{nasution2020language}, and in Lhokseumawe in Aceh province~\cite{saleh2018chinese}.
It is a tonal language and the basic word order is SVO.
It is written in Han script and also Latin script.

\textbf{Indonesian} (ind) is the national language of Indonesia in 1945 Constitution, Article 36. 
Its lexical similarity to Standard Malay is over 80\%. 
The word order is SVO. 
It is non-tonal and has 19 consonants, 6 vowels, and 3 diphthongs. The stress is on the penultimate syllable. 
It has a rich affixation system, including a variety of prefixes, suffixes, circumfixes, and reduplication. Most of the affixes in Indonesian are derivational~\cite{pisceldo-etal-2008-two}.
It is developed from literary `Classical Malay’ of the Riau-Johor sultanate \cite{sneddon2003} and has regional variants. It is written in Latin script.

\textbf{Javanese} (jav) is a language spoken mainly in Java island.
It is the de facto language of provincial identity in central and eastern Java. 
The word order is SVO. It has 21 consonants and 8 vowels. 
It used to be written in Javanese script but since 20th century is mostly written in Latin script.
Javanese differs from most other languages of western Indonesia in contrasting dental and retroflex stops, and in the feature of breathy voice or murmur as a phonetic property of its voiced obstruents. Javanese also differs from most languages of the Philippines and western Indonesia in allowing a number of word-initial consonant clusters. It has an elaborate system of speech levels~\cite{blust2013austronesian}.

\textbf{Madurese} (mad) is a language spoken in the East Java province, mainly on Madura Island, south and west of Surabaya city, Bawean, Kangean, and Sapudi islands. 
It has vowel harmony, gemination, rich affixation, three types of reduplication, and SVO basic word order \cite{davies2010grammar}.

\textbf{Minangkabau} (min) is a language spoken mainly in West Sumatra and other provinces on Sumatra Island such as Bengkulu and Riau. Although it is classified as Malay, it is not intelligible with Indonesian. 
The word order is SVO written in Latin script. 
Standard Minangkabau voice can be characterised as an Indonesian-type system whereas colloquial Minangkabau voice is more effectively characterised as a Sundic-type system~\cite{crouch2009voice}.

\textbf{Min Nan} (nan) is a language spoken in Southeastern China. One of its dialects is Chaozhou-Shantou (Chao-Shan dialect) or Teochew dialect. It is spoken by Chinese descendants in some parts of Indonesia such as in Jambi~\cite{peng2011head} and in Pontianak in West Kalimantan province~\cite{veniranda2015perfective}. 
While Teochew is historically Chinese, its contact with languages in Indonesia has resulted in some changes uncharacteristic of Chinese languages.
For example, regarding word order, Teochew spoken in Jambi exhibits both head-final and head-initial relative clauses even though head-initial relative clauses are generally ungrammatical in Chinese languages. In addition to the head-initial word order, Jambi Teochew has also borrowed the Malay relativizer \textit{yang}~\cite{peng2011head}. 
It is a tonal language with tone sandhi. The word order is SVO~\cite{ethnologue}. 

\textbf{Ngaju} (nij) is a language spoken in the Central Kalimantan province. It is widely used as a language of wider communication for trade in much of Kalimantan, from the Barito to the Sampit river. It is used in many domains (church, school, village-level government, market, etc.). It has various affixes and reduplication, similar to Indonesian. The active voice is marked by prefix `maN-' and the passive voice is marked by prefix `iN-'. The word order is similar to the one in Indonesian. The pronouns have enclitic forms to mark possessors in a noun phrase or agents in a passive sentence~\cite{UchiboriShibata1988}.

\textbf{Sundanese} (sun) is a language spoken mainly in the Banten and West Java provinces. It is the de facto language of provincial identity in western Java. The main dialects are Bogor (Krawang), Pringan, and Cirebon. 
It is non-tonal and has 18 consonant and 7 vowel phonemes. The stress is on the penultimate syllable. It has elaborate coding of respect levels. It is written in Latin script since the middle of the 19th century but was previously written in Arabic, Javanese, and Sundanese scripts.
Sundanese is a predominantly SVO language. It has voice marking and incorporates some (optional) actor-verb agreement, i.e., number and person~\cite{kurniawan2013sundanese}.

\textbf{Tok Pisin} (tpi) is an English-based creole and de facto the national language of Papua New Guinea, a neighboring country of Indonesia. 
Dialect differences exist among lowlands, highlands, and islands. Highlands lexicon has more English influence.
It is a non-tonal language and has 16 consonant and 5 vowel phonemes. It has inclusive/exclusive pronouns and the basic word order is SVO.
It is written in Latin script~\cite{ethnologue}.

\textbf{Tetun Dili} (tdt) is a Tetun-based creole spoken in Dili district, East Timor north coast as the first language and scattered in western part of East Timor as the second language. 
It is a statutory national language according to the 2002 Constitution, Article 13. It has heavy Portuguese (por) and Mambae (mgm) influence as well as some Indonesian (ind) or Malay influence. 
It is a non-tonal language with 22 consonants and 5 vowels. The stress is most commonly on the penultimate syllable. It has inclusive/exclusive pronouns. 
The basic word order is SVO with prepositions and tense-aspect markers. It is a head-initial language, except for possessors.
The speakers of Tetun Dili also use Tetun [tet], some bilingually, but many others have significant difficulty understanding it in many domains.
It is written in Latin script~\cite{ethnologue}.

\textbf{Malayic Dayak} (xdy) is a language widely dispersed in Central and West Kalimantan provinces. It has many dialects and it is written in Latin script~\cite{ethnologue}. 
Malayic Dayak is not a proper subgroup, but refers to the large number of unclassified but clearly Malayic languages of Borneo which have a three voice system~\cite{sommerlot2020syntax}.

\section{Schemas in NusaCrowd}
\label{app:task-schema}

Schema serves to define and format the attributes of the dataset returned by a data loader. For each data loader, we implement a source schema, which is responsible to present the dataset in a format similar to its original structure, and a nusantara schema, which supports the standardization data structure across similar tasks.

We define the nusantara schemas as follows. Labels are in string format unless indicated otherwise.

\begin{compactitem}
    \item \textbf{Image-text (\texttt{IMTEXT})}. This schema could be used for image captioning, text-to-image generation, and vision-language pre-training. It consists of \texttt{(id, text, image\_paths, metadata)}, where \texttt{id} denotes a unique row identifier of the dataset, \texttt{text} denotes an input text, \texttt{image\_paths} denotes a list of paths to the input image sources, and \texttt{metadata} denotes relevant details such as visual concepts and labels (if required).

    \item \textbf{Speech-text (\texttt{SPTEXT})}. This could be used for speech recognition, text-to-speech (TTS) or speech synthesis, and speech-to-text translation. It consists of \texttt{(id, path, audio, text, speaker\_id, metadata)}, where \texttt{id} denotes a unique row identifier of the dataset, \texttt{path} denotes the file path to an input audio source, \texttt{audio} denotes the audio data loaded from the corresponding \texttt{path}, \texttt{text} denotes an input text, \texttt{speaker\_id} denotes a unique identifier of the speaker, \texttt{metadata} denotes relevant details such as the age and gender of the speaker (if required).

    \item \textbf{Speech-to-speech (\texttt{S2S})}. This could be used for speech-to-speech translation. It consists of \texttt{(id, path\_1, audio\_1, text\_1, metadata\_1, path\_2, audio\_2, text\_2, metadata\_2)}, where \texttt{id} denotes a unique row identifier of the dataset, \texttt{path\_1} and \texttt{path\_2} denote the file path to a respective input audio source, \texttt{audio\_1} and \texttt{audio\_2} denote the audio data loaded from the corresponding \texttt{path}, \texttt{text\_1} and \texttt{text\_2} denote input texts, and \texttt{metadata\_1} and \texttt{metadata\_2} denote relevant details such as the age of the speaker and their gender (if required).

    \item \textbf{Unlabeled text (\texttt{SSP})}. This schema could be used for language modeling in self-supervised pre-training. It consists of \texttt{(id, text)}, where \texttt{id} denotes a unique row identifier of the dataset and \texttt{text} denotes an input text.
    
    \item \textbf{Single-label text classification (\texttt{TEXT})}. This schema could be used for sentiment analysis, emotion classification, legal classification, and others. It consists of \texttt{(id, text, label)}, where \texttt{id} denotes a unique row identifier of the dataset, \texttt{text} denotes an input text, and \texttt{label} denotes a deterministic target variable.

    \item \textbf{Multi-label text classification (\texttt{TEXT MULTI})}. This schema could be used for hate speech detection and aspect-based sentiment analysis. It consists of \texttt{(id, text, labels)}, where \texttt{id} denotes a unique row identifier of the dataset, \texttt{text} denotes an input text, and \texttt{labels} denotes a list of deterministic target variables.
    
    \item \textbf{Text-to-text (\texttt{T2T})}. This schema could be used for machine translation, summarization, and paraphrasing. It consists of \texttt{(id, text\_1, text\_2, text\_1\_name, text\_2\_name)}, where \texttt{id} denotes a unique row identifier of the dataset, \texttt{text\_1} and \texttt{text\_2} denote an input text pair, and \texttt{text\_1\_name} and \texttt{text\_2\_name} denote the names of the input text pair (e.g., \texttt{ind} and \texttt{jav} for translation input text pairs, or \texttt{document} and \texttt{summary} for summarization input text pairs).

    \item \textbf{Sequence labeling (\texttt{SEQ LABEL})}. This schema could be used for named entity recognition (NER), POS tagging, and others. It consists of \texttt{(id, tokens, labels)}, where \texttt{id} denotes a unique row identifier of the dataset, \texttt{tokens} denotes a list of tokens of an input text, and \texttt{labels} denotes a list of targets for the tokens.

    \item \textbf{Question answering (\texttt{QA})}. This schema could be used for extractive QA, multiple-choice QA, and others. It consists of \texttt{(id, question\_id, document\_id, question, type, choices, context, answer)}, where \texttt{id} denotes a unique row identifier of the dataset, \texttt{question\_id} denotes a unique identifier of the question, \texttt{document\_id} denotes a unique identifier of the context document, \texttt{question} denotes an input question to be answered, \texttt{type} denotes the type of the QA task (e.g., extractive, multiple-choice, open-generative, closed-generative, etc.), \texttt{choices} denotes a list of answer choices (if required), \texttt{context} denotes a passage that serves as the background information of the question (if required), and \texttt{answer} denotes the gold answer to the question (if required).

    \item \textbf{Single-label text pair classification \texttt{(PAIRS)}}. This could be used for textual entailment and next sentence prediction. It consists of \texttt{(id, text\_1, text\_2, label)}, where \texttt{id} denotes a unique row identifier of the dataset, \texttt{text\_1} and \texttt{text\_2} denote an input text pair, and \texttt{label} denotes the target variable.

    \item \textbf{Single-label text pair classification with continuous values or regression \texttt{(PAIRS SCORE)}}. This could be used for answer grading and semantic textual similarity. It consists of \texttt{(id, text\_1, text\_2, label)}, where \texttt{id} denotes a unique row identifier of the dataset, \texttt{text\_1} and \texttt{text\_2} denote an input text pair, and \texttt{label} denotes a target variable as a continuous value.

    \item \textbf{Multi-label text pair classification (\texttt{PAIRS MULTI})}. This could be used for morphological inflection. It consists of \texttt{(id, text\_1, text\_2, labels)}, where \texttt{id} denotes a unique row identifier of the dataset, \texttt{text\_1} and \texttt{text\_2} denote an input text pair, and \texttt{labels} denotes a list of target variables.

    \item \textbf{Knowledge base (\texttt{KB})}. This schema could be used for constituency parsing, dependency parsing, coreference resolution, dialogue system, and other tasks with complex structures. It consists of \texttt{(id, passages, entities, events, coreferences, relations)}. Considering its intricate structure, we encourage readers to take a look at the implementation of the knowledge base schema.
\end{compactitem}



\section{Details for Zero-Shot Setting Experiment in NusaNLU}
\label{app:exp-zeroshot-nlu}

\paragraph{Model Checkpoints}
For the NLU experiment, we utilize 4 model checkpoints, which are: 1) BLOOMZ fine-tuned on English prompt with 3B parameters\footnote{\url{https://huggingface.co/bigscience/bloomz}}, 2) XGLM with 2.9B parameters\footnote{\url{https://huggingface.co/facebook/xglm-2.9B}}, 3) off-the-shelf XLM-R fine-tuned on XNLI\footnote{\url{https://huggingface.co/joeddav/xlm-roberta-large-xnli}}, and 4) XLM-R large fine-tuned on IndoNLI. For XLM-R large fine-tuned on IndoNLI, we fine-tuned the XLM-R large model with batch size of 128 and initial learning rate of 1e-5 for 50 epochs. We use AdamW optimizer with a linear learning rate decay and apply early stopping of 5 epochs based on the validation accuracy score.

\paragraph{Prompts}
We run the prompting experiment using 3 different prompts for each task type. We cover several different task types in our NLG experiments, i.e., sentiment analysis, abusive detection, hate speech detection, emotion classification, natural language inference (NLI), and next tweet prediction. The prompt templates used for each task type are shown in \Cref{tab:senti-prompt,tab:nli-prompt}.

\section{Details for Zero-Shot Setting Experiment in NusaNLG}
\label{app:exp-zeroshot-nlg}

\paragraph{Model Checkpoints}
For the NLG experiment, we utilize 2 model checkpoints, i.e., BLOOMZ fine-tuned on English prompt with 3B parameters and XGLM with 2.9B parameters. We use the same checkpoint as the one used in the zero-shot NLU experiment.

\paragraph{Generation Hyperparameters}
For generating the prediction sequence, we generate sequence using greedy decoding with sampling, using top-k of 50 and top-p of 1.0. We force the model to at least generate one token and limit the generation sequence length to 100 tokens.

\paragraph{Prompts}
We run the prompting experiment using 3 different prompts for each task type. We cover two different task types in our NLG experiments, i.e., machine translation and summarization. The prompt templates used in our NLG experiment are shown in \Cref{tab:nlg-summary-prompt} and \Cref{tab:nlg-translation-prompt}.

\section{Details of Speech Recognition Experiment in NusaASR}
\label{app:exp-multilingual-asr}

\paragraph{Model Checkpoints} 
For both the monolingual and multilingual ASR experiment, we employ 2 wav2vec 2.0$_{LARGE}$ model checkpoints (both with $\sim$300M parameters) as follows: 1) pre-trained XLSR wav2vec 2.0 model\footnote{\url{wav2vec2-large-xlsr-53: https://huggingface.co/facebook/wav2vec2-large-xlsr-53}} and an off-the-shelf fine-tuned XLSR wav2vec 2.0 model to Indoensian, Sundanese, and Javanese speech data\footnote{\url{https://huggingface.co/indonesian-nlp/wav2vec2-indonesian-javanese-sundanese}}. For Whisper model we employ the Whisper$_{SMALL}$\footnote{\url{https://huggingface.co/openai/whisper-small}} model with 244M parameters. For the monolingual experiment, we explore training using the 3 largest and widely-used languages in Indonesia, i.e., Indonesian (ind), Javanese (jav), and Sundanese (sun).

\paragraph{Fine-Tuning Hyperparameters}
We apply fine-tuning to both XLSR wav2vec 2.0 and Whisper models for single-task training, monolingual multi-task training, and multilingual multi-task training settings. We fine-tune the models using the following hyperparameters, i.e., Adam optimizer with a learning rate of 5e-5 for the wav2vec 2.0 model and 1e-4 for the Whisper model, training batch size of 16, fine-tuning epoch of 30, and apply an early stopping of 5 epoch based on the validation word error rate (WER). For each model, we search for the best learning rate ranging from [5e-4 \dots 1e-5]. We run all experiments on a single A100 GPU.

\section{Zero-Shot Results of NusaNLU}
\label{app:zeroshot-nlu}
\begin{figure*}[!ht]
\centering
\includegraphics[width=\linewidth]{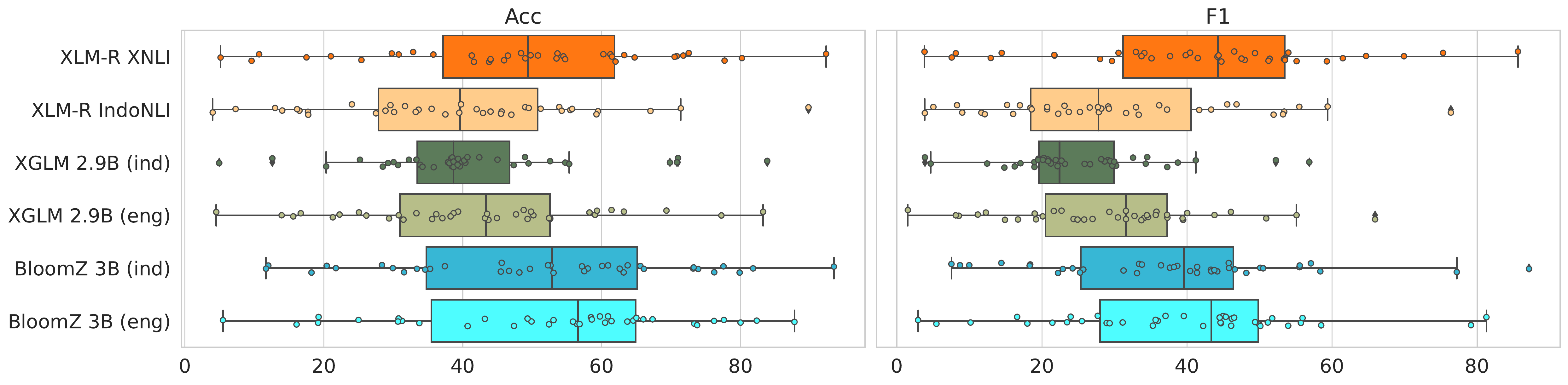}
\caption{Zero-shot generalization to NLU tasks in NusaNLU. Box plots show summary statistics on accuracy \textbf{(left)} and \textbf{F1} (right). Points are per-dataset scores from the average of performances using three different prompts. }
\label{fig:zeroshotnlu-full}
\end{figure*}

Here we elaborate further on the analysis in \Cref{sec:nusanlu}. We report  the overall performances of each model in \Cref{fig:zeroshotnlu-full} and per task performance in \Cref{tab:appendix_zeroshotnlu}. Predictions derived by prompting BLOOMZ outperform all the other models and perform on average on par with zero-shot cross-task prompting using the XLM-R model trained on XNLI. In detail, predictions using cross-task prompting actually are better in F1 than using BLOOMZ in 17 tasks, while it's actually worse in accuracy in 13 tasks, all out of the 26 NLU tasks sampled. One extreme example can be observed in their performance comparison on the id\_abusive task, where predicting by cross-task prompting XLM-R trained on XNLI nearly triples the F1 on prompting BLOOMZ. These results suggest that methods like cross-task prompting are worth exploring, benefitting better efficiency through cross-task transfer on low-resource language tasks compared to large multilingual LMs.

Comparing the languages of the prompt, although on both XGLM and BLOOMZ it's better to use the English prompt, the difference is actually more apparent on average when prompting is done using XGLM. However, when we zoom into each of the tasks, the difference is much larger in prompting using BLOOMZ. The largest spread is observed on utilizing the English prompt when predicting for the indolem sentiment analysis task, where the accuracy differs by $\sim$30\%, and the F1 differs by $\sim$37.8\%. Comparing the same variables in XGLM, the largest accuracy difference of $\sim$24\% is observed on id\_google\_play\_review\_posneg, and the largest F1 difference of $\sim$19.1\% is observed on Madurese (mad) sentiment analysis task. Furthermore, utilizing Indonesian prompts is not always the case, worse. On Buginese (bug) sentiment analysis utilizing BLOOMZ we can get $\sim$23\% more accuracy by using Indonesian prompt. On classifying emotion in emotcmt task utilizing XGLM, we can get $\sim$7\% more F1 by using also the Indonesian prompt. On the indolem next-tweet-prediction task, utilizing both BLOOMZ and XGLM using also the Indonesian prompt, we can get additional $\sim$14\% accuracy and $\sim$23\% F1 respectively.

\section{Zero-Shot Results of NusaNLG}
\label{app:zeroshot-nlg}
\begin{figure*}[!ht]
\centering
	\includegraphics[width=\linewidth]{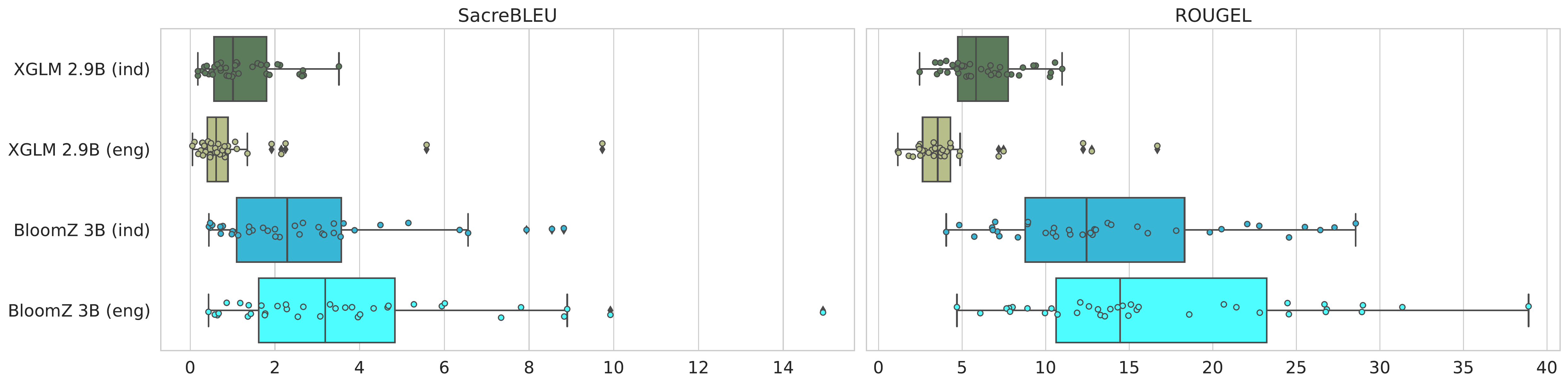}
	\caption{Zero-shot generalization to NLG tasks in NusaNLG. Box plots show summary statistics on SacreBLEU \textbf{(left)} and ROUGE-L \textbf{(right)}. Points are per-dataset scores from the average of performances using 3 different prompts.}
	\label{fig:zeroshotnlg-full}
\end{figure*}

Here we elaborate further on the analysis in \Cref{sec:nusanlg}. We report  the overall performances of each model in \Cref{fig:zeroshotnlg-full} and per task performance in \Cref{tab:appendix_zeroshotnlg}. Generations derived by prompting BLOOMZ are better than prompting XGLM in all of the tasks except in indosum\_fold0\_nusantara\_t2t, where the scores differ slightly. The performances in the summarization tasks are generally lower than the performances in the machine translation tasks. On the machine translation tasks, the performance in translating to the Indonesian language as the target language is generally higher than translating to the local languages, while translating from English to Indonesian is generally performing the highest.

Prompting using BLOOMZ yields better performances in most of the tasks, when prompting using English prompts than using Indonesian prompts. In general, prompting using XGLM yields better generation using Indonesian prompts than using English prompts. This is especially the case in the machine translation tasks, where most of them yield better performances except when translating to Toba Batak (bbc) and Banjarese (bjn) from Indonesian (ind), and also when translating to Minangkabau (min) to Indonesia (ind) and vice versa. In the summarization task, prompting using XGLM with English prompts produce better results than with Indonesian prompts.

It's worth noting that the translation quality is extremely poor for local languages, especially in Banjarese (bjn), Acehnese (ace), Toba Batak (bbc), Ngaju (nij), Madurese (mad), and Sundanese (sun). This is even more severe when those local languages become the target languages. This finding suggests that both BLOOMZ and XGLM still fail to learn the representation of these local languages.

\section{ASR Results of NusaASR}
\label{app:nusaasr-results}

Here we elaborate further on the analysis in \Cref{sec:nusaasr}. We report  the per-task performance of each model in \Cref{tab:appendix_multilingualasr_wav2vec_pt} for the wav2vec 2.0-pt model, \Cref{tab:appendix_multilingualasr_wav2vec_ft} for the wav2vec 2.0 model-ft, and \Cref{tab:appendix_multilingualasr_whisper} for the Whisper model. The best overall performance is achieved by \textbf{wav2vec 2.0-pt} fine-tuned in multilingual multi-task setting, achieving 17.03\% average WER over all tasks. The model also performs better in most cases for languages other than Indonesian compared to the  \textbf{wav2vec 2.0-ft} model. While for Indonesian, \textbf{wav2vec 2.0-ft} fine-tuned in all 3 training settings, i.e., multilingual multi-task, monolingual (ind) multi-task, and single-task settings, achieve much better scores, i.e., $<$5\% WER over all Indonesian tasks.

Comparing the performance per language, the best Indonesian ASR model achieves very low WER on Indonesian (ind) speech corpora, i.e., indspeech\_digit\_cdsr, indspeech\_news\_lvcsr, indspeech\_teldialog\_lvcsr, and indspeech\_teldialog\_svcsr. Compared to local languages, i.e., Minangkabau (min), Sundanese (sun), Javanese (jav), Balinese (ban), Acehnese (ace), Batak (btk), and Buginese (bug), the performance of the best ASR model only achieves $\sim$10-30\% WER. The performance is especially low for Buginese (bug), which suggests distinct speech features are required for handling speech recognition in Buginese (bug) language. This fact aligns with the result of prior work in Indonesian local languages~\cite{winata-etal-2023-nusax}, where Buginese (bug) has inferior performance in the leave-one-language-out setting.

\section{Private Datasets in NusaCrowd}
\label{app:private-dataset}

NusaCrowd offers access to 14 previously private datasets. We provide the details of all 14 previously private datasets listed in NusaCrowd along with the task, languages, and modality in \Cref{tab:private-dataset}.

\section{Comparison with Other Initiatives}
\label{app:initiative-comparison}

To provide a broader perspective of the impact of NusaCrowd, we provide the comparison of NusaCrowd initiatives with other global, regional, and Indonesian data initiatives in \Cref{tab:data-comparison}.



\section{Details of Datasets in NusaCrowd}
\label{app:data-description}

\Cref{tab:stats-nusacrowd} provides the details description, license, languages, dataset volume, annotation quality, and other metadata of all 137 datasets collected in NusaCrowd.

\newpage

\begin{table*}[!t]
    \centering
    \begin{tabular}{cp{0.85\linewidth}}
        \toprule
        \textbf{Language} & \textbf{Prompt in \colorbox{yellow!25}{Sentiment Analysis} Task}  \\
        \midrule
        \multirow{6}{*}{\textbf{Indonesian (ind)}} & \texttt{[INPUT]\textbackslash nApakah sentimen dari teks tersebut? [LABELS\_CHOICE]} \\
        \cmidrule{2-2}
        & \texttt{Apakah sentimen dari teks berikut?\textbackslash nTeks: [INPUT]\textbackslash nSentimen: [LABELS\_CHOICE]} \\
        \cmidrule{2-2}
        & \texttt{Teks: [INPUT]\textbackslash n\textbackslash nTolong prediksikan sentimen dari teks diatas: [LABELS\_CHOICE]} \\
        \cmidrule{1-2}
        \multirow{6}{*}{\textbf{English (eng)}} & \texttt{[INPUT]\textbackslash nWhat would be the sentiment of the text above? [LABELS\_CHOICE]} \\
        \cmidrule{2-2}
        & \texttt{What is the sentiment of this text?\textbackslash nText: [INPUT]\textbackslash nSentiment: [LABELS\_CHOICE]} \\
        \cmidrule{2-2}
        & \texttt{Text: [INPUT]\textbackslash n\textbackslash nPlease classify the sentiment of above text: [LABELS\_CHOICE]} \\
        \bottomrule
    \end{tabular}
    \caption{Prompt used for Sentiment Analysis task}
    \label{tab:senti-prompt}
\end{table*}

\begin{table*}[!t]
    \centering
    \begin{tabular}{cp{0.85\linewidth}}
        \toprule
        \textbf{Language} & \textbf{Prompt in \colorbox{yellow!25}{Emotion Classification} Task}  \\
        \midrule
        \multirow{5}{*}{\textbf{Indonesian (ind)}} & \texttt{[INPUT]\textbackslash nApakah emosi dari teks diatas? [LABELS\_CHOICE]} \\
        \cmidrule{2-2}
        & \texttt{Apakah emosi dari teks ini?\textbackslash n Teks: [INPUT]\textbackslash n Emosi: [LABELS\_CHOICE]} \\
        \cmidrule{2-2}
        & \texttt{Teks: [INPUT]\textbackslash n\textbackslash nTolong prediksikan emosi dari teks diatas: [LABELS\_CHOICE]} \\
        \cmidrule{1-2}
        \multirow{6}{*}{\textbf{English (eng)}} & \texttt{[INPUT]\textbackslash nWhat would be the emotion of the text above? [LABELS\_CHOICE]} \\
        \cmidrule{2-2}
        & \texttt{What is the emotion of this text?\textbackslash nText: [INPUT]\textbackslash nEmotion: [LABELS\_CHOICE]} \\
        \cmidrule{2-2}
        & \texttt{Text: [INPUT]\textbackslash n\textbackslash nPlease classify the emotion of above text: [LABELS\_CHOICE]} \\
        \bottomrule
    \end{tabular}
    \caption{Prompt used for Emotion Classification task}
    \label{tab:emot-prompt}
\end{table*}

\begin{table*}[!t]
    \centering
    \begin{tabular}{cp{0.85\linewidth}}
        \toprule
        \textbf{Language} & \textbf{Prompt in \colorbox{yellow!25}{Abusive Detection} Task}  \\
        \midrule
        \multirow{5}{*}{\textbf{Indonesian (ind)}} & \texttt{[INPUT]\textbackslash nApakah teks diatas kasar? [LABELS\_CHOICE]} \\
        \cmidrule{2-2}
        & \texttt{Apakah teks berikut ini kasar?\textbackslash n[INPUT]\textbackslash nJawab dengan [OPTIONS]: [LABELS\_CHOICE]} \\
        \cmidrule{2-2}
        & \texttt{[INPUT]\textbackslash nApakah menurutmu teks diatas itu [OPTIONS]? [LABELS\_CHOICE]} \\
        \cmidrule{1-2}
        \multirow{5}{*}{\textbf{English (eng)}} & \texttt{[INPUT]\textbackslash nIs the text abusive? [LABELS\_CHOICE]} \\
        \cmidrule{2-2}
        & \texttt{Is the following text abusive?\textbackslash n[INPUT]\textbackslash nAnswer with [OPTIONS]: [LABELS\_CHOICE]} \\
        \cmidrule{2-2}
        & \texttt{[INPUT]\textbackslash nDo you think the text is [OPTIONS]? [LABELS\_CHOICE]} \\
        \bottomrule
    \end{tabular}
    \caption{Prompt used for Abusive Detection task}
    \label{tab:abuse-prompt}
\end{table*}

\begin{table*}[!t]
    \centering
    \begin{tabular}{cp{0.85\linewidth}}
        \toprule
        \textbf{Language} & \textbf{Prompt in \colorbox{yellow!25}{Clickbait Detection} Task}  \\
        \midrule
        \multirow{5}{*}{\textbf{Indonesian (ind)}} & \texttt{[INPUT]\textbackslash nApakah judul diatas clickbait? [LABELS\_CHOICE]} \\
        \cmidrule{2-2}
        & \texttt{Apakah judul berikut ini clickbait?\textbackslash n[INPUT]\textbackslash nJawab dengan [OPTIONS]: [LABELS\_CHOICE]} \\
        \cmidrule{2-2}
        & \texttt{[INPUT]\textbackslash nApakah menurutmu teks diatas itu [OPTIONS]? [LABELS\_CHOICE]} \\
        \cmidrule{1-2}
        \multirow{5}{*}{\textbf{English (eng)}} & \texttt{[INPUT]\textbackslash nIs the title clickbait? [LABELS\_CHOICE]} \\
        \cmidrule{2-2}
        & \texttt{Is the following title a clickbait?\textbackslash n[INPUT]\textbackslash nAnswer with [OPTIONS]: [LABELS\_CHOICE]} \\
        \cmidrule{2-2}
        & \texttt{[INPUT]\textbackslash nDo you think the text is [OPTIONS]? [LABELS\_CHOICE]} \\
        \bottomrule
    \end{tabular}
    \caption{Prompt used for Clickbait Detection task}
    \label{tab:cbd-prompt}
\end{table*}

\begin{table*}[!t]
    \centering
    \begin{tabular}{cp{0.85\linewidth}}
        \toprule
        \textbf{Language} & \textbf{Prompt in \colorbox{yellow!25}{Rating Review Regression} Task}  \\
        \midrule
        \multirow{6}{*}{\textbf{Indonesian (ind)}} & \texttt{[INPUT]\textbackslash nBerapa rating dari teks review tersebut, dari 1 sampai 5? [LABELS\_CHOICE]} \\
        \cmidrule{2-2}
        & \texttt{[INPUT]\textbackslash nDari 1 sampai 5, berapa rating dari review diatas? [LABELS\_CHOICE]} \\
        \cmidrule{2-2}
        & \texttt{[INPUT]\textbackslash nDari 1 sampai 5 bintang, bagaimana menurutmu rating dari review tersebut? [LABELS\_CHOICE]} \\
        \cmidrule{1-2}
        \multirow{6}{*}{\textbf{English (eng)}} & \texttt{[INPUT]\textbackslash nWhat is the rating of the review above, from 1 to 5? [LABELS\_CHOICE]} \\
        \cmidrule{2-2}
        & \texttt{[INPUT]\textbackslash nFrom 1 to 5, what is the rating of the review above? [LABELS\_CHOICE]} \\
        \cmidrule{2-2}
        & \texttt{[INPUT]\textbackslash nFrom 1 to 5 stars, how would you rate the previous review? [LABELS\_CHOICE]} \\
        \bottomrule
    \end{tabular}
    \caption{Prompt used for Rating Review Regression task}
    \label{tab:rr-prompt}
\end{table*}

\begin{table*}[!t]
    \centering
    \begin{tabular}{cp{0.85\linewidth}}
        \toprule
        \textbf{Language} & \textbf{Prompt in \colorbox{yellow!25}{Hate Speech Detection} Task} \\
        \midrule
        \multirow{5}{*}{\textbf{Indonesian (ind)}} & \texttt{[INPUT]\textbackslash nApakah teks diatas hatespeech? [LABELS\_CHOICE]} \\
        \cmidrule{2-2}
        & \texttt{Apakah teks berikut ini hatespeech\textbackslash n[INPUT]\textbackslash nJawab dengan [OPTIONS]: [LABELS\_CHOICE]} \\
        \cmidrule{2-2}
        & \texttt{[INPUT]\textbackslash nApakah menurutmu teks diatas itu [OPTIONS]? [LABELS\_CHOICE]} \\
        \cmidrule{1-2}
        \multirow{6}{*}{\textbf{English (eng)}} & \texttt{[INPUT]\textbackslash nDo you think the text is hatespeech? Answer: [LABELS\_CHOICE]} \\
        \cmidrule{2-2}
        & \texttt{Is the following text a hatespeech?\textbackslash n[INPUT]\textbackslash nAnswer with [OPTIONS]: [LABELS\_CHOICE]} \\
        \cmidrule{2-2}
        & \texttt{[INPUT]\textbackslash nDo you think the text is [OPTIONS]? [LABELS\_CHOICE]} \\
        \bottomrule
    \end{tabular}
    \caption{Prompt used for Hate Speech Detection task}
    \label{tab:hsd-prompt}
\end{table*}

\begin{table*}[!t]
    \centering
    \begin{tabular}{cp{0.85\linewidth}}
        \toprule
        \textbf{Language} & \textbf{Prompt in \colorbox{yellow!25}{Next Tweet Prediction} Task}  \\
        \midrule
        \multirow{6}{*}{\textbf{Indonesian (ind)}} & \texttt{Diberikan dua tweet\textbackslash nA: [INPUT\_A]\textbackslash nB: [INPUT\_B]\textbackslash n\textbackslash nApakah tweet B adalah sambungan dari tweet A? [LABELS\_CHOICE]} \\
        \cmidrule{2-2}
        & \texttt{Apakah tweet "[INPUT\_B]" adalah sambungan dari tweet "[INPUT\_A]"? [LABELS\_CHOICE]} \\
        \cmidrule{2-2}
        & \texttt{Tweet pertama: [INPUT\_A].\textbackslash nApakah "[INPUT\_B]" merupakan sambungan dari tweet pertama? [LABELS\_CHOICE]} \\
        \cmidrule{1-2}
       \multirow{6}{*}{\textbf{English (eng)}} & \texttt{Given two tweets\textbackslash nA: [INPUT\_A]\textbackslash nB: [INPUT\_B]\textbackslash n\textbackslash nIs tweet B is a continuation of tweet A? [LABELS\_CHOICE]} \\
        \cmidrule{2-2}
        & \texttt{Is tweet "[INPUT\_B]" a continuation of tweet "[INPUT\_A]"? [LABELS\_CHOICE]} \\
        \cmidrule{2-2}
        & \texttt{First Tweet: [INPUT\_A].\textbackslash nWould "[INPUT\_B]" a continuation of the first tweet? [LABELS\_CHOICE]} \\
        \bottomrule
    \end{tabular}
    \caption{Prompt used for Next Tweet Prediction task}
    \label{tab:ntp-prompt}
\end{table*}

\begin{table*}[!t]
    \centering
    \begin{tabular}{cp{0.85\linewidth}}
        \toprule
        \textbf{Language} & \textbf{Prompt in \colorbox{yellow!25}{NLI} Task} \\
        \midrule
        \multirow{6}{*}{\textbf{Indonesian (ind)}} & \texttt{[INPUT\_A]\textbackslash nBerdasarkan kutipan sebelumnya, apakah benar bahwa "[INPUT\_B]"? [OPTIONS]? [LABELS\_CHOICE]} \\
        \cmidrule{2-2}
        & \texttt{[INPUT\_A]\textbackslash n\textbackslash nPertanyaan: Apakah kalimat tersebut mengimplikasikan bahwa "[INPUT\_B]"? [OPTIONS]? [LABELS\_CHOICE]} \\
        \cmidrule{2-2}
        & \texttt{Diberikan [INPUT\_A]. Apakah kalimat tersebut sesuai dengan [INPUT\_B]? [OPTIONS]? [LABELS\_CHOICE]} \\
        \cmidrule{1-2}
        \multirow{6}{*}{\textbf{English (eng)}} & \texttt{[INPUT\_A]\textbackslash nBased on the previous passage, is it true that "[INPUT\_B]"? Yes, no, or maybe? [LABELS\_CHOICE]} \\
        \cmidrule{2-2}
        & \texttt{[INPUT\_A]\textbackslash n\textbackslash nQuestion: Does this imply that "[INPUT\_B]"? Yes, no, or maybe? [LABELS\_CHOICE]} \\
        \cmidrule{2-2}
        & \texttt{Given that [INPUT\_A]. Does it follow that [INPUT\_B]? Yes, no, or maybe? [LABELS\_CHOICE]} \\
        \bottomrule
    \end{tabular}
    \caption{Prompt used for Natural Language Inference task}
    \label{tab:nli-prompt}
\end{table*}


\begin{table*}[]
    \centering
    \begin{tabular}{cp{0.85\linewidth}}
        \toprule
        \textbf{Language} & \textbf{Prompt in \colorbox{yellow!25}{Summary} Task}  \\
        \midrule
        \multirow{6}{*}{\textbf{Indonesian (ind)}} & \texttt{[INPUT]\textbackslash n===\textbackslash nTulis rangkuman dari teks diatas dalam bahasa Indonesia:} \\
        \cmidrule{2-2}
        & \texttt{Artikel dalam bahasa Indonesia: [INPUT]\textbackslash nRangkuman dalam bahasa Indonesia:} \\
        \cmidrule{2-2}
        & \texttt{[SOURCE]\textbackslash nBagaimana kamu merangkum teks diatas dalam bahasa Indonesia?} \\
        \midrule
        \multirow{5}{*}{\textbf{English (eng)}} & \texttt{[INPUT]\textbackslash n===\textbackslash nWrite a summary of the text above in Indonesian:} \\
        \cmidrule{2-2}
         & \texttt{Article in Indonesian: [INPUT]\textbackslash nSummary in Indonesian:} \\
        \cmidrule{2-2}
         & \texttt{[SOURCE]\textbackslash nHow would you rephrase that briefly in Indonesian?} \\
        \bottomrule
    \end{tabular}
    \caption{Prompt used for Summary task}
    \label{tab:nlg-summary-prompt}
\end{table*}

\begin{table*}[]
    \centering
    \begin{tabular}{cp{0.85\linewidth}}
        \toprule
        \textbf{Language} & \textbf{Prompt in \colorbox{yellow!25}{Translation} Task}  \\
        \midrule
        \multirow{6}{*}{\textbf{Indonesian (ind)}} & \texttt{Terjemahkan teks berikut dari bahasa [SOURCE] ke bahasa [TARGET].\textbackslash nTeks: [INPUT]\textbackslash nTerjemahan:} \\
        \cmidrule{2-2}
        & \texttt{[INPUT]\textbackslash nTerjemahkan teks diatas dari bahasa [SOURCE] ke bahasa [TARGET].} \\
        \cmidrule{2-2}
        & \texttt{Teks dalah bahasa [SOURCE]: [INPUT]\textbackslash nBagaimana kamu menterjemahkan teks diatas dalam bahasa [TARGET]?} \\
        \midrule
        \multirow{6}{*}{\textbf{English (eng)}} & \texttt{Translate the following text from [SOURCE] to [TARGET].\textbackslash nText: [INPUT]\textbackslash nTranslation:} \\
        \cmidrule{2-2}
         & \texttt{[INPUT]\textbackslash nTranslate the text above from [SOURCE] to [TARGET].} \\
        \cmidrule{2-2}
         & \texttt{Text in [SOURCE]: [INPUT]\textbackslash nHow would you translate that in [TARGET]?} \\
        \bottomrule
    \end{tabular}
    \caption{Prompt used for Translation task}
    \label{tab:nlg-translation-prompt}
\end{table*}

\clearpage

\begin{table*}[!ht]
\centering
\resizebox{\linewidth}{!}{%
\begin{tabular}{l|c|c|cc|cc|cc|cc|cc|cc}
\toprule
\multirow{2}{*}{\textbf{Dataset Name}} & \multirow{2}{*}{\textbf{Lang}} & \multirow{2}{*}{\textbf{Task}} & \multicolumn{2}{|c|}{\textbf{XLM-R XNLI}} & \multicolumn{2}{|c|}{\textbf{XLM-R IndoNLI}} & \multicolumn{2}{|c|}{\textbf{XGLM 2.9B (id)}} & \multicolumn{2}{|c|}{\textbf{XGLM 2.9B (en)}} & \multicolumn{2}{|c|}{\textbf{BLOOMZ 3B (id)}} & \multicolumn{2}{|c}{\textbf{BLOOMZ 3B (en)}} \\  \cmidrule(lr){4-15}
& & & \textbf{acc} & \textbf{f1} & \textbf{acc} & \textbf{f1} & \textbf{acc} & \textbf{f1} & \textbf{acc} & \textbf{f1} & \textbf{acc} & \textbf{f1} & \textbf{acc} & \textbf{f1} \\ \midrule
code\_mixed\_jv\_id\_id & ind & SA & 29.79 & 29.66 & 16.48 & 15.20 & 71.00 & 29.10 & 15.59 & 14.90 & 21.73 & 22.88 & 19.17 & 23.96 \\
code\_mixed\_jv\_id\_jv & jav & SA & 32.86 & 34.12 & 17.71 & 18.42 & 70.83 & 28.65 & 13.92 & 12.27 & 18.22 & 18.38 & 16.07 & 16.58 \\
emot & ind & EMOT & 49.77 & 47.95 & 33.64 & 29.14 & 30.68 & 20.82 & 29.39 & 19.18 & 45.61 & 38.66 & 43.18 & 35.29 \\
emotcmt & ind & EMOT & 43.81 & 41.48 & 28.87 & 25.24 & 32.25 & 25.88 & 26.12 & 19.00 & 45.48 & 33.38 & 56.41 & 44.68 \\
emotion\_id\_opinion & ind & EMOT & 50.83 & 49.37 & 12.98 & 11.65 & 29.25 & 21.90 & 31.46 & 30.48 & 48.15 & 45.79 & 49.90 & 49.98 \\
id\_abusive & ind & AD & 49.01 & 35.11 & 42.91 & 32.97 & 40.38 & 19.72 & 77.24 & 34.08 & 77.54 & 33.11 & 55.87 & 28.92 \\
id\_abusive\_news\_comment & ind & AD & 30.78 & 21.74 & 7.29 & 8.29 & 25.20 & 23.16 & 22.26 & 20.11 & 28.37 & 22.21 & 31.04 & 25.51 \\
id\_google\_play\_review & ind & RR & 9.59 & 12.94 & 51.22 & 23.70 & 42.38 & 14.82 & 43.70 & 16.70 & 73.90 & 37.64 & 63.70 & 35.97 \\
id\_google\_play\_review\_posneg & ind & SA & 92.32 & 85.63 & 89.78 & 76.38 & 83.84 & 52.25 & 83.25 & 65.93 & 93.44 & 87.15 & 87.76 & 81.28 \\
id\_hatespeech & ind & HSD & 77.70 & 75.31 & 59.47 & 59.39 & 40.21 & 34.32 & 36.19 & 32.76 & 62.60 & 40.44 & 64.56 & 42.23 \\
id\_hoax\_news & ind & HD & 48.40 & 46.51 & 45.60 & 45.54 & 37.87 & 30.23 & 39.33 & 34.64 & 53.07 & 38.20 & 53.07 & 44.96 \\
id\_hsd\_nofaaulia & ind & HSD & 72.53 & 51.40 & 67.03 & 51.94 & 30.04 & 28.18 & 30.77 & 29.30 & 63.74 & 41.40 & 76.19 & 51.75 \\
id\_short\_answer\_grading & ind & SAG & 17.50 & 7.56 & 17.75 & 5.03 & 12.58 & 4.67 & 21.29 & 8.55 & 20.41 & 8.70 & 19.24 & 5.45 \\
id\_stance & ind & SD & 10.68 & 8.12 & 24.04 & 16.04 & 48.96 & 26.64 & 63.20 & 25.82 & 31.55 & 14.41 & 58.46 & 29.33 \\
imdb\_jv & jav & SA & 21.01 & 14.45 & 33.22 & 20.71 & 49.47 & 38.74 & 48.77 & 39.44 & 37.42 & 31.25 & 31.27 & 31.13 \\
indo\_law & ind & LG & 53.64 & 53.51 & 53.90 & 53.40 & 52.58 & 34.52 & 52.59 & 34.46 & 52.64 & 43.29 & 47.38 & 35.67 \\
indolem\_ntp & ind & NTP & 61.26 & 32.95 & 31.69 & 20.72 & 33.87 & 32.56 & 43.20 & 40.03 & 76.21 & 57.08 & 77.60 & 53.95 \\
indolem\_sentiment & ind & SA & 70.82 & 69.91 & 55.49 & 55.48 & 69.83 & 56.86 & 59.05 & 55.10 & 81.80 & 77.19 & 82.33 & 79.15 \\
indonli & ind & NLI & 35.77 & 28.02 & 35.52 & 27.84 & 35.82 & 29.71 & 35.59 & 24.36 & 52.27 & 41.38 & 56.82 & 44.62 \\
indotacos & ind & LG & 5.13 & 3.81 & 16.15 & 9.01 & 4.93 & 3.87 & 4.51 & 1.51 & 12.00 & 7.52 & 5.47 & 2.92 \\
jadi\_ide & jav & DI & 41.61 & 33.73 & 30.12 & 28.18 & 33.33 & 21.06 & 33.33 & 22.70 & 33.40 & 18.31 & 33.75 & 17.99 \\
karonese\_sentiment & btx & SA & 41.30 & 39.82 & 27.50 & 16.96 & 34.20 & 17.04 & 37.07 & 21.63 & 35.30 & 25.70 & 40.70 & 21.44 \\
nusax\_senti\_ace & ace & SA & 53.50 & 44.74 & 39.50 & 22.25 & 38.42 & 19.52 & 44.92 & 33.71 & 60.92 & 46.59 & 60.92 & 46.17 \\
nusax\_senti\_ban & ban & SA & 54.50 & 44.20 & 45.50 & 31.61 & 39.33 & 21.23 & 50.17 & 37.30 & 60.08 & 45.75 & 61.42 & 46.50 \\
nusax\_senti\_bjn & bjn & SA & 61.50 & 53.97 & 47.00 & 33.34 & 38.33 & 18.97 & 38.67 & 26.98 & 49.67 & 36.42 & 52.42 & 39.56 \\
nusax\_senti\_bug & bug & SA & 44.00 & 40.43 & 37.50 & 18.58 & 39.58 & 22.15 & 49.33 & 35.79 & 65.58 & 50.09 & 67.33 & 51.14 \\
nusax\_senti\_eng & eng & SA & 71.75 & 61.48 & 55.75 & 43.33 & 38.08 & 18.95 & 38.25 & 24.91 & 46.67 & 33.78 & 49.33 & 37.04 \\
nusax\_senti\_ind & ind & SA & 70.50 & 59.28 & 59.25 & 46.83 & 54.75 & 41.21 & 58.25 & 43.79 & 73.17 & 55.51 & 73.33 & 55.70 \\
nusax\_senti\_jav & jav & SA & 64.75 & 55.11 & 54.25 & 41.69 & 45.00 & 29.42 & 61.42 & 46.05 & 73.25 & 55.54 & 73.75 & 55.93 \\
nusax\_senti\_mad & mad & SA & 60.25 & 51.24 & 44.00 & 29.23 & 40.67 & 22.70 & 52.58 & 39.47 & 66.08 & 50.50 & 66.00 & 50.11 \\
nusax\_senti\_min & min & SA & 62.00 & 53.36 & 49.00 & 36.17 & 38.50 & 19.46 & 47.67 & 35.69 & 58.00 & 44.17 & 60.50 & 46.09 \\
nusax\_senti\_nij & nij & SA & 54.75 & 47.52 & 42.00 & 26.59 & 38.67 & 20.25 & 52.42 & 39.39 & 63.17 & 48.18 & 65.00 & 49.39 \\
nusax\_senti\_sun & sun & SA & 63.25 & 53.50 & 49.50 & 37.25 & 38.67 & 20.12 & 43.50 & 31.58 & 57.17 & 43.36 & 59.75 & 45.34 \\
nusax\_senti\_bbc & bbc & SA & 46.50 & 37.67 & 39.75 & 23.11 & 39.25 & 20.85 & 49.83 & 37.25 & 57.50 & 43.77 & 58.58 & 44.50 \\
sentiment\_nathasa\_review & ind & SA & 25.41 & 21.71 & 14.00 & 12.12 & 20.33 & 12.44 & 16.67 & 8.12 & 29.93 & 24.23 & 30.76 & 27.69 \\
smsa & ind & SA & 80.20 & 64.69 & 71.40 & 53.24 & 55.33 & 37.28 & 69.33 & 50.92 & 79.87 & 58.38 & 80.00 & 58.50 \\
su\_emot & sun & SA & 45.95 & 44.35 & 29.59 & 27.72 & 28.49 & 16.25 & 25.06 & 11.16 & 34.59 & 25.25 & 24.99 & 10.16 \\
wrete & ind & ENT & 44 & 30.56 & 4.00 & 3.85 & 47.33 & 29.97 & 59.33 & 31.59 & 11.67 & 9.98 & 30.67 & 23.46 \\
\bottomrule
\end{tabular}
}
\caption{Details of zero-shot generalization to NLU tasks in NusaNLU. \textbf{EMOT} denotes emotion classification, \textbf{AD} denotes abusive detection, \textbf{RR} denotes review rating, \textbf{SA} denotes sentiment analysis, \textbf{HD} denotes hoax detection, \textbf{HSD} denotes hate speech detection, \textbf{SAG} denotes short answer grading, \textbf{SF} denotes stance detection,  \textbf{LG} denotes legal classification, \textbf{NLI} denotes natural language inference, \textbf{NTP} denotes next tweet prediction, and \textbf{ENT} denotes entailment.}
\label{tab:appendix_zeroshotnlu}
\end{table*}

\clearpage

\begin{table*}[!ht]
\centering
\resizebox{\linewidth}{!}{%
\begin{tabular}{l|c|c|cc|cc|cc|cc}
\toprule
\multirow{2}{*}{\textbf{Dataset Name}} & \multirow{2}{*}{\textbf{Lang}} & \multirow{2}{*}{\textbf{Task}} & \multicolumn{2}{|c|}{\textbf{XGLM 2.9B (id)}} & \multicolumn{2}{|c|}{\textbf{XGLM 2.9B (en)}} & \multicolumn{2}{|c|}{\textbf{BLOOMZ 3B (id)}} & \multicolumn{2}{|c}{\textbf{BLOOMZ 3B (en)}} \\ \cmidrule{4-11}
 & & & \small{\textbf{SacreBLEU}} & \small{\textbf{ROUGEL}} & \small{\textbf{SacreBLEU}} & \small{\textbf{ROUGEL}} & \small{\textbf{SacreBLEU}} & \small{\textbf{ROUGEL}} & \small{\textbf{SacreBLEU}} & \small{\textbf{ROUGEL}} \\ \midrule
\textbf{bible\_en\_id} & eng-ind & MT & 1.11 & 8.41 & 0.49 & 3.69 & 3.39 & 20.52 & 3.82 & 22.81 \\
\textbf{bible\_jv\_id} & jav-ind & MT & 0.18 & 5.13 & 0.10 & 1.16 & 0.47 & 7.23 & 0.43 & 8.02 \\
\textbf{bible\_su\_id} & sun-ind & MT & 0.18 & 4.95 & 0.05 & 1.20 & 0.44 & 7.11 & 0.64 & 7.81 \\
\textbf{id\_panl\_bppt} & eng-ind & MT & 2.58 & 10.56 & 0.48 & 2.77 & 6.36 & 26.43 & 7.81 & 28.98 \\
\textbf{indo\_general\_mt\_en\_id} & eng-ind & MT & 3.51 & 9.41 & 1.35 & 3.40 & 8.82 & 25.51 & 9.92 & 26.82 \\
\textbf{indo\_religious\_mt\_en\_id} & eng-ind & MT & 1.80 & 9.27 & 0.25 & 1.81 & 4.49 & 22.06 & 5.94 & 26.76 \\
\textbf{minangnlp\_mt} & min-ind & MT & 2.12 & 7.14 & 2.25 & 7.19 & 5.15 & 17.81 & 7.34 & 26.68 \\
\textbf{news\_en\_id} & eng-ind & MT & 2.66 & 10.26 & 0.48 & 2.62 & 6.56 & 24.56 & 8.83 & 28.92 \\
\textbf{nusax\_mt\_ace\_ind} & ace-ind & MT & 0.72 & 6.55 & 0.47 & 3.19 & 1.47 & 11.47 & 1.77 & 13.28 \\
\textbf{nusax\_mt\_ban\_ind} & ban-ind & MT & 1.09 & 6.94 & 0.82 & 3.68 & 2.58 & 12.92 & 2.54 & 14.61 \\
\textbf{nusax\_mt\_bbc\_ind} & bbc-ind & MT & 1.59 & 7.94 & 0.89 & 4.33 & 3.03 & 15.49 & 3.96 & 18.59 \\
\textbf{nusax\_mt\_bjn\_ind} & bjn-ind & MT & 0.30 & 4.43 & 0.28 & 2.06 & 0.52 & 6.80 & 0.58 & 7.66 \\
\textbf{nusax\_mt\_bug\_ind} & bug-ind & MT & 2.06 & 10.31 & 0.19 & 2.50 & 3.12 & 22.78 & 4.33 & 24.55 \\
\textbf{nusax\_mt\_eng\_ind} & eng-ind & MT & 0.72 & 4.77 & 0.66 & 3.38 & 1.00 & 6.84 & 1.76 & 10.35 \\
\textbf{nusax\_mt\_ind\_ace} & ind-ace & MT & 1.03 & 5.26 & 0.77 & 4.30 & 1.72 & 8.34 & 3.43 & 15.45 \\
\textbf{nusax\_mt\_jav\_ind} & jav-ind & MT & 0.53 & 5.47 & 0.30 & 2.48 & 1.13 & 10.00 & 1.38 & 10.71 \\
\textbf{nusax\_mt\_mad\_ind} & mad-ind & MT & 1.81 & 8.64 & 1.06 & 4.83 & 3.39 & 16.11 & 5.28 & 20.66 \\
\textbf{nusax\_mt\_min\_ind} & min-ind & MT & 0.64 & 6.14 & 0.63 & 3.38 & 1.39 & 11.40 & 1.68 & 12.07 \\
\textbf{nusax\_mt\_nij\_ind} & nij-ind & MT & 0.84 & 6.73 & 0.81 & 4.08 & 2.01 & 12.80 & 2.26 & 13.54 \\
\textbf{nusax\_mt\_sun\_ind} & sun-ind & MT & 0.39 & 5.00 & 0.29 & 2.40 & 0.98 & 8.94 & 1.18 & 9.96 \\
\textbf{nusax\_mt\_ind\_ban} & ind-ban & MT & 1.67 & 7.19 & 1.10 & 4.88 & 1.83 & 10.43 & 3.66 & 15.10 \\
\textbf{nusax\_mt\_ind\_bbc} & ind-bbc & MT & 0.33 & 2.46 & 0.36 & 2.64 & 0.47 & 4.05 & 0.67 & 4.69 \\
\textbf{nusax\_mt\_ind\_bjn} & ind-bjn & MT & 0.57 & 3.70 & 2.15 & 12.76 & 3.16 & 19.82 & 4.66 & 24.47 \\
\textbf{nusax\_mt\_ind\_bug} & ind-buh & MT & 0.86 & 4.75 & 0.70 & 3.73 & 1.39 & 8.92 & 2.28 & 12.59 \\
\textbf{nusax\_mt\_ind\_eng} & ind-buh & MT & 0.44 & 3.50 & 0.42 & 3.00 & 0.77 & 5.73 & 0.86 & 6.09 \\
\textbf{nusax\_mt\_ind\_jav} & ind-jav & MT & 1.87 & 7.69 & 0.85 & 4.30 & 2.47 & 10.62 & 6.01 & 21.41 \\
\textbf{nusax\_mt\_ind\_mad} & ind-mad & MT & 0.71 & 4.69 & 0.59 & 3.31 & 0.72 & 6.98 & 1.36 & 8.91 \\
\textbf{nusax\_mt\_ind\_min} & ind-min & MT & 0.99 & 5.41 & 0.76 & 3.80 & 2.00 & 10.50 & 4.01 & 15.56 \\
\textbf{nusax\_mt\_ind\_nij} & ind-nij & MT & 0.50 & 3.39 & 0.47 & 3.30 & 0.71 & 4.83 & 1.43 & 7.87 \\
\textbf{nusax\_mt\_ind\_sun} & ind-sun & MT & 1.14 & 7.27 & 0.59 & 3.74 & 2.11 & 13.00 & 2.06 & 13.87 \\
\textbf{parallel\_su\_id} & sun-ind & MT & 0.91 & 6.70 & 0.89 & 3.84 & 3.88 & 13.71 & 4.68 & 14.95 \\
\textbf{ted\_en\_id} & eng-ind & MT & 2.68 & 10.99 & 0.49 & 2.44 & 7.94 & 28.55 & 8.90 & 31.34 \\
\textbf{ud\_id\_csui} & ind-eng & MT & 1.06 & 3.69 & 5.58 & 16.68 & 8.54 & 27.28 & 14.94 & 38.89 \\
\textbf{indosum\_fold0} & ind & SUM & 2.64 & 5.53 & 9.73 & 12.24 & 3.55 & 12.21 & 3.07 & 11.88 \\
\textbf{xl\_sum} & ind & SUM & 0.35 & 4.05 & 0.33 & 3.95 & 2.66 & 12.68 & 3.30 & 14.31 \\
\textbf{liputan6} & ind & SUM & 1.47 & 4.12 & 1.92 & 7.48 & 3.62 & 13.92 & 2.67 & 13.13 \\ 
\midrule
 & & & \small{\textbf{EM}} & \small{\textbf{F1}} & \small{\textbf{EM}} & \small{\textbf{F1}} & \small{\textbf{EM}} & \small{\textbf{F1}} & \small{\textbf{EM}} & \small{\textbf{F1}} \\ \midrule
\textbf{facqa} & ind & QA & 0.21 & 5.61 & 0.21 & 4.72 & 61.95$^*$ & 81.97$^*$ & 61.95$^*$ & 82.02$^*$ \\ 
\textbf{idk\_mrc} & ind & QA & 0.71 & 3.67 & 0.28 & 4.31 & 37.05$^*$ & 45.51$^*$ & 34.91$^*$ & 43.57$^*$ \\ 
\textbf{tydiqa\_ind} & ind & QA & 0.31 & 5.52 & 0.27 & 4.67 & 63.08$^*$ & 77.28$^*$ & 62.85$^*$ & 77.09$^*$ \\ 
\bottomrule
\end{tabular}%
}
\caption{Details of zero-shot generalization to NLG tasks in NusaNLG. \textbf{MT} denotes machine translation, \textbf{SUM} denotes summarization, and \textbf{QA} denotes question answering. \textbf{*} the test data might be (partially) seen during the language pre-training and prompt tuning phase.}
\label{tab:appendix_zeroshotnlg}
\end{table*}

\clearpage

\begin{table*}
    \centering
    \resizebox{0.65\linewidth}{!}{
        \begin{tabular}{l|c|c|c|c|c|c}
        \toprule
        \multirow{2}{*}{\textbf{Dataset}} & \multirow{2}{*}{\textbf{Lang}} & \small{\textbf{Multilingual}}  & \multicolumn{3}{|c|}{\small{\textbf{Monolingual Multi-task}}} & \multirow{2}{*}{\small{\textbf{Single-task }}} \\
        \cmidrule{4-6}
         &  & \small{\textbf{Multi-task}} & \small{\textbf{ind}} & \small{\textbf{jav}} & \small{\textbf{sun}} &  \\
        \midrule
        indspeech\_digit\_cdsr & ind &  2.36 & 0.22 & 97.51 & 97.65 & 43.84 \\
        indspeech\_news\_lvcsr & ind &  13.04 & 3.31 & 84.88 & 81.39 & 1.41 \\
        indspeech\_teldialog\_lvcsr & ind & 1.59 & 0.37 & 81.16 & 77.65 & 0.45 \\
        indspeech\_teldialog\_svcsr & ind & 9.01 & 1.72 & 94.58 & 97.03 & 0.65 \\
        librivox\_indonesia\_ind & ind &  14.24 & 15.41 & 86.23 & >100 & 16.20 \\
        librivox\_indonesia\_ace & ace &  40.85 & 95.14 & >100 & 92.36 & 100 \\
        indspeech\_newstra\_ethnicsr\_ban & ban & 12.21 & >100 & 56 & 59.14 & 43.98 \\
        librivox\_indonesia\_ban & ban &  21.24 & >100 & 78.04 & >100 & 100 \\
        indspeech\_newstra\_ethnicsr\_btk & btk & 18.98 & >100 & 81.24 & 74.67 & 64.77 \\
        librivox\_indonesia\_bug & bug &  41.59 & 96.70 & >100 & >100 & 100 \\
        indspeech\_news\_ethnicsr\_jv & jav & 13.37 & >100 & 37.50 & 93.32 & 100 \\
        indspeech\_newstra\_ethnicsr\_jav & jav & 13.30 & >100 & 42.70 & 72.77 & 57.34 \\
        librivox\_indonesia\_jav & jav &  29.05 & 81.18 & 60.70 & >100 & 100 \\
        librivox\_indonesia\_min & min &  16.94 & 46.19 & 68.10 & 98.57 & 100 \\
        indspeech\_news\_ethnicsr\_su & sun & 20.54 & 84.08 & 84.17 & 50.17 & 100 \\
        indspeech\_newstra\_ethnicsr\_sun & sun & 12.41 & >100 & 65.40 & 47.38 & 49.03 \\
        librivox\_indonesia\_sun & sun & 8.85 & 60.69 & 60.12 & 23.70 & 100 \\
        \midrule
        \multicolumn{2}{c|}{\textbf{Average}} & \textbf{17.03} & \textbf{>100} & \textbf{75.98} & \textbf{82.37} & \textbf{63.39} \\
        \bottomrule
        \end{tabular}
    }
    \caption{Per task WER (lower is better) of the \textbf{wav2vec 2.0-pt} model on all 17 ASR tasks in NusaASR.}
    \label{tab:appendix_multilingualasr_wav2vec_pt}
\end{table*}

\begin{table*}
    \centering
    \resizebox{0.65\linewidth}{!}{
        \begin{tabular}{l|c|c|c|c|c|c}
        \toprule
        \multirow{2}{*}{\textbf{Dataset}} & \multirow{2}{*}{\textbf{Lang}} & \small{\textbf{Multilingual}}  & \multicolumn{3}{|c|}{\small{\textbf{Monolingual Multi-task}}} & \multirow{2}{*}{\small{\textbf{Single-task }}} \\
        \cmidrule{4-6}
         &  & \small{\textbf{Multi-task}} & \small{\textbf{ind}} & \small{\textbf{jav}} & \small{\textbf{sun}} &  \\
        \midrule
        indspeech\_digit\_cdsr & ind &  0.38 & 0.18 & 92.86 & 97.46 & 0.22 \\
        indspeech\_news\_lvcsr & ind &  0.82 & 0.65 & 67.33 & 80.12 & 1.37 \\
        indspeech\_teldialog\_lvcsr & ind & 0.59 & 0.09 & 68.65 & 74.26 & 0.22 \\
        indspeech\_teldialog\_svcsr & ind & 0.41 & 0.33 & 92.92 & 94.89 & 0.46 \\
        librivox\_indonesia\_ind & ind &  7.32 & 8.11 & 72.57 & >100 & 8.37 \\
        librivox\_indonesia\_ace & ace &  31.94 & 91.67 & 90.28 & 89.58 & 49.31 \\
        indspeech\_newstra\_ethnicsr\_ban & ban & 22.95 & >100 & 36.57 & 43.69 & 21.50 \\
        librivox\_indonesia\_ban & ban &  19.16 & >100 & 68.69 & >100 & 35.98 \\
        indspeech\_newstra\_ethnicsr\_btk & btk & 35.99 & >100 & 59.79 & 61.34 & 40.92 \\
        librivox\_indonesia\_bug & bug &  53.30 & >100 & >100 & >100 & 90.09 \\
        indspeech\_news\_ethnicsr\_jv & jav & 22.30 & >100 & 17.90 & 96.73 & 27.13 \\
        indspeech\_newstra\_ethnicsr\_jav & jav & 21.41 & >100 & 22.10 & 50.14 & 25.10 \\
        librivox\_indonesia\_jav & jav &  38.93 & 96.49 & 41.70 & >100 & 44.10 \\
        librivox\_indonesia\_min & min &  18.10 & 70.48 & 52.86 & 79.05 & 24.29 \\
        indspeech\_news\_ethnicsr\_su & sun & 35.47 & 87.02 & 74.13 & 43.25 & 44.38 \\
        indspeech\_newstra\_ethnicsr\_sun & sun & 19.39 & >100 & 39.67 & 25.71 & 20.45 \\
        librivox\_indonesia\_sun & sun & 7.51 & 67.63 & 49.13 & 6.36 & 15.03 \\
        \midrule
        \multicolumn{2}{c|}{\textbf{Average}} & \textbf{19.76} & \textbf{>100} & \textbf{61.76} & \textbf{75.76} & \textbf{26.41} \\
        \bottomrule
        \end{tabular}
    }
    \caption{Per task WER (lower is better) of the \textbf{wav2vec 2.0-ft} modes on all 17 ASR tasks in NusaASR.}
    \label{tab:appendix_multilingualasr_wav2vec_ft}
\end{table*}

\begin{table*}
    \centering
    \resizebox{0.65\linewidth}{!}{
        \begin{tabular}{l|c|c|c|c|c|c}
        \toprule
        \multirow{2}{*}{\textbf{Dataset}} & \multirow{2}{*}{\textbf{Lang}} & \small{\textbf{Multilingual}}  & \multicolumn{3}{|c|}{\small{\textbf{Monolingual Multi-task}}} & \multirow{2}{*}{\small{\textbf{Single-task }}} \\
        \cmidrule{4-6}
         &  & \small{\textbf{Multi-task}} & \small{\textbf{ind}} & \small{\textbf{jav}} & \small{\textbf{sun}} &  \\
        \midrule
        indspeech\_digit\_cdsr & ind & 0.27 & 0.30 & 43.61 & 79.08 & 3.94 \\
        indspeech\_news\_lvcsr & ind & 2.56 & 2.40 & 59.95 & 97.42 & 2.08 \\
        indspeech\_teldialog\_lvcsr & ind & 0.38 & 0.57 & 58.14 & 94.44 & 0.97 \\
        indspeech\_teldialog\_svcsr & ind & 50.85 & 50.91 & 72.47 & 87.69 & 100 \\
        librivox\_indonesia\_ind & ind & 87.43 & 17.32 & 88.15 & 99.70 & 10.13 \\
        librivox\_indonesia\_ace & ace & 97.92 & 99.31 & 98.61 & >100 & 37.50 \\
        indspeech\_newstra\_ethnicsr\_ban & ban & 51.89 & 51.81 & 57.85 & 95.56 & 70.66 \\
        librivox\_indonesia\_ban & ban & 88.32 & >100 & 87.38 & 99.07 & 39.25 \\
        indspeech\_newstra\_ethnicsr\_btk & btk & 63.85 & 56.52 & 82.16 & 98.39 & >100 \\
        librivox\_indonesia\_bug & bug & 100 & >100 & 99.06 & >100 & 56.60 \\
        indspeech\_news\_ethnicsr\_jv & jav & 58.24 & 96.31 & 30.82 & >100 & 53.84 \\
        indspeech\_newstra\_ethnicsr\_jav & jav & 47.55 & 49.45 & 42.77 & 95.99 & 73.60 \\
        librivox\_indonesia\_jav & jav & 85.71 & >100 & 46.94 & >100 & 51.84 \\
        librivox\_indonesia\_min & min & 80.95 & 82.38 & 78.57 & 90.95 & 36.67 \\
        indspeech\_news\_ethnicsr\_su & sun & 64.18 & 97.28 & 78.37 & 68.91 & 67.02 \\
        indspeech\_newstra\_ethnicsr\_sun & sun & 54.65 & 53.51 & 66.26 & 73.02 & 72.71 \\
        librivox\_indonesia\_sun & sun & 77.46 & 88.44 & 68.21 & 41.62 & 27.17 \\
        \midrule
        \multicolumn{2}{c|}{\textbf{Average}} & \textbf{59.54} & \textbf{62.83} & \textbf{68.20} & \textbf{89.98} & \textbf{49.44} \\
        \bottomrule
        \end{tabular}
    }
    \caption{Per task WER (lower is better) of the \textbf{Whisper} model on all 17 ASR tasks in NusaASR.}
    \label{tab:appendix_multilingualasr_whisper}
\end{table*}

\begin{table*}
    \centering 
    \begin{tabular}{p{0.58\linewidth}|c|c|p{0.18\linewidth}}
        \toprule
        \textbf{Dataset} & \textbf{Task} &  \textbf{Modal} & \textbf{Languages} \\
        \midrule
        Korpus Nusantara~\cite{indrayana2016meningkatkan,hasbiansyah2016tuning,ningtyas2018smt,etsa2018smt,darwis2019peningkatan,wahyuni2019smt,sujaini2019mdsun,sujaini2020improving,gunawan2021analisis} & MT & Text & ind, jav, xdy, bug, sun, mad, bjn, bbc, hak, msa, min, nan \\ \midrule
        Karonese dataset~\cite{karo2022sentiment} & SA & Text & btx \\ \midrule
        Sundanese-Indonesian Parallel Corpus~\cite{arie2022suidparallel} & MT & Text & sun, ind \\ \midrule
        PoSTagged Sundanese Monolingual Corpus~\cite{arie2022sunpos} & POS & Text & sun \\ \midrule
        Code-mixed Sentiment JV-ID~\cite{tho2021codemixed} & SA & Text & ind \\ \midrule
        ID-HSD-Nofaaulia~\cite{aulia2019hsd} & HSD & Text & ind \\ \midrule
        Indo Wiki Paralel Corpora~\cite{trisedya2014wiki-idjv} & MT & Text & ind, jav, sun, min \\ \midrule
        INDspeech\_DIGIT\_CDSR~\cite{sakti2004indonesian} & ASR & Speech & ind \\ \midrule
        INDspeech\_NEWS\_EthnicSR~\cite{sani2012cocosda} & ASR & Speech & sun, jav \\ \midrule        INDspeech\_NEWS\_LVCSR~\cite{sakti2008development,sakti2004indonesian,sakti-s2st-csl-2013} & ASR & Speech & ind \\ \midrule
        INDspeech\_NEWS\_TTS~\cite{sakti2008hmm,sakti-tts-malindo-2010,sakti-s2st-csl-2013} & TTS & Speech & ind \\ \midrule        INDspeech\_NEWSTRA\_EthnicSR~\cite{sakti2013towards,sakti-sltu-2014,novitasari-sltu-2020} & ASR & Speech & sun, jav, btk, ban \\ \midrule        INDspeech\_TELDIALOG\_LVCSR~\cite{sakti2008development,sakti2004indonesian,sakti-s2st-csl-2013} & ASR & Speech & ind \\ \midrule
        INDspeech\_TELDIALOG\_SVCSR~\cite{sakti2004indonesian} & ASR & Speech & ind \\
        \bottomrule
    \end{tabular}
    \caption{List of private datasets that have been made public through NusaCrowd initiative. \textbf{MT} denotes machine translation, \textbf{SA} denotes sentiment analysis, \textbf{POS} denotes POS tagging, \textbf{HSD} denotes hate speech detection, \textbf{ASR} denotes automatic speech recognition, and \textbf{TTS} denotes text-to-speech.}
    \label{tab:private-dataset}
\end{table*}

\begin{table*}
    \small
    \resizebox{\linewidth}{!}{
    \begin{tabular}{lccccc}
        \toprule
        \multirow{2}{*}{\textbf{Project}} & \multirow{2}{*}{\textbf{\#Dataset}} & \multirow{2}{*}{\textbf{Languages}} & \multirow{2}{*}{\textbf{Modality}} & \textbf{Open/crowd} & \textbf{Opening access}  \\
        & & & & \textbf{initiative?} & \textbf{to private data}  \\
        \midrule
        \multicolumn{6}{c}{\textit{Global  Data Initiative}} \\
        \midrule
        XTREME~\cite{hu2020xtreme} & 9 & 40 & text & \xmark & \xmark \\
        XGLUE~\cite{liang2020xglue} & 11 & 19 & text & \xmark & \xmark \\
        GEM~\cite{gehrmann2021gem} & 11 & 18 & text & \cmark & \xmark \\
        GEMv2~\cite{gehrmann2022gemv2} & 40 & 51 & text & \cmark & \xmark \\
        \midrule
        \multicolumn{6}{c}{\textit{Other Regional Data Initiative}} \\
        \midrule
        CLUE~\cite{xu2020clue} & 9 & 1 (zho) & text & \xmark & \xmark \\
        KLUE~\cite{park2021klue} & 8 & 1 (kor) & text & \xmark & \xmark \\
        ALUE~\cite{seelawi-etal-2021-alue} & 9 & 1 (ara) & text & \xmark & \xmark \\
        IndicGLUE~\cite{kakwani2020indicnlpsuite} & 14 & 12 Indian languages & text & \xmark & \xmark \\
        IndicNLG~\cite{kumar-etal-2022-indicnlg} & 5 & 11 Indian languages & text & \xmark & \xmark \\
        IndicXTREME~\cite{doddapaneni2022indicxtreme} & 103 & 18 Indian languages & text & \xmark & \xmark \\
        \midrule
        \multicolumn{6}{c}{\textit{Indonesian  Data Initiative}} \\
        \midrule
        IndoNLU~\cite{wilie2020indonlu} & 12 & 1 (ind) & text & \xmark & \xmark \\
        IndoLEM~\cite{koto2020indolem} & 12 & 1 (ind) & text & \xmark & \xmark \\
        IndoNLG~\cite{cahyawijaya-etal-2021-indonlg} & 10 & 3 (ind, sun, jav) & text & \xmark & \xmark \\
        NusaCrowd & 137 & 19 Indonesian languages & text, speech, image & \cmark & \cmark \\
        \bottomrule
    \end{tabular}
    }
    \caption{Comparison of NusaCrowd with other similar initiatives.}
    \label{tab:data-comparison}
\end{table*}

\clearpage
\newgeometry{left=1cm,bottom=1.5cm,right=1cm,top=1cm}
\renewcommand\_{\textunderscore\allowbreak}

\begin{landscape}
\begin{longtable}{p{0.08\linewidth}|p{0.3\linewidth}|p{0.05\linewidth}|p{0.05\linewidth}|p{0.07\linewidth}|p{0.1\linewidth}|p{0.05\linewidth}|p{0.05\linewidth}|p{0.1\linewidth}}

    \toprule \textbf{Dataset} & \textbf{Description} & \textbf{License} & \textbf{Year} & \textbf{Language} & \textbf{Annotation Quality} & \textbf{Data Volume} & \textbf{Data Unit} & \textbf{Split} \\ \midrule \midrule
    \endfirsthead

    \multicolumn{9}{c}{{\bfseries \tablename\ \thetable{} -- continued from the previous page}} \\
    \toprule \textbf{Dataset} & \textbf{Description} & \textbf{License} & \textbf{Year} & \textbf{Language} & \textbf{Annotation Quality} & \textbf{Data Volume} & \textbf{Data Unit} & \textbf{Split} \\ \midrule
    \endhead

    \multicolumn{9}{c}{\textbf{Continued on the next page}} \\
    \endfoot
    
    \endlastfoot

        AM2iCo~\cite{liu2021am2ico} & We present AM2ICO (Adversarial and Multilingual Meaning in Context), a wide coverage cross-lingual and multilingual evaluation set; it aims to faithfully assess the ability of state-of-the-art (SotA) representation models to understand the identity of word meaning in cross-lingual contexts for 14 language pairs. & CC-BY 4.0 & 2021 & ind, eng & Crawling \& human annotation & 3,098 & examples & 1589 train, 500 validation, 1000 test \\ \midrule 
        Barasa & Barasa: Indonesian SentiWordNet for sentiment analysis & MIT & 2015 & ind & Unknown & 16 & MB & No dataset split \\ \midrule 
        CASA~\cite{ilmania2018absa} & CASA is an aspect-based sentiment analysis dataset consisting of around a thousand car reviews collected from multiple Indonesian online automobile platforms. The dataset covers six aspects of car quality, where each label represents a sentiment for a single aspect with three possible values: positive, negative, and neutral. & CC-BY-SA 4.0 & 2018 & ind & Crawling \& human annotation & 1,080 & sentences & 810 train, 90 validation, 180 test \\ \midrule 
        CC100~\cite{conneau2020unsupervised} & CC100 comprises of monolingual data for 100+ languages and also includes data for romanized languages. This was constructed using the urls and paragraph indices provided by the CC-Net repository by processing January-December 2018 Commoncrawl snapshots. Each file comprises of documents separated by double-newlines and paragraphs within the same document separated by a newline. The data is generated using the open source CC-Net repository. & Common Crawl's license & 2020 & ind, sun, jav & Machine generated / Crawling w/o curation & 36,052 & MB & No dataset split \\ \midrule 
        COCO Captions ID~\cite{sinurat2019pembangkitan} & COCO Captions contains over one and a half million captions describing over 330,000 images. For the training and validation images, five independent human generated captions are be provided for each image. This is an Indonesian version of COCO translated using Google Translate. & CC-BY 4.0 & 2019 & ind & Machine generated / Crawling w/o curation & 123,287 & sentences & 113287 train, 5000 validation, 5000 test \\ \midrule 
        CORD~\cite{park2019cord} & In this paper, we introduce a novel dataset called CORD, which stands for a Consolidated Receipt Dataset for post-OCR parsing. To the best of our knowledge, this is the first publicly available dataset which includes both box-level text and parsing class annotations. The parsing class labels are provided in two-levels. The eight superclasses include store, payment, menu, subtotal, and total. The eight superclasses are subdivided into 54 subclasses, e.g., store has nine subclasses including name, address, telephone, and fax. Furthermore, it also provides line annotations for the serialization task which is a newly emerging problem as a combination of the two tasks. & CC-BY 4.0 & 2019 & ind & Crawling \& human annotation & 1,000 & receipts & 800 train, 100 validation, 100 test \\ \midrule 
        CVSS~\cite{jia2022cvss} & We introduce CVSS, a massively multilingual-to-English speech-to-speech translation (S2ST) corpus, covering sentence-level parallel S2ST pairs from 21 languages into English. CVSS is derived from the Common Voice (Ardila et al., 2020) speech corpus and the CoVoST 2 (Wang et al., 2021b) speech-to-text translation (ST) corpus, by synthesizing the translation text from CoVoST 2 into speech using state-of-the-art TTS systems. Two versions of translation speech in English are provided: 1) CVSS-C: All the translation speech is in a single high-quality canonical voice; 2) CVSS-T: The translation speech is in voices transferred from the corresponding source speech. In addition, CVSS provides normalized translation text which matches the pronunciation in the translation speech. & CC-BY 4.0 & 2022 & ind & Crawling \& human annotation & 6 & hours & 2.6 train, 1.8 validation, 1.9 test \\ \midrule 
        Cendana~\cite{moeljadi2019cendana} & Cendana is a linguistically annotated corpus that includes some grammatical analyses, such as parts-of-speech, phrases, relations between entities, and meaning representations. Cendana is built using tools developed in the Deep Linguistic Processing with HPSG (DELPHIN) community. & GNU General Public License, version 2 & 2019 & ind & Human generation \& curation & 552 & sentences & No dataset split \\ \midrule 
        CoVoST 2 ~\cite{wang2021covost} & With the aim to foster research in massive multilingual ST and ST for low resource language pairs, we release CoVoST 2, a large-scale multilingual ST corpus covering translations from 21 languages into English and from English into 15 languages. This represents the largest open dataset available to date from total volume and language coverage perspective. & CC0 & 2020 & ind, eng & Crawling \& human annotation & 3 & hours & 1 train, 1 validation, 1 test \\ \midrule 
        Code-mixed Sentiment JV-ID~\cite{tho2021codemixed} & Dataset terdiri dari  3.963 kalimat code-mixing dalam bahasa Indonesia dan bahasa Jawa yang dikumpulkan dari twitter. Label dataset terdiri dari 3 kelas sentimen, yaitu: positif, negatif, dan netral. Label sentimen dikumpulkan dengan melakukan anotasi manual  untuk setiap tweet. & CC-BY 3.0 & 2021 & ind & Human generation \& curation & 977 & tweets & No dataset split \\ \midrule 
        Cross-lingual Outline- based Dialogue (COD)~\cite{majewska_tacl_a_00539} & Cross-lingual Outline-based Dialogue dataset (termed COD) enables natural language understanding, dialogue state tracking, and end-to-end dialogue modelling and evaluation in 4 diverse languages: Arabic, Indonesian, Russian, and Kiswahili. The data covers multi domain instances, e.g., bank, travel, weather, movies, music. & Unknown & 2022 & ind & Machine generated w/ human curation & 194 & dialogues & 0 train, 92 validation, 102 test \\ \midrule 
        Customer Review (Natasha Skincare)~\cite{nurlaila2017natashacare} & This dataset is collected from tweet costumer review from Natasha Skincare. This dataset contain of label emotion (joy, sad, angry, fear, disgust, surprise, or no emotion). & Unknown & 2017 & ind & Crawling \& human annotation & 124,263 & tweets & 87120 train, 37143 validation, 37143 test \\ \midrule 
        EmoT (IndoNLU Split)~\cite{saputri2018emotion} & EmoT is an emotion classification dataset collected from the social media platform Twitter. The dataset consists of around 4000 Indonesian colloquial language tweets, covering five different emotion labels: anger, fear, appiness, love, and sadness. & CC-BY-SA 4.0 & 2018 & ind & Human generation \& curation & 4,403 & sentence pairs & 3521 train, 440 validation, 442 test \\ \midrule 
        EmotCMT~\cite{Yulianti2021-kx} & EmotCMT is a Indonesian-English code-switching data collected from Twitter for emotion classification task. & Unknown & 2021 & ind, eng & Crawling \& Human annotation & 582 & sentences & No dataset split \\ \midrule 
        Emotion Indonesian Public Opinion~\cite{ricossan2022emotpubopini} & The dataset is formed from Indonesian tweet containing six emotion values, namely anger, fear, joy, love, sad, and neutral. The total data in this dataset is 7,080 and it is fully cleaned and fully annotated. Each label has a varied amount of data distribution, including 1,130 data for anger, 911 data for fear, 1,275 data for joy, 760 data for love, 1,003 data for sad, and 2,001 data for neutral. & CC-BY-NC-ND 4.0 & 2022 & ind & Human generation \& curation & 7,080 & tweets & No dataset split \\ \midrule 
        FacQA~\cite{purwarianti2007facqa} & The goal of the FacQA dataset is to find the answer to a question from a provided short passage from a news article . Each row in the FacQA dataset consists of a question, a short passage, and a label phrase, which can be found inside the corresponding short passage. There are six categories of questions: date, location, name, organization, person, and quantitative. & CC-BY-SA 4.0 & 2007 & ind & Human generation \& curation & 3,117 & documents & 2495 train, 311 dev, 311 test \\ \midrule 
        HoASA (IndoNLU Split)~\cite{azhar2019aspectcategorization} & HoASA is an aspect-based sentiment analysis dataset consisting of hotel reviews collected from the hotel aggregator platform, AiryRooms & CC-BY-SA 4.0 & 2019 & ind & Crawling \& human annotation & 9,450 & sentences & 7,560 train, 1890 test \\ \midrule 
        Human Instructions - Indonesian (wikihow)~\cite{ijcai2018p22} & Human Instructions - Indonesian (wikihow) is 39.246 Human Instructions in Indonesian Extracted from wikiHow. Step-by-step instructions in Indonesian extracted from wikiHow and decomposed into a formal graph representation in RDF. Instructions are represented in RDF following the PROHOW vocabulary and data model. For example, the category, steps, requirements and methods of each set of instructions have been extracted. This dataset has been produced as part of the The Web of Know-How project. & CC-BY-NC-SA 4.0 & 2017 & ind & Crawling \& human annotation & 39,246 & documents & No dataset split \\ \midrule 
        ID Abusive~\cite{ibrohim2018idabusive} & ID Abusive is a Twitter dataset for abusive language detection in Indonesian. Pre-defined abusive words are used as queries to collect the tweets. The dataset is labeled into 3 labels: not abusive language, abusive but not offensive, and offensive language by 20 volunteer annotators. & CC-BY-NC-SA 4.0 & 2018 & ind & Crawling \& human annotation & 2,016 & tweets & No dataset split \\ \midrule 
        ID Abusive Online News Comment~\cite{desrul2019abusivenewscomments} & The dataset consists of comments that are in some of the top news stories in 2019, obtained from several online news/forum, such as: kompas, kaskus, and detik. The labeling process is carried out by a total of 10 annotators and each comment is annotated by 3 annotators. Each comment was labeled with one of the following labels: not abusive, abusive but not offensive, abusive and offensive. & Unknown & 2019 & ind & Machine generated / Crawling w/o curation & 3,184 & comments & No dataset split \\ \midrule 
        ID Coreference Resolution~\cite{suherik2017idcoref} & ID Coreference resolution is news dataset aimed for coreference resolution task. This dataset consists of 1030 manually labelled sentences derived from IDENTIC parallel corpus. & Unknown & 2017 & ind & Crawling \& human annotation & 1,030 & sentences & 759 train, 0 validation, 108 test \\ \midrule 
        ID Multilabel HS~\cite{ibrohim2019multihs} & ID Multilabel HS consists of hate speech and abusive language Twitter dataset from several previous researches consisting of (Alfina et al., 2017, 2018), (Putri, 2018), and (Ibrohim and Budi, 2018), and new tweets queried using specific abusive words/phrases. Labels used in the dataset are: HS (hate speech label), Abusive (abusive language label), HS\_Individual (hate speech targeted to an individual), HS\_Group (hate speech targeted to a group), HS\_Religion (hate speech related to religion/creed), HS\_Race (hate speech related to race/ethnicity), HS\_Physical (hate speech related to physical/disability), HS\_Gender (hate speech related to gender/sexual orientation), HS\_Physical (hate related to other invective/slander), HS\_Weak (weak hate speech), HS\_Moderate (moderate hate speech), HS\_Strong (strong hate speech). & CC-BY-NC-SA 4.0 & 2019 & ind & Crawling \& human annotation & 13,169 & tweets & No dataset split \\ \midrule 
        ID Quora Question Pairs & Quora Question Pairs (QQP) dataset consists of over 400,000 question pairs, and each question pair is annotated with a binary value indicating whether the two questions are paraphrase of each other. This dataset is translated version of QQP to Indonesian Language. & Custom\footnote{\url{https://www.quora.com/about/tos}} & 2021 & ind & Machine generated / Crawling w/o curation & 149,011 & sentence pairs & 134084 train, 14927 validation, 0 test \\ \midrule 
        ID Short Answer Grading~\cite{JLK} & ID Short Answer Grading dataset is  dataset of questions using Edukasystem platform. It used 4 exams consisting of Biology and Geography subject.  Two  exams  are used for training data and 2 exams are used for testing data. Exam  used  for training data has 15 questions and exam for testing data has 3  questions. The dataset has 30  questions  + 7605  short answers  as  training  data  and  6  questions + 1560  short answers as testing data. The number of respondents is 534 different respondents. Assessment  was  carried  out  by  2  experts  for  Biology subject and 5 experts for Geography subject. The assessment is carried out on a scale of 0 to 5. & Unknown & 2020 & ind & Human generation \& curation & 9,165 & sentences & 7605 train, 0 validation, 1560 test \\ \midrule 
        ID-HSD-Nofaaulia~\cite{aulia2019hsd} & There have been many studies on detecting hate speech in short documents like Twitter data. But to our knowledge, research on long documents is rare, we suppose that the difficulty is increasing due to the possibility of the message of the text may be hidden. In this research, we explore in detecting hate speech on Indonesian long documents using machine learning approach. We build a new Indonesian hate speech dataset from Facebook.  & Unknown & 2022 & ind & Crawling \& human annotation & 906 & documents & 815 train, 0 validation, 91 test \\ \midrule 
        ID-HSD-Riomulia~\cite{alfina2017hsd} & ID-HSD-RioMulia composed of tweets about the Jakarta Governor Election 2017, whose selection of candidates potentially triggers hate speech in relation to race, religion, and gender. Each tweet is labelled as either containing hate speech or not by 30 human annotators. & GNU General Public License v3.0 & 2017 & ind & Crawling \& human annotation & 713 & tweets & No dataset split \\ \midrule 
        IDK-MRC~\cite{putri-oh-2022-idk} & IDK-MRC is an Indonesian Machine Reading Comprehension dataset that covers answerable and unanswerable questions. Based on the combination of the existing answerable questions in TyDiQA, the new unanswerable question in IDK-MRC is generated using question generation model and human-written question. Each paragraph in the dataset have a set of answerable and unanswerable question with the corresponding answer. (Note: the paper for this dataset is still under review on EMNLP 2022 -- the anonymity period ends on Oct 6, 2022) & CC-BY-SA 4.0 & 2022 & ind & Machine generated w/ human curation & 10,940 & paragraph, question, and answer pairs & 9332 train, 764 validation, 844 test \\ \midrule 
        IMDb Javanese~\cite{wongso2021imdbjv} & Large Movie Review Dataset translated to Javanese. This is a dataset for binary sentiment classification containing substantially more data than previous benchmark datasets. We provide a set of 25,000 highly polar movie reviews for training, and 25,000 for testing. & Unknown & 2021 & jav & Machine generated / Crawling w/o curation & 50,000 & sentences & 25000 train, 0 validation, 25000 test \\ \midrule 
        INDspeech\_DIGIT\_CDSR~\cite{sakti2004indonesian} & INDspeech\_DIGIT\_CDSR is the first Indonesian speech dataset for connected digit speech recognition (CDSR). The data was developed by TELKOMRisTI (R\&D Division, PT Telekomunikasi Indonesia) in collaboration with Advanced Telecommunication Research Institute International (ATR) Japan and Bandung Institute of Technology (ITB) under the Asia-Pacific Telecommunity (APT) project in 2004 [Sakti et al., 2004]. Although it was originally developed for a telecommunication system for hearing and speaking-impaired people, it can be used for other applications, i.e., automatic call centers that recognize telephone numbers. & CC-BY-NC-SA 4.0 & 2004 & ind & Human generation \& curation & 12444 [214] & utterances [speakers] & 8440 train, 0 validation, 4004 test \\ \midrule 
        INDspeech\_NEWSTRA\_EthnicSR~\cite{sakti2013towards} & INDspeech\_NEWSTRA\_EthnicSR is a collection of graphemically balanced and parallel speech corpora of four major Indonesian ethnic languages: Javanese, Sundanese, Balinese, and Bataks. It was developed in 2013 by the Nara Institute of Science and Technology (NAIST, Japan) [Sakti et al., 2013]. The data has been used to develop Indonesian ethnic speech recognition in supervised learning [Sakti et al., 2014] and semi-supervised learning [Novitasari et al., 2020] based on the Machine Speech Chain framework [Tjandra et al., 2020]. & CC-BY-NC-SA 4.0 & 2013 & ind, sun, jav, ban, btk & Human generation \& curation & 13000 [40] & utterances [speakers] & 9000 train, 0 validation, 4000 test \\ \midrule 
        INDspeech\_NEWS\_EthnicSR~\cite{sani2012cocosda} & INDspeech\_NEWS\_EthnicSR is a collection of Indonesian ethnic speech corpora  (Javanese and Sundanese) for Indonesian ethnic speech recognition. It was developed in 2012 by the Nara Institute of Science and Technology (NAIST, Japan) in collaboration with the Bandung Institute of Technology (ITB, Indonesia) [Sani et al., 2012]. Furthermore, as all speakers utter the same sentences, it can also be used for voice conversion tasks. & CC-BY-NC-SA 4.0 & 2012 & sun, jav & Human generation \& curation & 2300 [20] & utterances [speakers] & 2000 train, 0 validation, 300 test \\ \midrule 
        INDspeech\_NEWS\_LVCSR~\cite{sakti2008development} & INDspeech\_NEWS\_LVCSR is the first Indonesian speech dataset for large vocabulary continuous speech recognition (LVCSR) with more than 40 hours of speech and 400 speakers [Sakti et al., 2008]. R\&D Division of PT Telekomunikasi Indonesia (TELKOMRisTI) developed the data in 2005-2006, in collaboration with Advanced Telecommunication Research Institute International (ATR) Japan, as the continuation of the Asia-Pacific Telecommunity (APT) project [Sakti et al., 2004]. It has also been successfully used for developing Indonesian LVCSR in the Asian speech translation advanced research (A-STAR) project [Sakti et al., 2013]. & CC-BY-NC-SA 4.0 & 2008 & ind & Human generation \& curation & 44000 [400] & utterances [speakers] & 39600 train, 0 validation, 4400 test \\ \midrule 
        INDspeech\_NEWS\_TTS~\cite{sakti2008hmm} & INDspeech\_NEWS\_TTS is a speech dataset for developing an Indonesian text-to-speech synthesis system [Sakti et al., 2008, Sakti et al., 2010]. The data was developed by Advanced Telecommunication Research Institute International (ATR) Japan under the Asian speech translation advanced research (A-STAR) project [Sakti et al., 2013]. & CC-BY-NC-SA 4.0 & 2008 & ind & Human generation \& curation & 2,012 & utterances & 1972 train, 0 validation, 40 test \\ \midrule 
        INDspeech\_TELDIALOG\_LVCSR~\cite{sakti2008development} & INDspeech\_TELDIALOG\_LVCSR is one of the first Indonesian speech datasets for large vocabulary continuous speech recognition (LVCSR) [Sakti et al., 2008]. R\&D Division of PT Telekomunikasi Indonesia (TELKOMRisTI) developed the data in 2005-2006, in collaboration with Advanced Telecommunication Research Institute International (ATR) Japan, as the continuation of the Asia-Pacific Telecommunity (APT) project [Sakti et al., 2004]. It has also been successfully used for developing Indonesian LVCSR in the Asian speech translation advanced research (A-STAR) project [Sakti et al., 2013]. & CC-BY-NC-SA 4.0 & 2008 & ind & Human generation \& curation & 40000 [400] & utterances [speakers] & 36000 train, 0 validation, 4000 test \\ \midrule 
        INDspeech\_TELDIALOG\_SVCSR~\cite{sakti2004indonesian} & INDspeech\_TELDIALOG\_SVCSR is the first Indonesian speech dataset for small vocabulary continuous speech recognition (SVCSR). The data was developed by TELKOMRisTI (R\&D Division, PT Telekomunikasi Indonesia) in collaboration with Advanced Telecommunication Research Institute International (ATR) Japan and Bandung Institute of Technology (ITB) under the Asia-Pacific Telecommunity (APT) project in 2004 [Sakti et al., 2004]. Although it was originally developed for a telecommunication system for hearing and speaking impaired people, it can be used for other applications, i.e., automatic call centers. Furthermore, as all speakers utter the same sentences, it can also be used for voice conversion tasks. & CC-BY-NC-SA 4.0 & 2004 & ind & Human generation \& curation & 20000 [200] & utterances [speakers] & 10000 train, 0 validation, 10000 test \\ \midrule 
        Identic (Larasati, 2012)~\cite{larasati-2012-identic} & IDENTIC is an Indonesian-English parallel corpus for research purposes. The corpus is a bilingual corpus paired with English. The aim of this work is to build and provide researchers a proper Indonesian-English textual dataset and also to promote research in this language pair. The corpus contains texts coming from different sources with different genres. & Unknown & 2012 & eng, ind & Crawling \& human annotation & 45,000 & sentences & No dataset split \\ \midrule 
        Idn-tagged-corpus-CSUI~\cite{dinakaramani2014idn-tag-csui} & Id-tagged-corpus-CSUI is a POS tagging dataset contains about 10,000 sentences, collected from the PAN Localization Project tagged with 23 POS tag classes. & CC-BY-SA 4.0 & 2014 & ind & Crawling \& human annotation & 10,000 & sentences & 8000 train, 1000 validation, 1029 test \\ \midrule 
        InSet Lexicon~\cite{koto2017inset} & Sentiment analysis from Twitter & Unknown & 2017 & ind & Crawling \& human annotation & 2,630 & sentences &  2630 test \\ \midrule 
        IndQNER & IndQNER is a NER dataset created by manually annotating 8 chapters in the Indonesian translation of Quran text. The dataset consists of 2476 named entities from 18 categories. Each named entity is labeled using BIO (Beginning-Inside-Outside) tagging format. & Unknown & 2022 & ind & Human generation \& curation & 3,118 & sentences & 2494 train, 312 validation, 312 test \\ \midrule 
        Indo4B~\cite{wilie2020indonlu} & Indo4B is an Indonesian pre-training corpus collected from multiple online sources, Indo4B consists of ~3.6B tokens  and over more than 250M sentences. Indo4B has been used to pre-trained a large pre-trained language model called IndoBERT and IndoBERT-lite. & CC-BY-NC-SA 4.0 & 2020 & ind & Machine generated / Crawling w/o curation & ~3.6B & tokens & No dataset split \\ \midrule 
        Indo4B Plus~\cite{cahyawijaya-etal-2021-indonlg} & Indo4BPlus is an Indonesian pre-training corpus derived from Indo4B. Indo4BPlus covers three languages in Indonesia, i.e., Indonesian, Javanese, and Sundanese. Indo4BPlus consists of ~4B tokens with over more than 300M documents. & CC-BY-NC-SA 4.0 & 2021 & ind, sun, jav & Machine generated / Crawling w/o curation & ~4B & tokens & No dataset split \\ \midrule 
        IndoAMR~\cite{Ilmy2020parsing} & IndoAMR is annotated Indonesia AMR parser from Indonesian simple sentences. & Unknown & 2021 & ind & Crawling \& human annotation & 1,130 & sentences & 700 train, 0 validation, 300 test \\ \midrule 
        IndoCollex~\cite{wibowo2021indocollex} & List of Colloquial word Transformation with its label. e.g.: makan -> mkn (shortening). Data is published on IndoCollex: A Testbed for Morphological Transformation of Indonesian Word Colloquialism Research Paper published on ACL-IJCNLP 2021. Useful for morphological research. & MIT & 2021 & ind & Human generation \& curation & 2,126 & sentence pairs & 1637 train, 182 validation, 193 test \\ \midrule 
        IndoCoref~\cite{artari2021indocoref} & IndoCoref is a coreference resolution dataset collected from Wikipedia. IndoCoref consists of 201 passages from wikipedia with manually labelled coreference by five annotators. & MIT & 2021 & ind & Human generation \& curation & 201 & documents & No dataset split \\ \midrule 
        IndoLEM NTP~\cite{koto2020indolem} & IndoLEM next tweet prediction is a next tweet prediction dataset collected from tweeter & CC BY-SA 3.0 & 2020 & ind & Crawling \& human annotation & 8,382 & instances &  5681 train, 811 validation, 1890 test \\ \midrule 
        IndoLEM Sentiment Analysis~\cite{koto2020indolem} & IndoLEM Sentiment Analysis is a textual sentiment analysis dataset collected from twitter & CC BY-SA 3.0 & 2020 & ind & Crawling \& human annotation & 5,048 & sentences &  3638 train, 399 validation, 1011 test \\ \midrule 
        IndoLEM Tweet Ordering~\cite{koto2020indolem} & IndoLEM tweet ordering is a text ordering dataset collected from tweeter & CC BY-SA 3.0 & 2020 & ind & Crawling \& human annotation & 7,608 & instances &  5327 train, 760 validation, 1521 test \\ \midrule 
        IndoNLG Bible En-Id~\cite{cahyawijaya-etal-2021-indonlg} & Bible En-Id is a machine translation dataset containing Indonesian-English parallel sentences collected from the bible. We also add a Bible dataset to the English Indonesian translation task. Specifically, we collect an Indonesian and an English language Bible and generate a verse-aligned parallel corpus for the English-Indonesian machine translation task. We split the dataset and use 75\% as the training set, 10\% as the validation set, and 15\% as the test set. Each of the datasets is evaluated in both directions, i.e., English to Indonesian (En $\rightarrow$ Id) and Indonesian to English (Id $\rightarrow$ En) translations. & CC-BY-SA 4.0 & 2021 & eng, ind & Crawling \& human annotation & 31,078 & sentences &  23308 training, 3109 validation, 4661 test \\ \midrule 
        IndoNLG Bible Jv-Id~\cite{cahyawijaya-etal-2021-indonlg} & Bible Jv-Id is a machine translation dataset containing Indonesian-Javanese parallel sentences collected from the bible. Analogous to the En $\leftrightarrow$ Id and Su $\leftrightarrow$ Id datasets, we create a new dataset for Javanese and Indonesian translation generated from the verse-aligned Bible parallel corpus with the same split setting. In terms of size, both the Su $\leftrightarrow$ Id and Jv $\leftrightarrow$ Id datasets are much smaller compared to the En $\leftrightarrow$ Id dataset, because there are Bible chapters for which translations are available for Indonesian, albeit not for the local languages. & CC-BY-SA 4.0 & 2021 & jav, ind & Crawling \& human annotation & 7,957 & sentences &  5967 train, 797 validation, 1193 test \\ \midrule 
        IndoNLG Bible Su-Id~\cite{cahyawijaya-etal-2021-indonlg} & Bible Su-Id is a machine translation dataset containing Sundanese Indonesian parallel sentences collected from the bible. Analogous to the En $\leftrightarrow$ Id and Su $\leftrightarrow$ Id datasets, we create a new dataset for Javanese and Indonesian translation generated from the verse-aligned Bible parallel corpus with the same split setting. In terms of size, both the Su $\leftrightarrow$ Id and Jv $\leftrightarrow$ Id datasets are much smaller compared to the En $\leftrightarrow$ Id dataset, because there are Bible chapters for which translations are available for Indonesian, albeit not for the local languages & CC-BY-SA 4.0 & 2021 & sun, ind & Crawling \& human annotation & 7,958 & sentences &  5968 train, 797 validation, 1193 test \\ \midrule 
        IndoNLI~\cite{mahendra2021indonli} & IndoNLI is the first human-elicited Natural Language Inference (NLI) dataset for Indonesian. IndoNLI is annotated by both crowd workers and experts. The expert-annotated data is used exclusively as a test set. It is designed to provide a challenging test-bed for Indonesian NLI by explicitly incorporating various linguistic phenomena such as numerical reasoning, structural changes, idioms, or temporal and spatial reasoning. & CC-BY-SA 4.0 & 2021 & ind & Human generation \& curation & 17,712 & sentence pairs &  10330 train, 2197 validation, 5183 test \\ \midrule 
        IndoNLU NERGrit~\cite{wilie2020indonlu} & NER Grit dataset is a NER dataset taken from the Grit-ID repository, and the labels are spans in IOB chunking representation. The dataset consists of three kinds of named entity tags, PERSON (name of person), PLACE (name of location), and ORGANIZATION (name of organization). & CC-BY-SA 4.0 & 2020 & ind & Unknown & 2,090 & sentences & 1672 train, 209 validation, 209 test \\ \midrule 
        IndoPuisi & Puisi is an Indonesian poetic form. The dataset contains 7223 Indonesian puisi (poem) with its title and author. The data was scraped online using BeautifulSoup. The title and author column was produced using regex. & MIT & 2020 & ind & Machine generated / Crawling w/o curation & 7,223 & documents & No dataset split \\ \midrule 
        IndoSum~\cite{kurniawan2018indosum} & The Indosum dataset was collected from news aggregators covering six topics: entertainment, inspiration, sport, showbiz, headline, and technology. Compared to Liputan6, the summary label of Indosum is less abstractive, with novel 1-gram and novel 4-gram rates of 3.1\% and 20.3\%, respectively (Koto et al., 2020a). & CC-BY-SA 4.0 & 2021 & ind & Crawling \& human annotation & 18,773 & sentences &  14083 train, 1880 validation, 2810 test \\ \midrule 
        IndoTacos & IndoTacos dataset is tax court verdict summary collected from perpajakan.ddtc,co.id. It contains ~12k tax court summary with its verdict: mengabulkan seluruhnya, mengabulkan sebagian, menolak, mengabulkan, lain-lain.  This legal document is spesific for Indonesia tax cases. & CC-BY-NC-SA 4.0 & 2021 & ind & Machine generated / Crawling w/o curation & 12,291 & documents & No dataset split \\ \midrule 
        Indo\_MultiModal\_CC\_12M~\cite{changpinyo2021conceptual} & Conceptual 12M (CC12M) is a dataset with 12 million image-text pairs specifically meant to be used for visionand-language pre-training. Its data collection pipeline is a relaxed version of the one used in Conceptual Captions 3M (CC3M). Indo\_MultiModal\_CC\_12M is the Indonesian language version. & Custom\footnote{\url{https://github.com/google-research-datasets/conceptual-12m/blob/main/LICENSE}} & 2022 & ind & Machine generated / Crawling w/o curation & 1 & GB & No dataset split \\ \midrule 
        Indo\_MultiModal\_LAION~\cite{schumann2021laion} & Indo\_MultiModal\_LAION is a translated subset of the LAION-400M dataset with 70M image-text pairs specifically meant to be used for vision-language pre-training in Indonesian language. LAION-400M is a dataset with 400M English (image, text) pairs, filtered with OpenAI‘s CLIP by calculating the cosine similarity between the text and image embeddings and dropping those with a similarity below 0.3. The threshold of 0.3 had been determined through human evaluations and seemed to be a good heuristic for estimating semantic image-text-content matching. The image-text-pairs have been extracted from the Common Crawl web data dump and are from random web pages crawled between 2014 and 2021. More info for LAION-400M: https://laion.ai/blog/laion-400-open-dataset/. & CC-BY 4.0 & 2022 & ind & Machine generated / Crawling w/o curation & 7 & GB & No dataset split \\ \midrule 
        Indo\_MultiModal\_PMD\_ID~\cite{singh2022flava} & Introduced in the FLAVA paper, Public Multimodal Dataset (PMD) is a collection of publicly-available image-text pair datasets. PMD contains 70M image-text pairs in total with 68M unique images. The dataset contains pairs from Conceptual Captions, Conceptual Captions 12M, WIT, Localized Narratives, RedCaps, COCO, SBU Captions, Visual Genome and a subset of YFCC100M dataset. Indo\_MultiModal\_PMD\_Indonesia is the Indonesian language version. & CC-BY-4.0 & 2022 & ind & Machine generated / Crawling w/o curation & 15 & GB & 0 train, 0 validation, 0 test \\ \midrule 
        Indonesian Clickbait~\cite{william2020clickid} & The CLICK-ID dataset is a collection of Indonesian news headlines that was collected from 12 local online news publishers; detikNews, Fimela, Kapanlagi, Kompas, Liputan6, Okezone, Posmetro-Medan, Republika, Sindonews, Tempo, Tribunnews, and Wowkeren. This dataset is comprised of mainly two parts; (i) 46,119 raw article data, and (ii) 15,000 clickbait annotated sample headlines. Annotation was conducted with 3 annotator examining each headline. Judgment were based only on the headline. The majority then is considered as the ground truth. In the annotated sample, our annotation shows 6,290 clickbait and 8,710 non-clickbait. & CC-BY 4.0 & 2020 & ind & Crawling \& human annotation & 15,000 & headlines & No dataset split \\ \midrule 
        Indonesian Frog Storytelling Corpus~\cite{moeljadi2012usage} & Indonesian written and spoken storytelling corpus, based on the twenty-eight pictures. & Unknown & 2014 & ind & Unknown & 0 & documents & No dataset split \\ \midrule 
        Indonesian Google Play Review & Indonesian Google Play Review, dataset scrapped from e-commerce app on Google Play for sentiment analysis. & CC-BY 4.0 & 2022 & ind & Machine generated / Crawling w/o curation & 10,041 & sentences & 7028 train, 3012 validation, 0 test \\ \midrule 
        Indonesian Hoax News Detection~\cite{pratiwi2017study} & Indonesian Hoax News Detection is a dataset for hoax news detection. 600 Data are retrieved in Indonesian language with 372 valid news and 228 fake news.All data are manually labelled. & CC-BY 4.0 & 2018 & ind & Human generation \& curation & 600 & documents & No dataset split \\ \midrule 
        Indonesian Poem Tweets & Indonesian Poem tweets is dataset crawled from Twitter. The purpose of this data is to create text generation model for short text and make sure they are all coherence and rhythmic & CC-BY 4.0 & 2022 & ind & Machine generated / Crawling w/o curation & 16,427 & tweets & No dataset split \\ \midrule 
        Indonesian Stance~\cite{jannati2018stance} & ID Stance is a collection of Kompasiana articles that match with a pre-defined list of Indonesian politician names. Each article possesses a stance towards a candidate entity and election event, determined by annotators. Since the task is framed as a binary classification task, articles with no stance (neutral) are excluded from the gold-standard set. & CC-BY-NC-SA 4.0 & 2018 & ind & Crawling \& human annotation & 337 & documents & No dataset split \\ \midrule 
        Indonesian WSD~\cite{mahendra2018cross} & Indonesian WSD is a word sense dissambiguation dataset automatically collected using CrossLingual WSD (CLWSD) approach by utilizing WordNet and parallel corpus GIZA++. The monolingual WSD model is built from training data and it is used to assign the correct sense to any previously unseen word in a new context. The dataset covers 6 commonly ambiguous words, i.e., alam, atas, kayu, anggur, perdana, and dasar, with a total of 2416 sentences. & Unknown & 2018 & ind & Machine generated / Crawling w/o curation & 2,416 & sentences & No dataset split \\ \midrule 
        Indonesian general domain MT En-Id~\cite{guntara2020benchmarking} & For the general domain, both Tatoeba and TALPCo are manually curated, but their sentences (especially Tatoeba) are very short compared to Wikimatrix. Therefore, for these two datasets, we do a random split involving all datasets in the domain for validation and testing, each having 2000 unique pairs not present in the training set. For the general domain, we mix shorter sentences from TALPCo and the longer ones from Wikimatrix as our validation and test data. We observe that Tatoeba has similar types of high-quality sentences like TALPCo has, albeit shorter. Therefore we choose TALPCo to be in the validation and test sets instead, because longer sentences mean more difficult and meaningful evaluation. Tatoeba dataset contains short sentences. However, they contain high-quality full-sentence pairs with precise translation and is widely used in previous work in other languages (Artetxe and Schwenk, 2019b). Due to its simplicity, we do not use Tatoeba as our test and validation sets. We find that the Wikipedia scraper for Wikimatrix is faulty in some cases, causing some noise coming from unfiltered markup tags. & CC-BY-SA 4.0 & 2020 & eng, ind & Human generation \& curation & 1,811,300 & sentences &  1729472 train, 2000 validation, 2000 test \\ \midrule 
        Indonesian religious domain MT En-Id~\cite{guntara2020benchmarking} & Religious domain consists of religious manuscripts or articles. These articles are different from news as they are not in a formal, informative style. Instead, they are written to advocate and inspire religious values, often times citing biblical or quranic anecdotes.  The Tanzil dataset is a Quran translation dataset which has a relatively-imbalanced sentence length between the two languages, evidenced in Table 2, where an average Indonesian sentence in this dataset is about 50\% longer than an average English one. Furthermore, an average pair of sentences in this dataset would, on average, have one of them twice as long as the other. However, we still decide to include the dataset in the domain to avoid overfitting because the remaining datasets are all about Christianity. Another interesting property in the religion domain corpus is the localized names, for example, David to Daud, Mary to Maryam, Gabriel to Jibril, and more. In contrast, entity names are usually kept unchanged in other domains. We also find quite a handful of Indonesian translations of JW300 are missing the end sentence dot (.), even though the end sentence dot is present in their English counterpart. Lastly, we also find some inconsistency in the transliteration, for example praying is sometimes written as \ex{salat} or \ex{shalat}, or repentance as \ex{tobat} or \ex{taubat}. & CC-BY-SA 4.0 & 2020 & eng, ind & Human generation \& curation & 1,068,400 & sentences &  579544 train, 5000 validation, 4823 test \\ \midrule
        JATI~\cite{moeljadi2017building} & JATI is a treebank built from a subset of parsed dictionary definition sentences. The main data for this study comes from the fifth edition of Kamus Besar Bahasa Indonesia (KBBI) (Amalia, 2016), the official and the most comprehensive dictionary for the Indonesian language. The dictionary definition sentences are parsed using the Indonesian Resource Grammar (INDRA) (Moeljadi, Bond, and Song 2015), a computational grammar for Indonesian in the Head-Driven Phrase Structure Grammar (HPSG) framework (Sag, Wasow, and Bender, 2003). JATI will be employed to build an ontology, in which knowledge is extracted from the semantic representation in Minimal Recursion Semantics (MRS) (Copestake et al., 2005). & Unknown & 2017 & ind & Human generation \& curation & 1,253 & sentences & No dataset split \\ \midrule 
        JV-ID ASR~\cite{kjartanssoncrowd} & This dataset contains transcribed audio data for Javanese. The dataset consists of wave files, and a TSV file. The file utt\_spk\_text.tsv contains a FileID, UserID and the transcription of audio in the file. The dataset has been manually quality checked, but there might still be errors. This dataset was collected by Google in collaboration with Reykjavik University and Universitas Gadjah Mada in Indonesia. & CC-BY-SA 4.0 & 2018 & jav & Human generation \& curation & 185,076 & utterances & No dataset split \\ \midrule 
        JV-ID TTS~\cite{sodimana2018step} & This dataset contains high-quality transcribed audio data for Javanese. The dataset consists of wave files, and a TSV file. The file line\_index.tsv contains a filename and the transcription of audio in the file. Each filename is prepended with a speaker identification number. The dataset has been manually quality checked, but there might still be errors.This dataset was collected by Google in collaboration with Gadjah Mada University in Indonesia. & CC-BY-SA 4.0 & 2018 & ind, jav & Human generation \& curation & 5,800 & sentences & No dataset split \\ \midrule 
        JaDi-Ide~\cite{hidayatullah2020attention} & The dataset is collected from Twitter. We named the dataset as Javanese dialect identification (JaDi-Ide). The dialect is classified into Standard Javanese, Ngapak Javanese, and East Javanese dialects.  & Unknown & 2020 & jav & Crawling \& human annotation & 16,000 & tweets & No dataset split \\ \midrule 
        KEPS~\cite{mahfuzh2019improving} & KEPS is a keyphrase extraction dataset consists of text from Twitter discussing banking products and services and is written in the Indonesian language. A phrase containing important information is considered a keyphrase. Text may contain one or more keyphrases since important phrases can be located at different positions. The dataset follows the IOB chunking format, which represents the position of the keyphrase. & CC-BY-SA 4.0 & 2019 & ind & Crawling \& human annotation & 1,247 & documents & 1000 train, 247 test \\ \midrule 
        KaWAT~\cite{kurniawan2019kawat} & We introduced KaWAT (Kata Word Analogy Task), a new word analogy task dataset for Indonesian. We evaluated on it several existing pretrained Indonesian word embeddings and embeddings trained on Indonesian online news corpus. We also tested them on two downstream tasks and found that pretrained word embeddings helped either by reducing the training epochs or yielding significant performance gains. & Apache 2.0 & 2019 & ind & Human generation \& curation & 34,000 & sentence pairs & No dataset split \\ \midrule 
        Kamus Alay~\cite{salsabila2018colloquial} & We provide a lexicon for text normalization of Indonesian colloquial words. We gathered 3,592 unique colloquial words-also known as \ex{bahasa alay} -and manually annotated them with the normalized form. We built this lexicon from Instagram comments provided by Septiandri \& Wibisono (2017). & Unknown & 2018 & ind & Human generation \& curation & 3,592 & tokens & No dataset split \\ \midrule 
        Karonese dataset~\cite{karo2022sentiment} & Karonese dataset consist karonese text and the label (positive, negaitive or neutra). karonese text comes from multi domain social media, such us facebook, twitter, Instagram and Youtube & Unknown & 2022 & btx & Crawling \& human annotation & 1,001 & sentences & 0 train, 0 validation, 0 test \\ \midrule 
        Kethu~\cite{arwidarasti2019converting} & Kethu is a constituency treebank derived from Universitas Indonesia Constituency Treebank (UI-CTB) corpus Kethu converts UI-CTB treebank format into the widely accepted Penn Treebank format by adjusting the bracketing format for compound words as well as the POS tagset according to the Penn Treebank format. In addition, word segmentation and POS tagging of a number of tokens are also revised from the original UI-CTB corpus. & Unknown & 2019 & ind & Machine generated w/ human curation & 1,030 & sentences & No dataset split \\ \midrule 
        KoPI-CC (Korpus Perayapan Indonesia) & KoPI-CC (Korpus Perayapan Indonesia)-CC is Indonesian Only Extract from Common Crawl snapshots ,each snapshots get extracted using ungoliant oscar tools and get extra filtering using deduplication technique (Exact Hash Dup and Minhash LSH) & CC0 & 2022 & ind & Machine generated / Crawling w/o curation & 106 & GB & No dataset split \\ \midrule 
        KoPI-CC\_News & KoPI(Korpus Perayapan Indonesia)-CC\_News is Indonesian Only Extract from CC NEWS Common Crawl from 2016-2022(july) ,each snapshots get extracted using warcio and filter using fasttext & CC0 & 2022 & ind & Machine generated / Crawling w/o curation & 4 & GB & No dataset split \\ \midrule 
        KoPI-NLLB & KopI(Korpus Perayapan Indonesia)-NLLB, is Indonesian family language(aceh,bali,banjar,indonesia,jawa,minang,sunda) only extracted from NLLB Dataset each language set also filtered using some deduplicate technique such as exact hash(md5) dedup technique and minhash LSH neardup & ODC-BY & 2022 & ind, sun, jav, min, ban, bjn, ace & Machine generated / Crawling w/o curation & 18 & GB & No dataset split \\ \midrule 
        Korpus Nusantara~\cite{sujaini2020improving} & The dataset is a combination of multiple machine translation works from the author, Herry Sujaini, covering  Indonesian to 25 local dialects in Indonesia. Since not all dialects have ISO639-3 standard coding, as agreed with Pak Herry , we decided to group the dataset into the closest language family, i.e., Javanese, Dayak, Buginese, Sundanese, Madurese, Banjar, Batak Toba, Hakka (Khek), Malay, Minangkabau, and Min Nan (Teochew). & Unknown & 2022 & ind, sun, jav, min, mad, bbc, bug, msa, xdy, hak, nan & Human generation \& curation & 68,856 & sentence pairs & No dataset split \\ \midrule
        LibriVox-Indonesia \cite{wirawan_2022} & The LibriVox Indonesia dataset consists of MP3 audio and a corresponding text file we generated from the public domain audiobooks LibriVox. We collected only languages in Indonesia for this dataset. The original LibriVox audiobooks or sound files' duration varies from a few minutes to a few hours. Each audio file in the speech dataset now lasts from a few seconds to a maximum of 20 seconds. & CC0 & 2022 & ind, sun, jav, min, bug, ban, ace & Machine generated / Crawling w/o curation & 7,815 & utterances & No dataset split \\ \midrule 
        Liputan6 Summarization~\cite{koto2020liputan6} & The Liputan6 dataset was crawled from an online Indonesian news portal, which covers a wide range of topics, such as politics, sport, technology, business, health, and entertainment. There are two different experimental settings for Liputan6: Canonical, which includes all the test samples, and Xtreme, which only includes test samples with more than 90\% novel 4-grams in the summary label.  & CC-BY-SA 4.0 & 2021 & ind & Crawling \& human annotation & 224,637 & sentences &  193883 train, canonical: 10972 validation, 10972 test), extreme: 4948 validation, 3862 test),  \\ \midrule 
        Local ID Abusive~\cite{putri2021abusive} & Local ID Abusive is dataset aimed to be used for abusive and hate speech detection available in Javanese and Sundanese. The Javanese and Sundanese were annotated manually by annotator from each region. The annotation process involved multiple-step processes. It was carried by two annotators for each language, after an initial step where the guidelines were discussed and refined to reach unanimous comprehension. The annotation process gives 3449 and 2207 tweets for Javanese and Sundanese dataset respectively with 100\% agreement.  & Unknown & 2021 & sun, jav & Crawling \& human annotation & 5,656 & sentences & No dataset split \\ \midrule 
        MaRVL~\cite{liu2021visually} & Multicultural Reasoning over Vision and Language (MaRVL) is a dataset based on an ImageNet-style hierarchy representative of many languages and cultures (Indonesian, Mandarin Chinese, Swahili, Tamil, and Turkish). The selection of both concepts and images is entirely driven by native speakers. Afterwards, we elicit statements from native speakers about pairs of images. The task consists in discriminating whether each grounded statement is true or false. The present file contains all the dataset images and annotations.  & CC-BY 4.0 & 2021 & ind & Crawling \& human annotation & 1,128 & image, image, concept, caption & No dataset split \\ \midrule 
        MinangNLP MT~\cite{koto2020towards} & MinangNLP MT is a machine translation dataset generated from two sentiment analysis dataset by manual translation to Minangkabau language & MIT & 2020 & ind, min & Crawling \& human annotation & 5,000 & sentences &  11,571 train, 1600 validation, 3200 test \\ \midrule 
        MultiLexNorm~\cite{van2021multilexnorm} & MultiLexNorm is multilingual benchmark dataset for lexical normalization task for 12 languages, including Indonesian-English (code-mixed). Lexical normalization is the task of transforming an utterance into its standard form, word by word, including both one-to-many (1-n) and many-to-one (n-1) replacements. ID-EN dataset actually originates from Barik et.al. (2019) work. However, there is preprocessing work upon the original dataset. & CC-BY-NC-SA 4.0 & 2021 & ind, eng & Crawling \& human annotation & 13,949 & tokens & 13950 train, 4810 validation, 4367 test \\ \midrule 
        Multilingual Open Relations~\cite{faruqui2015multilingual} & This dataset provides the set of automatically extracted relations obtained using cross-lingual annotation projection method. The data covers 61 languages, including Indonesian. Relation extraction is the task of assigning a semantic relationship between a pair of arguments. For example, from the sentence Soekarno lahir di Jawa Timur, the relation <Soekarno, born\_in, Jawa Timur> is expected to be extracted. & Unknown & 2015 & ind & Machine generated / Crawling w/o curation & 1,876 & relations & No dataset split \\ \midrule 
        NER UGM (IndoLEM split)~\cite{fachri2014named} & NER UGM is a named entity recognition dataset collected by UGM. We use IndoLEM split for the dataset. & CC BY-SA 3.0 & 2014 & ind & Crawling \& human annotation & 2,343 & sentences &  1687 train, 187 validation, 469 test \\ \midrule 
        NER UI (IndoLEM split)~\cite{gultom2017automatic} & NER UI is a named entity recognition dataset collected by UI. We use IndoLEM split for the dataset. & CC BY-SA 3.0 & 2017 & ind & Human generation \& curation & 2,125 & sentences &  1530 train, 170 validation, 425 test \\ \midrule 
        NERGrit & NER Grit dataset is a NER dataset taken from the Grit-ID repository, and the labels are spans in IOB chunking representation. The dataset consists of three kinds of named entity tags, PERSON (name of person), PLACE (name of location), and ORGANIZATION (name of organization). & custom & 2020 & ind & Unknown & 17,437 & sentences & 12518 train, 2521 validation, 2398 test \\ \midrule 
        NERP (IndoNLU Split)~\cite{hoesen2018investigating} & NERP is a NER dataset which is collected from several Indonesian news websites, labelled with 5 entity classes: PER (name of person), LOC (name of location), IND (name of product or brand), EVT (name of the event), and FNB (name of food and beverage). & CC-BY-SA 4.0 & 2018 & ind & Human generation \& curation & 8,400 & sentences & 6720 train, 840 validation, 840 test \\ \midrule 
        NLLB Seed~\cite{costa2022no} & NLLB Seed is a set of professionally-translated sentences in the Wikipedia domain. Data for NLLB-Seed was sampled from Wikimedia’s List of articles every Wikipedia should have, a collection of topics in different fields of knowledge and human activity. NLLB-Seed consists of around six thousand sentences in 39 languages. NLLB-Seed is meant to be used for training rather than model evaluation. Due to this difference, NLLB-Seed does not go through the human quality assurance process present in FLORES-200. & CC-BY-NC 4.0 & 2022 & eng, ace, bjn, bug & Crawling \& human annotation & 30,965 & sentence pairs & No dataset split \\ \midrule 
        Netifier & Netifier dataset is a collection of scraped posts on famous social media sites in Indonesia, such as Instagram, Twitter, and Kaskus aimed to do multi-label toxicity classification. The dataset consists of 7,773 texts. The author manually labelled ~7k samples into 4 categories: pornography, hate speech, racism, and radicalism. & CC-BY-NC-SA 4.0 & 2018 & ind & Crawling \& human annotation & 7,773 & tweets & 6995 train, 0 validation, 778 test \\ \midrule 
        News En-Id MT~\cite{moeljadi2020ojw} & News En-Id is a machine translation dataset containing Indonesian-English parallel sentences collected from news translation dataset (Guntara et al., 2020).  The news dataset (Guntara et al., 2020) is collected from multiple sources: Pan Asia Networking Localization (PANL), Bilingual BBC news articles, Berita Jakarta, and GlobalVoices & CC-BY-SA 4.0 & 2021 & eng, ind & Crawling \& human annotation & 44,325 & sentences &  38469 train, 1953 validation, 1954 test \\ \midrule 
        NusaX MT~\cite{winata-etal-2023-nusax} & The first-ever parallel resource for 10 low-resource languages in Indonesia. & CC-BY-SA 4.0 & 2022 & ind, ace, jav, sun, min, bug, bbc, ban, nij, mad, bjn, eng & Human generation \& curation & 132,000 & sentence pairs & 500 train, 100 dev, 400 test \\ \midrule 
        NusaX Sentiment~\cite{winata-etal-2023-nusax} & The first-ever parallel resource for 10 low-resource languages in Indonesia. & CC-BY-SA 4.0 & 2022 & ind, ace, jav, sun, min, bug, bbc, ban, nij, mad, bjn, eng & Human generation \& curation & 12,000 & sentences & 500 train, 100 dev, 400 test \\ \midrule 
        OJW~\cite{moeljadi2020ojw} & OJW is abbreviation of Old Javanese Wordnet. & Unknown & 2020 & kaw & Unknown & 5,038 & tokens & No dataset split \\ \midrule 
        PANL BPPT~\cite{riza2009resource} & Parallel Text Corpora for Multi-Domain Translation System created by BPPT (Indonesian Agency for the Assessment and Application of Technology) for PAN Localization Project (A Regional Initiative to Develop Local Language Computing Capacity in Asia). The dataset contains around 24K sentences divided in 4 difference topics (Economic, international, Science and Technology and Sport). & Unknown & 2009 & eng, ind & Crawling \& human annotation & 24,000 & sentences & No dataset split \\ \midrule 
        POSP (IndoNLU Split)~\cite{hoesen2018investigating} & POSP is an Indonesian part-of-speech tagging (POS) dataset collected from Indonesian news websites. The dataset consists of around 8000 sentences with 26 POS tags following the Indonesian Association of Computational Linguistics (INACL) POS Tagging Convention. & CC-BY-SA 4.0 & 2018 & ind & Human generation \& curation & 8,400 & sentences & 6720 train, 840 validation, 840 test \\ \midrule 
        ParaCotta~\cite{aji2021paracotta} & ParaCotta is a synthetic parallel paraphrase corpus generated from monolingual data and a neural machine translation system. Multiple translations were generated using beam search, and then paraphrase pairs were selected based on the lexical difference determined by their sentence BLEU. & Unknown & 2021 & ind, eng & Machine generated / Crawling w/o curation & 6,000,000 & sentence pairs & No dataset split \\ \midrule 
        Parallel: Indonesian -  Lampung Nyo~\cite{abidin2021effect} & Parallel Indonesian - Lampung Nyo corpus is constructed from documents taken from the Lampung language book for elementary and junior high school levels in the province of Lampung. The document data that has been collected will then be manually typed to be made into a parallel corpus in Indonesian – Lampung dialect of nyo and mono corpus in Lampung dialect of nyo. There are 3000 parallel corpus sentences collected in Indonesian - Lampung dialect of nyo and 3000 mono corpus sentences in Indonesian - Lampung dialect of nyo. & Unknown & 2021 & abl, ind & Crawling \& human annotation & 3,000 & sentences & No dataset split \\ \midrule 
        PoSTagged Sundanese Monolingual Corpus~\cite{arie2022sunpos} & The dataset consists of 3616 Sundanese sentences collected from several Sundanese online magazine (Mangle, Dewan Dakwah Jabar, and Balebat). The dataset is mannualy annotated with Part of Speech label by several Sundanese Language faculty  students from UPI Bandung. & CC0 & 2015 & sun & Crawling \& human annotation & 3,616 & sentences & No dataset split \\ \midrule 
        SQuAD ID~\cite{muis2020sequence} & Stanford Question Answering Dataset (SQuAD) is a reading comprehension dataset, consisting of questions posed by crowdworkers on a set of Wikipedia articles, where the answer to every question is a segment of text, or span, from the corresponding reading passage, or the question might be unanswerable. This version is translated version of SQuAD to Indonesian Language. & Unknown & 2020 & ind & Machine generated / Crawling w/o curation & 124,631 & documents & 102657 train, 11407 validation, 10567 test \\ \midrule 
        STIF-Indonesia~\cite{wibowo2020semi} & STIF-Indonesia is an Indonesian text style transfer dataset collected from Twitter. The style transfer is done from informal style to a formal style. STIF-Indonesia consists of 52.5k sentences with 2.5k of which is manually annotated. & MIT & 2020 & ind & Crawling \& human annotation & 52.5k & sentences & 1922 train, 214 validation, 363 test \\ \midrule 
        SU-ID ASR~\cite{kjartanssoncrowd} & This dataset contains transcribed audio data for Sundanese. The dataset consists of wave files, and a TSV file. The file utt\_spk\_text.tsv contains a FileID, UserID and the transcription of audio in the file. The dataset has been manually quality checked, but there might still be errors. This dataset was collected by Google in Indonesia. & CC-BY-SA 4.0 & 2018 & sun & Human generation \& curation & 220,000 & sentences & No dataset split \\ \midrule 
        SU-ID TTS~\cite{sodimana2018step} & This dataset contains high-quality transcribed audio data for Sundanese. The dataset consists of wave files, and a TSV file. The file line\_index.tsv contains a filename and the transcription of audio in the file. Each filename is prepended with a speaker identification number. The dataset has been manually quality checked, but there might still be errors. This dataset was collected by Google in collaboration with Universitas Pendidikan Indonesia. & CC-BY-SA 4.0 & 2018 & sun & Human generation \& curation & 4,213 & utterances & No dataset split \\ \midrule 
        Sampiran~\cite{siallagan2023} & Sampiran is a dataset for pantun generation. It consists of 7.8K Indonesian pantun, collected from various sources (online). Pantun is a traditional Malay poem consisting of four lines: two lines of deliverance and two lines of message. This dataset filtered the gathered Pantun to follow the general rules of Pantun; four lines with ABAB rhyme and eight to twelve syllables per line. & AGPL-3.0 & 2023 & ind & Crawling \& human annotation & 7,879 & sentences & No dataset split \\ \midrule 
        Semeval STS Indo & SemEval is a series of international natural language processing (NLP) research workshops whose mission is to advance the current state of the art in semantic analysis and to help create high-quality annotated datasets in a range of increasingly challenging problems in natural language semantics. This is a translated version of SemEval Dataset from 2012-2016 for Semantic Textual Similarity Task to Indonesian language. & Unknown & 2019 & ind, eng & Machine generated / Crawling w/o curation & 12,901 & sentence pairs & 10293 train, 0 validation, 2608 test \\ \midrule 
        Singgalang~\cite{alfina2017modified} & Singgalang is an automatically tagged Indonesian NER dataset collected from Wikipedia articles which uses DBpedia as the reference of the entity type. The dataset used expanded DBpedia of MDEE\_Gazetteer as the reference to label the token and contains 48,957 sentences. & Unknown & 2017 & ind & Machine generated / Crawling w/o curation & 48,957 & sentences & No dataset split \\ \midrule 
        SmSA~\cite{purwarianti2019improving} & SmSA is a sentence-level sentiment analysis dataset consisting of of comments and reviews in Indonesian obtained from multiple online platforms with 3 possible sentiments:positive, negative, and neutral. The text was crawled and then annotated by several Indonesian linguists to construct this dataset.  & CC-BY-SA 4.0 & 2020 & ind & Crawling \& human annotation & 12,760 & sentences & 11000 train, 1260 validation, 500 test \\ \midrule 
        Sundanese Twitter Dataset for Emotion~\cite{putra2020sundanese} & Sunda Emotion dataset gathered from Twitter API between January and March 2019 with 2518 tweets in total. The tweets ﬁltered by using some hashtags which are represented Sun-danese emotion, for instance, \#persib, \#corona, \#saredih, \#nyakakak, \#garoblog, \#sangsara, \#gumujeng, \#bungah, \#sararieun, \#ceurik, and \#hariwang. This dataset contains four distinctive emotions: anger, joy, fear, and sadness.Each tweet is annotated using related emotion. For data validation, we consulted a Sundanese language teacher for expert validation & Unknown & 2020 & sun & Crawling \& human annotation & 2,518 & sentences & No dataset split \\ \midrule 
        Sundanese-Indonesian Parallel Corpus~\cite{arie2022suidparallel} & The dataset consists of 3616 Sundanese sentences taken from a Sundanese online magazing (Mangle), Dewan Dakwah Jabar, and Balebat. The dataset is manually translated to Indonesian language by several Sundanese Language faculty  students from UPI Bandung. & CC0 & 2015 & ind, sun & Human generation \& curation & 3,616 & sentence pairs & No dataset split \\ \midrule 
        TALPCo~\cite{nomoto2018tufs} & The TUFS Asian Language Parallel Corpus (TALPCo) is an open parallel corpus consisting of Japanese sentences and their translations into Korean, Burmese (Myanmar; the official language of the Republic of the Union of Myanmar), Malay (the national language of Malaysia, Singapore and Brunei), Indonesian, Thai, Vietnamese and English & CC-BY 4.0 & 2018 & ind, eng, kor, jpn, vie, tha, msa & Unknown & 0 & sentences & No dataset split \\ \midrule 
        TED En-Id MT~\cite{qi2018and} & TED En-Id is a machine translation dataset containing Indonesian-English parallel sentences collected from TED translation dataset (Qi et al., 2018), collected from TED talk transcripts & CC-BY-SA 4.0 & 2021 & eng, ind & Crawling \& human annotation & 93,262 & sentences &  87406 train, 2677 validation, 3179 test \\ \midrule 
        TICO-19~\cite{anastasopoulos2020tico} & TICO-19 (Translation Initiative for COVID-19) is sampled from a variety of public sources containing COVID-19 related content, representing different domains (e.g., news, wiki articles, and others). TICO-19 includes 30 documents (3071 sentences, 69.7k words) translated from English into 36 languages: Amharic, Arabic (Modern Standard), Bengali, Chinese (Simplified), Dari, Dinka, Farsi, French (European), Hausa, Hindi, Indonesian, Kanuri, Khmer (Central), Kinyarwanda, Kurdish Kurmanji, Kurdish Sorani, Lingala, Luganda, Malay, Marathi, Myanmar, Nepali, Nigerian Fulfulde, Nuer, Oromo, Pashto, Portuguese (Brazilian), Russian, Somali, Spanish (Latin American), Swahili, Congolese Swahili, Tagalog, Tamil, Tigrinya, Urdu, Zulu. & CC0 & 2020 & ind, eng, ara, spa, fra, hin, por, rus, zho & Machine generated w/ human curation & 21,497 & sentence pairs & 0 train, 6797 validation, 14700 test \\ \midrule 
        TITML-IDN~\cite{lestari2006large} & TITML-IDN (Tokyo Institute of Technology Multilingual - Indonesian) is collected to build a pioneering Indonesian Large Vocabulary Continuous Speech Recognition (LVCSR) System. In order to build an LVCSR system, high accurate acoustic models and large-scale language models are essential. Since Indonesian speech corpus was not available yet, we tried to collect speech data from 20 Indonesian native speakers (11 males and 9 females) to construct a speech corpus for training the acoustic model based on Hidden Markov Models (HMMs). A text corpus which was collected by ILPS, Informatics Institute, University of Amsterdam, was used to build a 40K-vocabulary dictionary and a n-gram language model. & For research purposes only. If you use this corpus, you have to cite (Lestari et al, 2006). & 2006 & ind & Human generation \& curation & 6,679 & utterances & No dataset split \\ \midrule 
        TUFS Indonesia Constituency Tree~\cite{nomoto2022tufs} & TUFS Indonesia Constituency Tree is annotated dataset for Indonesian language constituency tree. & CC-BY 4.0 & 2022 & ind, msa & Human generation \& curation & 1,385 & sentences & No dataset split \\ \midrule 
        TermA~\cite{fernando2019aspect} & The TermA span-extraction dataset is collected from the hotel aggregator platform, AiryRooms. The dataset consists of thousands of hotel reviews, which each contain a span label for aspect and sentiment words representing the opinion of the reviewer on the corresponding aspect. The labels use Inside-Outside-Beginning (IOB) tagging representation with two kinds of tags, aspect and sentiment. & CC-BY-SA 4.0 & 2019 & ind & Human generation \& curation & 5,000 & documents & 3000 train, 1000 validation, 1000 test \\ \midrule 
        Toxicity-200~\cite{costa2022no} & Toxicity-200 is a wordlist to detect toxicity in 200 languages. It contains files that include frequent words and phrases generally considered toxic because they represent: 1) frequently used profanities; 2) frequently used insults and hate speech terms, or language used to bully, denigrate, or demean; 3) pornographic terms; and 4) terms for body parts associated with sexual activity. & CC-BY-NC 4.0 & 2022 & ind, ace, bjn, bug, jav & Human generation \& curation & 226 & phrases & No dataset split \\ \midrule 
        TyDiQA Id~\cite{cahyawijaya-etal-2021-indonlg} & For the question answering task, we use the TyDiQA (Clark et al., 2020) dataset. This dataset is collected from Wikipedia articles with human-annotated question and answer pairs covering 11 languages. The question-answer pairs are collected for each language without using translation services. We use the Indonesian data from the secondary Gold passage task of the TyDiQA dataset. As the original dataset only provides training and validation sets, we randomly split off 15\% of the training data and use it as the test set. & CC-BY-SA 4.0 & 2021 & ind & Human generation \& curation & 6,267 & sentences &  4847 train, 565 validation, 855 test \\ \midrule 
        UD\_Indonesian-CSUI~\cite{alfina2020tree} & The UD\_Indonesian-CSUI is a dependency treebank in Indonesian in the CoNLL-U format. It was converted from a consituency treebank (Kethu) while Kethu was also converted from another consituency treebank (IDN treebank). Currently, this treebank, consist of 1030 sentences. & MIT & 2020 & ind & Machine generated w/ human curation & 1,030 & sentences & 656 train, 0 validation, 374 test \\ \midrule 
        UD\_Indonesian-GSD~\cite{mcdonald2013universal} & UD\_Indonesian-GSD is is an Indonesian-GSD treebank dataset originally converted from the content head version of the universal dependency treebank v2.0 (legacy) in 2015. In order to comply with the latest Indonesian annotation guidelines, the treebank has undergone a major revision between UD releases v2.8 and v2.9 (2021). & CC BY-SA 3.0 & 2013 & ind & Human generation \& curation & 5,593 & sentences &  4477 train, 559 validation, 557 test \\ \midrule 
        UD\_Indonesian-PUD ~\cite{alfina2019gold} & An Indonesian dependency treebank that is part of a collection of ~18 Parallel Universal Dependencies (PUD)  treebanks. & CC BY-SA 3.0 & 2019 & ind & Human generation \& curation & 1,000 & sentences &  10 fold cross validation according to UD standard \\ \midrule 
        UKARA 1.0 Challenge~\cite{septiandri2020ukara} & Ukara 1.0 Challenge dataset is a dataset for automatic short answer scoring system which is a collaboration project between FMIPA UGM and PUSPENDIK, Ministry of Education and Culture of Indonesia.  It was intended to build supervised machine learning approach which is able to assign a score to student’s answer. The student’s answer usually consists of maximum 2-3 sentences. & Unknown & 2020 & ind & Human generation \& curation & 2,861 & sentences & 268 train, 215 validation, 855 test \\ \midrule 
        Unimorph ID~\cite{pimentelqsigmorphon} & The Universal Morphology (UniMorph) project is a collaborative effort to improve how NLP handles complex morphology in the world’s languages. The goal of UniMorph is to annotate morphological data in a universal schema that allows an inflected word from any language to be defined by its lexical meaning, typically carried by the lemma, and by a rendering of its inflectional form in terms of a bundle of morphological features from our schema & Creative Commons Attribution-ShareAlike 3.0 Unported (CC BY-SA 3.0) & 2021 & ind & Crawling \& human annotation & 27,714 & forms &  70\% train, 10\% validation, 20\% test \\ \midrule 
        VoxLingua107~\cite{valk2021voxlingua107} & VoxLingua107 is a speech dataset for training spoken language identification models. The dataset consists of short speech segments automatically extracted from YouTube videos and labeled according the language of the video title and description, with some post-processing steps to filter out false positives. VoxLingua107 contains data for 107 languages, including Indonesian, Javanese, and Sundanese.  & CC-BY 4.0 & 2021 & ind, sun, jav & Crawling \& human annotation & 157 & hours & No dataset split \\ \midrule 
        WReTe~\cite{setyasemi} & WReTe is an entailment dataset which consists of 450 sentence pairs constructed from Wikipedia revision history. The dataset contains pairs of sentences and binary semantic relations between the pairs. The data are labeled as entailed when the meaning of the second sentence can be derived from the first one, and not entailed otherwise. & CC-BY-SA 4.0 & 2018 & ind & Crawling \& human annotation & 450 & sentence pairs & 300 train, 50 validation, 100 test \\ \midrule 
        WikiAnn~\cite{pan2017cross} & We developed a simple yet effective framework that can extract names from 282 languages and link them to an English KB. This framework follows a fully automatic training and testing pipeline, without the needs of any manual annotations or knowledge from native speakers. We release the following resources for each of these 282 languages: “silver-standard'' name tagging and linking annotations with multiple levels of granularity, morphology analyzer if it’s a morphologically-rich language, and an end to-end name tagging and linking system. & Attribution License (ODC-By) & 2017 & ind, eng, sun, jav, min, bug, bjn, tpi, ace, tdt, msa, jav-bms & Machine generated / Crawling w/o curation & 254,240 & Number of name mentions & No dataset split \\ \midrule 
        WikiLingua~\cite{ladhak2020wikilingua} & We introduce WikiLingua, a large-scale, multilingual dataset for the evaluation of crosslingual abstractive summarization systems. We extract article and summary pairs in 18 languages from WikiHow12, a high quality, collaborative resource of how-to guides on a diverse set of topics written by human authors. We create gold-standard article summary alignments across languages by aligning the images that are used to describe each how-to step in an article. & CC-BY-NC-SA 3.0 & 2020 & ind, eng & Crawling \& human annotation & 47,511 & article-summary pairs & No dataset split \\ \midrule 
        X-FACT~\cite{gupta2021x} & The largest publicly available multilingual dataset for factual verification of naturally existing real world claims. The dataset contains short statements in 25 languages and is labeled for veracity by expert fact-checkers. The dataset includes a multilingual evaluation benchmark that measures both out-of-domain generalization, and zero-shot capabilities of the multilingual models. & MIT & 2021 & ind, eng & Crawling \& human annotation & 3,548 & evidences-links-claim & 2231 train, 297 validation, 448 test \\ \midrule 
        XCOPA~\cite{ponti2020xcopa} & Cross-lingual Choice of Plausible Alternatives (XCOPA), a typologically diverse multilingual dataset for causal commonsense reasoning in 11 languages, including Indonesian. The causal commonsense reasoning task consists two task variations, forward causal reasoning and backward reasoning. In forward causal reasoning, given the premise sentence, the model is asked to predict most reasonable result from two alternative. In backward causal reasoning, the model is asked to predict what causes the premise happens & CC-BY 4.0 & 2021 & ind & Human generation \& curation & 600 & sentences & 0 train, 100 validation, 500 test \\ \midrule 
        XL-Sum~\cite{hasan2021xl} & XL-Sum is a comprehensive and diverse dataset comprising 1 million professionally annotated article-summary pairs from BBC, extracted using a set of carefully designed heuristics. The dataset covers 44 languages, including Indonesian. & CC-BY-NC-SA 4.0 & 2021 & ind & Crawling \& human annotation & 47,802 & document-summary pairs & 38242 train, 4780 validation, 4780 test \\ \midrule 
        XPersona Id~\cite{lin2021xpersona} & XPersona is a open-domain dialogue system on 7 languages including Indonesia. The test set is manually translated by exper annotators, while the training and validation set isareautomatically translated from the persona chat dataset with an additional manual keyword correction phase. & CC-BY-SA 4.0 & 2021 & ind & Machine generated w/ human curation & 17,866 & utterances &  16878 train, 484 validation, 484 test \\ \midrule 
        id-en-code-mixed~\cite{barik2019normalization} & This dataset contain 825 tweet instances of Indonesian-English, corresponding to four NLP tasks, i.e., tokenization, language identification, lexical normalization, and word translation. Data for lexical normalization task is curated in MultiLexNorm (already in Nusa Catalogue), but other tasks are not. Tokenization for social media data is not as trivial as splitting the token using white space delimiter. In this data, language identification is performed in token-level granularity. & CC-BY-NC-SA 4.0 & 2019 & ind, eng & Crawling \& human annotation & 22,736 & tokens & No dataset split \\ \midrule 
        Indo Wiki Paralel Corpora~\cite{trisedya2014wiki-idjv} & Manually aligned parallel corpora from Wikipedia & Unknown & 2014 & ind, sun, jav, min & Crawling \& human annotation & 2,422 & sentence pairs & No dataset split \\ \midrule 
        indo-law~\cite{nuranti2022lawprediction} & This dataset consists of Indonesian court decision documents for general criminal cases that have been annotated for the document sections. The documents were taken from the website of the Indonesian Supreme Court Decision. There are 22,630 documents with xml format in this dataset, which each contains 11 tags that enclose the annotated sections of the court decision documents.  & Unknown & 2022 & ind & Crawling \& human annotation & 22,630 & documents & No dataset split \\ \midrule 
        xSID~\cite{van2021masked} & We introduce XSID, a new benchmark for cross-lingual (X) Slot and Intent Detection in 13 languages from 6 language families, including a very low-resource dialect. & CC-BY-SA 4.0 & 2021 & ind, eng & Crawling \& human annotation & 5,370,460 & sentences & No dataset split \\
        \bottomrule
    \caption{Overview of all datasets in NusaCrowd.}
    \label{tab:stats-nusacrowd}
\end{longtable}
\end{landscape}

\end{document}